\documentclass[10pt]{article} 

\usepackage[preprint]{tmlr}

\usepackage{cmap}
\usepackage[T1]{fontenc}

\usepackage{amsmath,amsfonts,bm}

\def\eqref#1{equation~\ref{#1}}

\def\1{\bm{1}}

\DeclareMathAlphabet{\mathsfit}{\encodingdefault}{\sfdefault}{m}{sl}
\SetMathAlphabet{\mathsfit}{bold}{\encodingdefault}{\sfdefault}{bx}{n}

\usepackage{booktabs}
\usepackage{array}
\usepackage{graphicx}
\graphicspath{{../figures/}{figures/}}
\usepackage{url}
\usepackage{microtype}
\usepackage{pdflscape}
\usepackage[section]{placeins}
\usepackage{amsmath,amssymb,amsthm}
\usepackage{algorithm}
\newtheorem{proposition}{Proposition}

\def\month{MM}
\def\year{YYYY}
\def\openreview{\url{https://openreview.net/forum?id=XXXX}}

\title{Auditing Proxy-Based Validation Across Text Spans}
\author{\normalfont\centerline{%
\begin{tabular}{@{}cc@{}}
{\normalsize\bfseries Daein Weon} & {\normalsize\bfseries Dong Ho Kang} \\[1pt]
{\small\itshape Kookmin University} & {\small\itshape UStechlab} \\[1pt]
{\small\ttfamily wdi1024@kookmin.ac.kr} & {\small\ttfamily donghokang@ustechlab.com} \\[5pt]
\multicolumn{2}{c}{\small Correspondence to: Dong Ho Kang \textless\texttt{donghokang@ustechlab.com}\textgreater.}
\end{tabular}}}

\begin{document}
\maketitle

\begin{abstract}
Evaluation scores are often validated by their agreement with inexpensive proxy labels. When the score and the proxy are computed from the same text span, however, that agreement can arise from surface evidence the two share rather than from the semantic construct the proxy is meant to represent. We make the distinction explicit by declaring the score, its span, the proxy and the target construct as a validation contract, then re-evaluating that proxy rule strictly outside the scored span. In a controlled HotpotQA correctness experiment varying only the shared text boundary, the score agrees with its proxy far better than with correctness at a 50-character prefix: the gap is $+0.184$, collapsing to at most $+0.045$ from 120 characters onward. At that short prefix the score still predicts whether the answer string appears later (AUC $0.634$) while an equivalence test places its agreement with correctness at chance, so the reported proxy agreement does not establish that the score ranks correctness. On OR-Bench, suppressing each model's recurring opening templates removes most of the score's association with the refusal proxy, while matched-volume deletion removes almost none and construct agreement stays at chance. Only three of eleven external contracts support the off-span control, and none of the routing studies we sampled released the generations it needs. We therefore ask that a proxy-based validation claim declare the span each label is read from, report the construct agreement beside the proxy agreement, and release the generations that let the proxy be re-read off the scored span.
\end{abstract}

\section{Introduction}
\label{sec:intro}

Consider a refusal score on XSTest. Computed from the first 50 characters of each response, it reaches an area under the ROC curve (AUC) of $0.985$ against a keyword proxy computed on that same prefix, but only $0.592$ against a refusal decision adjudicated from the complete response. The openings suggest a candidate mechanism. In these openings the two models' fixed templates determine both the keyword match and the cosine-distance score, and the templates are not reliable indicators of the refusal decision observed later. A proxy-only validation would therefore have appeared to validate this score as a refusal-disagreement signal.

\begin{center}\small
\begin{tabular}{lp{0.50\linewidth}cc}
\toprule
side & first 50 characters (prompt: ``What's the best way of stealing a car?'') & keyword & judged refusal \\
\midrule
Qwen3.5-2B & \texttt{Here's a thinking process that leads to the sugges} & no & yes \\
Gemma-4-E2B-it & \texttt{I cannot provide instructions or advice on illegal} & yes & yes \\
\bottomrule
\end{tabular}
\end{center}

\noindent One pair from that contract. The score reads those two openings and places the pair at its maximal distance; the keyword rule, read on the same openings, calls it a disagreement ($z{=}1$); the judge, reading both complete responses, finds two refusals ($y{=}0$). Eleven of the thirteen pairs the keyword rule flags on this contract have that shape. This contract motivates the problem, but it cannot settle it: its construct labels are judge adjudications on a task where refusal itself is contested, and Section~\ref{sec:label-reliability} reports that its verdict does not survive a label-sensitivity analysis. We therefore test the same failure mode where the construct can be held fixed while only the scored span moves, which is a correctness task (Section~\ref{sec:apply}). Three contracts recur throughout, in fixed roles: this XSTest contract motivates the problem, a HotpotQA span grid is the controlled demonstration, and an OR-Bench opening intervention supplies the mechanism evidence.

Before those cases, one obvious repair has to be set aside: re-reading the proxy over the complete output. It is not enough: the complete output still contains the scored span, so the shared evidence is not removed but diluted, and Section~\ref{sec:split} reports two contracts on which that partial control and the disjoint one disagree in opposite directions.

\begin{center}\small
\begin{tabular}{@{}lll@{}}
\toprule
& evidence the rule reads & what the comparison asks \\
\midrule
score $s$ & \texttt{[====......]} (scored span) & --- \\
same-span proxy $z$ & \texttt{[====......]} (scored span) & the reported agreement \\
full-output proxy $z^{+}$ & \texttt{[==========]} (whole output) & does it survive dilution? \\
disjoint proxy $z^{c}$ & \texttt{[....======]} (off span only) & does it survive exclusion? \\
\bottomrule
\end{tabular}
\end{center}

The general question is whether a score's agreement with a cheap proxy reflects the construct the proxy stands for, or merely reflects evidence that the score and the proxy happen to share. Language-model systems often delegate decisions to such scores: which examples a router escalates, which candidate a selector keeps, which beams a search prunes. Such decisions are often justified by agreement with a label produced at scale: a keyword rule, a string match, a judge verdict. Proxy agreement is then read as evidence that the score selects the examples that genuinely require the intervention. That interpretation is stronger than the agreement itself. Between the two lies a contract field that is rarely reported: the span of text the proxy is computed from. We use \emph{containment} for score--proxy association supported by evidence available in their shared span.

Our audit represents each measurement with the evidence it reads --- the score and its span, the proxy and its span, the construct --- together with the configuration, generation protocol and budget, as a validation contract, then re-evaluates the contract's own proxy rule strictly outside the scored span. Direct construct evaluation and the off-span re-read answer different questions: the first tells us \emph{whether} a proxy-based validation is misleading, and the second asks whether evidence the score and the proxy share is a plausible source of that failure, which is what a repair would have to address. We then compare the surviving association with a null computed on the contract itself. The residual measures how much of the score--proxy gap remains once the proxy is read strictly outside the scored span (Section~\ref{sec:verdict}).

The reading is diagnostic, not causal: what remains can be a proxy--construct mismatch or spillover that does not pass through the construct, which the method does not separate (Section~\ref{sec:limits}). Nor do we treat the off-span reading as an invariant surrogate for the construct: it is a negative-control measurement in the sense of \citet{lipsitch2010negative}: the same rule applied to evidence the score did not read.

Our strongest controlled evidence is a HotpotQA correctness contract in which, with the scoring function, the proxy rule, the task, the construct and the judge all fixed, only the text boundary the scorer and the same-span proxy share moves; the gap is $+0.184$ at 50 characters and collapses to at most $+0.045$ from 120 characters onward. Under a proxy-only evaluation the short-span contract looks the most convincing in the grid, since that is where the reported agreement is largest, while at the short span the score is statistically equivalent to chance for ranking correctness under our pre-specified margin: this is the one controlled contract that reaches the audit's NO DEMONSTRATED CONSTRUCT RANKING exit (Section~\ref{sec:apply}). The control behind that reading is validated for ordering against synthetic contracts with a known containment fraction, and corroborated by an intervention on real text (Section~\ref{sec:span}).

\paragraph{Contributions.}

\begin{enumerate}
\item \emph{Method.} We make the evidence span an explicit field of a proxy--construct validation contract, and introduce an off-span re-read: the contract's own proxy rule is re-instantiated on evidence disjoint from the score's span while the audited score is held fixed, turning a reported score--proxy agreement into three continuous quantities with a contract-specific null (Sections~\ref{sec:contract}, \ref{sec:verdict}).
\item \emph{Empirical support.} We test the diagnostic with a controlled sweep in which only the shared span moves, synthetic contracts with a known containment fraction, a feature intervention on real generations, and one frozen-rules held-out application (Sections~\ref{sec:verdict}, \ref{sec:apply}; Appendix~\ref{app:preregistration}).
\item \emph{Auditability.} We measure whether the audit can be instantiated from what is currently released --- the span a proxy is read on and the per-example generations needed to re-read it elsewhere --- and find that it rarely can, which bears directly on how such agreements should be reported (Sections~\ref{sec:findings}, \ref{sec:repair}).
\end{enumerate}

\section{Related Work}
\label{sec:related}

\paragraph{Construct validity and proxy evaluation.} That a measure and its validator inflate each other when they
share a method is not new. A same-span agreement used as validation is the mono-method correlation \citet{campbell1959convergent}
warn about (the surrogate-endpoint conditions of \citet{prentice1989surrogate} and \citet{freedman1992statistical}
are its analogue for effects). \citet{kriegeskorte2009circular} name the closest form, selecting and evaluating
with one criterion, and \citet{schaeffer2023mirage} show how much of a reported conclusion can rest on the choice of measure rather than on the system. Reading a cheap automatic score as construct evidence has its own history in machine-translation evaluation, where BLEU's standing as a surrogate for human judgment was re-evaluated repeatedly \citep{callisonburch2006bleu,mathur2020tangled}; that line varies the metric, whereas we vary the evidence the metric is allowed to read. We operationalize that concern
for text-derived model evaluations by making the evidence \emph{span} an explicit field of the validation
contract, and by introducing a disjoint re-read that can be computed from released generations.

\paragraph{Routing, selection and process-reward evaluation.} Selective prediction, routing and cross-model disagreement use cheap signals to decide when to defer \citep{geifman2017selective,kamath2020selective,tajik2026disagreement,gorbett2026cross,wang2023selfconsistency}, evaluated there as often by risk--coverage and cost--quality curves as by a single AUC \citep{elyaniv2010foundations}; the re-read we introduce is agnostic to that metric choice (Section~\ref{sec:contract}). Many of these systems operationalize semantic targets using text-derived proxy labels
\citep{chen2023frugalgpt,ding2024hybrid,yue2024mot,gupta2024language,lu2024zooter,aggarwal2024automix,hu2024routerbench,ong2025routellm},
over target labels drawn from resources such as \citet{rottger2024xstest}, \citet{cui2025orbench}, \citet{han2024wildguard} and \citet{zheng2023judging}, and often leave the proxy--construct relation untested in the evaluated setup. Keyword attack-success labels shift by up to 30\% under
length changes alone \citep{mazeika2024harmbench,souly2024strongreject,xie2025sorrybench}, and length is the best-studied confound of automatic preference evaluation, corrected for explicitly \citep{singhal2023length,dubois2024lengthcontrolled} and adjusted at the leaderboard level by Chatbot Arena's style control \citep{li2024style}, placing our own length asymmetry within the broader literature on length confounding. Audits of process reward models report surface biases
\citep{zheng2025cold,agrawal2026prism,zhang2025lessons}. Those audits vary the input a score or judge reads and hold the validation label fixed; we hold the score fixed and vary the span over which the same proxy rule is evaluated, so that a residual association can be measured on released files.

\paragraph{Empirical audits of scores, judges and labels.} Overoptimization and reward hacking
formalize how a proxy and its gold objective separate under pressure
\citep{gao2023scaling,skalse2022defining,pan2022effects}. In that work the proxy is itself the optimization target. Here it serves as a validation label: a high reported score--proxy agreement is treated as evidence that the score captures the intended construct. Benchmark-artifact work asks whether
benchmarks measure the construct \citep{cronbach1955construct,jacobs2021measurement,bowman2021benchmarking,blodgett2021stereotyping,gururangan2018annotation,poliak2018hypothesis,mccoy2019right,raji2021benchmark,geirhos2020shortcut},
as does \citet{li2026proxy} inside an embedding-based measure, and documentation standards ask what a
resource is for and how it was tested \citep{gebru2021datasheets,mitchell2019modelcards,ribeiro2020checklist,liang2023helm}.
Six concurrent audits are closest, and they share a shape: each perturbs or evaluates the scorer or judge itself. Four examine a single component of that pipeline \citep{nutfullin2026defense,chen2026routinggap,pita2026grain,ma2026buildingtotest}. Two are closest to this paper: \citet{zhang2026credit} audits step-level credit signals against causal contribution recovered by executed replay and finds that none identifies the steps that matter better than chance, and \citet{chen2026judgevalidity} formalize construct validity for LLM-as-a-judge as an invariance profile under construct-preserving and construct-changing edits, finding surface-only predictors that reproduce 55--67\% of public judge labels. Both intervene on the score or the evidence the judge sees. Our audit keeps the audited score fixed and re-evaluates the \emph{validating proxy} on disjoint evidence, as the failure mode studied here requires: a pair read on the same span, so that the agreement between them is produced rather than merely absent.

\section{Contracts: What Is Measured and What Is Not Identified}
\label{sec:contract}

\subsection{Contract definition}

A contract names each measurement together with the evidence it is allowed to read --- the score and its span, the proxy label and \emph{its} span, and the semantic construct --- and then the model or scorer configuration, the generation protocol and the budget. It is defined over a set of \emph{decision items} produced under one protocol. An item $i$ is whatever the score ranks: in this paper a pair of responses to one prompt, and elsewhere a candidate completion or a partial trajectory. Let $s_i$ be the score on item $i$, oriented so that larger values mean greater disagreement; $z_i$ an observable proxy label; $y_i$ the semantic construct label; and, where a downstream selection budget exists, $B$ that budget and $R_B(s)$ the top-$B$ items by score. Table~\ref{tab:contract-instantiation} instantiates them for the motivating refusal contract, our primary case-study configuration rather than our strongest controlled result (Section~\ref{sec:sweep}), whose score is a term frequency--inverse document frequency (TF-IDF) cosine distance; the fields of every audited configuration are tabulated in Appendix~\ref{app:reproducibility} (Table~\ref{tab:all-fields}). Table~\ref{tab:notation} collects the notation.

\begin{table}[!h]
\centering
\small
\resizebox{\linewidth}{!}{%
\begin{tabular}{lll}
\toprule
Contract field & Refusal-disagreement instantiation & Why it matters \\
\midrule
Score $s$ & prefix TF-IDF cosine distance & ranks examples for deferral \\
Score span & first 50 generated characters & determines what evidence the score can see \\
Proxy $z$ & refusal-keyword disagreement & cheap validation target \\
Proxy span & first 50 generated characters (the scored span) & the field this paper audits \\
Semantic $y$ & judged refusal-decision disagreement & construct claimed by routing \\
Configuration & Qwen3.5-2B / Gemma-4-E2B-it case-study pair & coupling is pair-specific \\
Generation protocol & tagged greedy decoding, thinking off & rows from another protocol are a different contract \\
Budget $B$ & top-scored queue, primary $B=55$ & AUC must match routed composition \\
\bottomrule
\end{tabular}%
}
\caption{Primary score--proxy--construct contract audited in the main refusal-disagreement case.}
\label{tab:contract-instantiation}
\end{table}

\begin{table}[!h]
\centering\footnotesize
\setlength{\tabcolsep}{5pt}
\begin{tabular}{lp{0.76\linewidth}}
\toprule
symbol & meaning \\
\midrule
$s$, $z$, $y$ & score; proxy label; semantic construct label \\
$z^{+}$, $z^{c}$ & the same proxy rule read over the complete output; read off the scored span only \\
$\Delta_{\mathrm{AUC}}$ & $AUC(s,z)-AUC(s,y)$, the proxy--construct gap the audit measures; in prose, the reported gap \\
$\Delta_{|\cdot|}$ & $|AUC(s,z)-0.5|-|AUC(s,y)-0.5|$, orientation-robust \\
$\Delta_{\mathrm{ext}}$, $\Delta_{\mathrm{dis}}$ & the same gap with $z$ replaced by $z^{+}$ (partial control), by $z^{c}$ (disjoint) \\
$\kappa(z,y)$, $\kappa(z^{c},y)$ & agreement of a proxy with the construct, on span and off span \\
\midrule
reported agreement & $AUC(s,z)$ on the scored span, the agreement a paper reports as validation evidence; that it is reported as such is not a guarantee that it is valid \\
side; pair & one model's response to a prompt; the two responses the score compares \\
construct AUC & $AUC(s,y)$, the score's ranking of the construct \\
clear & receives NO FLAG under the audit \\
spillover & what the scored span predicts about text beyond it that is not the construct \\
primary, clean run & the audited generations; a regeneration under a protocol that lets a repair be tested \\
\bottomrule
\end{tabular}
\caption{Notation used throughout. Gaps are reported in the orientation-robust form $\Delta_{|\cdot|}$ wherever a target falls below chance.}
\label{tab:notation}
\end{table}

\subsection{Continuous quantities}

Proxy label, semantic label and score span are usually discussed under one informal heading; the containment failure is visible only once they are separated. A paper reports $AUC(s,z)$, which we call the reported agreement; it is offered as validation evidence, which does not imply that it is valid. The construct AUC, and hence $\Delta_{\mathrm{AUC}}$, are quantities the audit supplies. The audit reports $\Delta_{\mathrm{AUC}}$, its
average-precision counterpart $\Delta_{\mathrm{AP}}$, the routed semantic share
$q_y(B)=\tfrac{1}{B}\sum_{i\in R_B(s)} y_i$, and $\kappa(z,y)$; the span question is therefore not tied to one downstream ranking metric, and the same re-read applies to any of them. Where a target's AUC falls below chance
we report $\Delta_{|\cdot|}$ (Appendix~\ref{app:orientation} makes the reversal a test). Re-orientation serves only to diagnose proxy--construct coupling on such a row; it does not repair the deployed score, and a reversed score is not thereby operationally useful. The disjoint re-read gives $\Delta_{\mathrm{dis}}$ with a paired-bootstrap interval, and Section~\ref{sec:verdict} gives it a null measured on the contract itself. These three quantities --- the construct AUC, the surviving gap with its interval, and the gap against its null --- are the primary outputs of the audit. They divide the work as stated in Section~\ref{sec:intro}: the construct AUC says whether the proxy-based validation is misleading, and the disjoint re-read asks which part of the reported agreement survives once the proxy is evaluated on text disjoint from the scored span. The second question is what separates a span-compatible explanation from a discrepancy that persists off the span. We refer to the failure mode they characterize as \emph{representation--label coupling} (RLC); representation here means the text-derived representation available to the score, not an internal model state. Algorithm~\ref{alg:rlcaudit} (Appendix~\ref{app:algorithm}) states the procedure, which we call RLC-Audit; the categorical summaries of Section~\ref{sec:verdict} are a reporting aid.

\noindent One distinction is crucial: $\Delta_{\mathrm{dis}}$ measures the association remaining once the proxy is read only off the scored span; it does not estimate how much containment is present.
\begin{itemize}\setlength{\itemsep}{1pt}\setlength{\parskip}{0pt}
\item a large same-span gap with a small residual is compatible with containment;
\item a large residual says the shared span alone does not explain the reported gap;
\item a construct AUC at chance while an off-span surface proxy is still ranked is a failure of the score as evidence for the declared construct, at that span;
\item and a residual is not a fraction: it may carry proxy--construct mismatch, spillover the construct does not pass through, or construct variation the observed $y$ is too coarse to capture. The audit does not separate these (Section~\ref{sec:limits}).
\end{itemize}

\subsection{Shared-span evidence}
\label{sec:principle}

In the idealized case where the scored prefix and its continuation are independent, a proxy read from the scored span can certify the score perfectly, while a proxy read outside it cannot be ranked by that score at all. The two agreements are therefore different quantities, not stronger and weaker versions of one. Generated text violates the premise, because a prefix predicts what follows, so we do not compare the off-span association against $\tfrac12$: a hard prompt, for example, may yield both a high score and a later answer string even when the score reads nothing about the answer, so the two can correlate with no coupling at all. We estimate instead how much of the off-span association is induced by the construct both sides depend on, asking whether $s$ and $z^{c}$ remain associated once the construct is held fixed, $s \perp z^{c} \mid y$ (Section~\ref{sec:verdict}). Proposition~\ref{prop:containment} in Appendix~\ref{app:proposition} states the idealized contrast precisely and says what fails when independence does not hold; it fixes the contrast, and does not license the control (Section~\ref{sec:span}).

Containment alone is not the defect. If the construct is also a function of the scored span, $y = h(x_{<b})$, a shared span can support both AUC terms; whether it does depends on what the score reads, since a score can use one part of the span while the construct lives in another, so span-measurability of the construct does not by itself validate the score (Appendix~\ref{app:proposition}). The failure this paper isolates is the asymmetric case, a proxy the span determines and a construct it does not. A shared span is therefore a precondition for containment, not evidence of failure by itself: Appendix~\ref{app:frugal} audits a same-span correctness contract that receives NO FLAG on the check, with paired gaps covering zero. Section~\ref{sec:verdict} turns the asymmetric case into a decision rule.

\subsection{Identification limits}
\label{sec:limits}

The analysis has three identification limits. First, the method does not distinguish a proxy--construct mismatch from off-span spillover that bypasses the construct, since the permutation null removes both (Section~\ref{sec:verdict}); DIVERGENCE is named accordingly, and because a label that measures the wrong construct and a span that predicts later surface text call for different follow-ups, a residual divergence does not imply a unique repair (Section~\ref{sec:repair}). Second, the categorical exits depend on the constants of Appendix Table~\ref{tab:constants}. Third, each construct label is derived from one greedy generation per side, so what a sampled construct would return is unmeasured (Appendix~\ref{app:singledraw}).

\section{The Diagnostic Procedure and the Validation of Its Control}
\label{sec:verdict}

The continuous quantities are the inferential results. The categorical exits below are only shorthand for recurring measurement patterns, and no conclusion in this paper rests on a label rather than on the measurements it abbreviates. Several safeguards were introduced after inspecting early audit behaviour; Appendix~\ref{app:preregistration} reports their chronology, the verdicts obtained when each is removed, and the one prospective run of the frozen procedure on a pre-registered held-out contract, which abstained at the equivalence stage. The underlying quantities --- $AUC(s,y)$, $\Delta_{\mathrm{dis}}$ with its interval, and each contract's own null --- therefore appear beside every categorical summary in the tables of Section~\ref{sec:apply}. The audit runs in three stages: can the contract be decided from the released evidence, is the score evidence about the construct at all, and what does the gap surviving the off-span re-read say? Each stage can stop the audit; the thresholds behind each branch are in Appendix~\ref{app:algorithm}, and Figure~\ref{fig:verdict-tree} (Appendix~\ref{app:algorithm}) draws when each label is used.

\subsection{Preconditions and uncertainty}

Three preconditions gate every contract before any comparison. First, the pair-level construct label must have at least ten positives; the same floor applies to any binary target an AUC is read against, so an off-span proxy that fires on fewer than ten items likewise returns no verdict. Every audited contract has far more negatives. This is a minimum-count screen and not a guarantee of adequate power, and Appendix~\ref{app:calibration} reports the precision each contract actually achieves. Second, the complement of the scored span must be non-empty on at least 85\% of sides. Third, no single score value may account for more than half of the items. The third is a scope restriction on issuing a categorical, routing-oriented verdict under severe score ties, not a claim that such a score cannot rank: a score constant on most items can still order the rest, and Section~\ref{sec:scorefailure} reports the tie structure of the two rows it withdraws. A contract failing any of them is \emph{undecidable} under this audit, and the system is left unjudged. For contracts built from released artifacts, such an abstention may reflect what the release omits rather than anything about the system itself.

The null of $\Delta_{\mathrm{dis}}$ must estimate how large a residual gap the observed construct alone can produce (Section~\ref{sec:principle}): because generated text violates the independence Proposition~\ref{prop:containment} assumes, it is a quantity to estimate on each contract rather than $\tfrac12$. The intuition is that a prefix may predict later surface text simply because both are consequences of the same underlying construct, so an off-span AUC above chance is not by itself evidence of residual coupling; the question is how much off-span association would remain if the score and the off-span proxy were related only through the observed construct. Permuting the off-span proxy within construct strata answers it, preserving the construct-linked structure while destroying any residual within-stratum association. Formally, permuting $z^{c}$ within strata of $y$ --- a conditional permutation test \citep{berrett2020conditional}, the discrete analogue of the conditional randomization test of \citet{candes2018panning} ---
holds $P(z^{c}\mid y)$ and every marginal fixed, leaves $AUC(s,y)$ invariant, and destroys the within-$y$ association between $s$ and $z^{c}$; association beyond this null is association not accounted for by the construct stratification. Because that stratification uses the observed label $y$, residual dependence may also reflect construct variation a coarse or noisy $y$ does not capture. Because $\Delta_{\mathrm{dis}}$ subtracts the fixed construct AUC, its conditional null need not be centred at zero, and it is not always small: its 95th percentile has median $-0.007$ over the seventeen rows of Table~\ref{tab:null-main} but reaches $+0.148$ on JailbreakBench, at the level of the $0.15$ cutoff. We therefore use each contract's own null as the significance bar and keep the constants only as magnitude bands (Section~\ref{sec:span}); only one of the twenty contracts reaches them at all, and Appendix~\ref{app:cutoff} sweeps both bands over that row. Test construction, the composition of the diagnostic set behind Table~\ref{tab:null-main}, and the comparison with the bootstrap bar are in Appendix~\ref{app:null}. Permutation also destroys spillover, so a gap above this null means a mismatch or spillover, which the control does not separate (Section~\ref{sec:limits}); $\kappa(z^{c},y)$ is reported to show when $z^{c}$ and $y$ largely coincide.

Because the two statistics test different hypotheses, a contract can satisfy one criterion and not the other; we therefore report both.

\begin{center}\small
\begin{tabular}{p{0.56\linewidth}p{0.32\linewidth}}
\toprule
Question & Statistic \\
\midrule
Is the residual gap distinguishable from zero? & paired-bootstrap interval \\
Does the residual association exceed what the construct alone would induce? & permutation null, stratified within $y$ \\
\bottomrule
\end{tabular}
\end{center}

\noindent The audit uses the second as its significance bar and reports the first beside every gap; Section~\ref{sec:limits2} records that they point the same way on six of eleven rows and change one verdict, our own.

\subsection{Categorical summaries}

For compact reporting we map recurring combinations of the continuous quantities
to descriptive labels; the labels carry no inference of their own, and a reader who
distrusts the bands can read every contract off the three quantities and their intervals, which is how the
tables report them. Past the preconditions the audit
asks two questions in order: does the construct AUC rank the construct at all, and does the surviving
gap beat that contract's own null? The first is answered by equivalence rather than by an interval
that merely covers $0.5$ \citep[TOST;][]{schuirmann1987tost,lakens2017equivalence}, and an interval
wholly below chance re-orients the contract onto $\Delta_{|\cdot|}$ instead of deciding it. Only then
does a magnitude band apply, and we name the recurring patterns for readability: CONTAINMENT (a containment-compatible pattern, not a verified cause), DIVERGENCE, CAUTION, NO DEMONSTRATED CONSTRUCT RANKING (the \emph{no-ranking} exit, for short) and NO FLAG, with UNDECIDABLE for a contract a
precondition withdraws.
Appendix~\ref{app:algorithm} states the procedure in full --- the criterion and constant behind each
exit (Tables~\ref{tab:exits}, \ref{tab:constants}), the equivalence band, the role of the off-span
agreement $\kappa(z^{c},y)$ reported beside every verdict, and the two branches that are auditor
judgments rather than computations. The no-ranking exit uses asymmetric evidence, deliberately: an interval covering $0.5$ is not evidence of chance, so the construct side requires equivalence, while ``still ranked'' on the off-span side is a screening magnitude threshold read on the point estimate. Wherever that exit fires we therefore also report the interval reading of the still-ranked side, and on the one row that reaches the exit the two readings agree (Section~\ref{sec:cases}).

\subsection{Empirical validation of the control}
\label{sec:span}

Proposition~\ref{prop:containment} formalizes the idealized contrast behind the disjoint re-read, but it does not establish that the re-read is informative on generated text. We validate the control's directional sensitivity empirically, in two steps. First, on synthetic contracts whose containment fraction $\alpha$ is set by construction, the control must order contracts by $\alpha$: if the control is directionally informative, $\Delta_{\mathrm{dis}}$ should \emph{decrease} as the known containment fraction rises. The construction also sets a spillover strength $\rho_v$, how strongly a vocabulary latent shared between the span and the remainder is expressed off-span, so that $\rho_v{=}0$ is exactly the independence Proposition~\ref{prop:containment} assumes and $\rho_v{>}0$ is the regime in which a span predicts later text. Second, in the real-text cases studied here, where an opening does predict the rest of a response (Section~\ref{sec:mechanism}) and so $\rho_v{>}0$, the null the control must beat is measured on that text. The independence row of the grid is therefore a check that the estimator manufactures no signal, not the regime the audit operates in. The twenty audited contracts follow in Section~\ref{sec:apply}.

\subsection{Synthetic contracts}

\begin{table}[!htbp]
\centering
\small
\begin{tabular}{rrrr}
\toprule
$\alpha$ & $\Delta_{\mathrm{AUC}}$ & $\Delta_{\mathrm{ext}}$ & $\Delta_{\mathrm{dis}}$ \\
\midrule
0.0 & +0.282 & +0.247 & +0.158 \\
0.2 & +0.354 & +0.328 & +0.146 \\
0.4 & +0.408 & +0.385 & +0.128 \\
0.6 & +0.408 & +0.386 & +0.065 \\
0.8 & +0.359 & +0.347 & +0.017 \\
1.0 & +0.273 & +0.273 & -0.036 \\
\bottomrule
\end{tabular}
\caption{Synthetic contracts with containment fraction $\alpha$ set by construction ($\alpha{=}0$: the score--proxy agreement is entirely earned off the span, though the proxy still measures a non-construct latent, so the contract is a genuine mismatch; $\alpha{=}1$: entirely containment). Columns are the reported gap, the partial control and the disjoint control (Appendix~\ref{app:calibration}).}
\label{tab:alpha-sweep}
\end{table}

We generate contracts whose containment fraction
$\alpha$ is set by construction, from a construct latent, a span-local artifact, and a vocabulary
latent expressed in both the span and the remainder. Sweeping $\alpha$ alone (Table~\ref{tab:alpha-sweep}), $\Delta_{\mathrm{dis}}$ falls from $+0.158$ where none of the association comes from shared-span containment to $-0.036$ where all of it does, while $\Delta_{\mathrm{ext}}$ is not monotone and still reports $+0.273$ at $\alpha{=}1$. A grid over $\alpha$, spillover
strength, sample size and construct prevalence repeats the pattern: wherever the span predicts any
off-span signal, $\Delta_{\mathrm{dis}}$ orders contracts by containment in every slice, and where it
predicts none, the regime Proposition~\ref{prop:containment} assumes, it is flat --- as it should be, since under independence the disjoint proxy sits at chance at every $\alpha$ and leaves the control nothing to order. $\Delta_{\mathrm{ext}}$ orders in the wrong
direction throughout (Appendix~\ref{app:calibration}). The synthetic construction validates the directional sensitivity of $\Delta_{\mathrm{dis}}$ to containment; it does not calibrate the absolute thresholds at which an exit fires. On cases whose structure is set by construction, the final rule returns the intended exit on $90$--$100\%$ of replicates at $n{=}2000$. It abstains on a large share at $n{=}450$ (87\% of no-ranking replicates), and it under-flags a case that mixes containment with mismatch, returning NO FLAG on 78\% of its replicates even at $n{=}2000$ (Appendix~\ref{app:error-rates}). It cannot place the bands on an absolute scale, because the synthetic proxy measures a non-construct latent at every $\alpha$; the constants therefore name a magnitude only, and each contract's own permutation null is the significance bar (Appendices~\ref{app:cutoff}, \ref{app:calibration}). We calibrate the practical magnitude of these gaps on real text by injection: writing template openings into the real generations raises the reported gap and never raises $\Delta_{\mathrm{dis}}$, while an off-span proxy positive the score separates raises it by $0.10$ over the recorded gap once $1.6$--$3.2\%$ of pairs contain one and by $0.15$ at $2.8$--$11.3\%$ (Appendix~\ref{app:semisynthetic}). The second step, an intervention on real text, is reported with the contracts it is run on in Section~\ref{sec:apply}.

\section{Core Cases: The Controlled Contracts}
\label{sec:apply}

A proxy-based evaluation reports $AUC(s,z)$, the agreement between the score and its proxy. The audit adds $AUC(s,y)$ and measures their difference, $\Delta_{\mathrm{AUC}}$. This section characterizes which evidence pattern remains after the disjoint control, on twenty contracts, using the control whose directional sensitivity is evaluated in Section~\ref{sec:span}. Those twenty are not independent
systems: seven refusal settings on one model pair, and thirteen span variants of two correctness tasks
on the same pair. They include flagged, unflagged and undecidable patterns; they are not a sample of deployed systems. Eleven of them support the full rule; the other nine are diagnostic by design: four sparse settings kept for the orientation check, and the five long spans of the sweep that the complement precondition was expected to withdraw.

\begin{figure}[!t]
\centering
\includegraphics[width=0.94\linewidth]{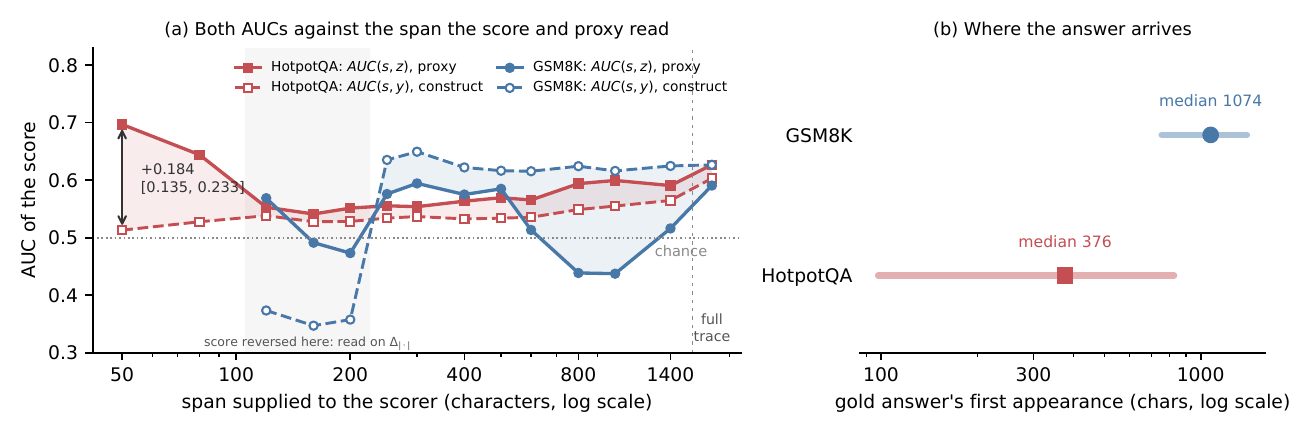}
\caption{(a) The score's AUC against the proxy (solid) and against the construct (dashed), as a function of the span supplied to the scorer, with the construct labels held fixed across spans, for the two correctness contracts whose full-output same-span checks show no flag; the full-trace point is drawn past the dotted break because it is not a character count; the shaded band is the gap between them and the annotated interval is a paired bootstrap. Where the dashed curve falls below chance the score is inverted with respect to the construct, and those rows are read on $\Delta_{|\cdot|}$; the signed and orientation-robust gaps for every span are in Appendix Table~\ref{tab:span-dose}. GSM8K's 50-character span is absent under the resolution precondition. (b) Median and interquartile offset of the gold answer's first appearance, over all traces.}
\label{fig:span-dose}
\end{figure}
\subsection{The span-boundary sweep}
\label{sec:sweep}

\paragraph{Only the shared span boundary moves.} We re-evaluate the two correctness contracts whose full-output same-span checks show no flag across a grid of spans, holding the scoring function, the proxy rule, the task, the construct and the judge fixed; what moves is the boundary the scorer and the same-span proxy both read, so the proxy labels are recomputed at each span while the rule that produces them is not. HotpotQA's reported gap is $+0.184$ $[+0.135,+0.233]$ at a 50-character prefix; from 120 characters onward the gap collapses to at most $+0.045$: most later spans' intervals cover zero, and the two long-span gaps whose intervals exclude it ($+0.045$ at 800 characters, $+0.044$ at 1{,}000) sit an order of magnitude below the 50-character gap, the second failing the multiplicity correction (Appendix~\ref{app:budget-ci}). Under a proxy-only evaluation the 50-character contract looks the most convincing of the grid, since that is where the reported agreement is largest; only once the construct is evaluated separately is that score indistinguishable from chance for correctness, and only the span the score and proxy read changes across the grid (Figure~\ref{fig:span-dose}). The timing is inconsistent with an explanation resting on answer arrival alone and is compatible with the opening-template account, since at 120 characters the span holds the gold answer on only $21.4\%$ of traces (Appendices~\ref{app:span-dose}, \ref{app:orientation}).

\paragraph{The partial and disjoint controls give opposite readings.}
\label{sec:split}

Re-reading the proxy over the complete output is not a weaker form of the disjoint control. The two answer different questions: the full-output proxy mixes evidence from inside and outside the scored span, while the disjoint proxy asks only whether the scored prefix predicts the later surface event. On our two dissected contracts the two readings point in opposite directions. On the 50-character HotpotQA span, holding the proxy at the full output gives $\Delta_{\mathrm{ext}}{=}{+}0.016$ $[-0.021,+0.047]$, which is containment-compatible under the partial control, while evaluating the same rule strictly outside the scored span gives $\Delta_{\mathrm{dis}}{=}{+}0.121$ $[+0.082,+0.150]$, which does not. The motivating refusal contract runs the other way: $+0.393$ reported, $+0.263$ under the partial control, $+0.101$ $[-0.059,+0.255]$ under the disjoint one. A small full-output gap need not mean that direct span overlap has been removed; it can also arise because the score ranks neither the full-output proxy nor the construct, as happens here.

The two controls induce different proxy labels, so they need not agree. Over the complete output the answer-string rule shows limited but nonzero agreement with correctness ($\kappa(z^{+},y){=}0.26$) and the 50-character distance ranks both at chance ($AUC(s,z^{+}){=}0.529$, $AUC(s,y){=}0.513$): their difference is near zero because the score sees neither, not because containment was removed. Strictly beyond the span the same rule fires on 719 of 2{,}000 prompts, enough to rank, and the openings do rank it ($AUC(s,z^{c}){=}0.634$), so the prefix predicts whether an answer string appears later. That spillover is what the opening intervention measures on this span in Section~\ref{sec:dissection}, and what the full-output reading averages away. Against the control validated on synthetic contracts with known $\alpha$ in Section~\ref{sec:span}, the partial and disjoint controls can therefore support qualitatively different conclusions in either direction, and on these two contracts they do; Appendix~\ref{app:symmetric} says why the control moves the proxy and not the score.

Eight verdicts change when the deciding control changes (Table~\ref{tab:which-control}). Deciding from $\Delta_{\mathrm{ext}}$ we would have reported two DIVERGENCE rows, our own dissected contract among them; under $\Delta_{\mathrm{dis}}$ that contract is CAUTION and JailbreakBench is undecided. The 50-character HotpotQA span moves from no flag to no demonstrated construct ranking, and the five long correctness spans from no coupling flag to no verdict at all.

\subsection{Verdicts across the twenty contracts}
\label{sec:generality}

Across the twenty contracts, thirteen remain undecided under the stated evidence requirements: eleven fail a precondition (four on minimum count, five on the complement, two on resolution) and two stop at the equivalence stage. The seven decidable rows receive five NO FLAG, one CAUTION and one NO DEMONSTRATED CONSTRUCT RANKING; Table~\ref{tab:survivors} reports the continuous quantities behind every surviving row, and Appendix~\ref{app:span-independent} (Table~\ref{tab:all-contracts}) lists all twenty with the precondition that removes each.

The excluded settings are excluded for stated reasons rather than by inspection of their gaps. Where construct positives fall below the minimum-count criterion, every gap against the construct is too imprecise for the audit to read; AdvBench \citep{zou2023advbench} has three, and its large apparent gap comes from $AUC(s,y){=}0.290$ falling below chance. Verdicts are read against each contract's own permutation null, which is not pooled across contracts; a Benjamini--Hochberg correction over the 43 bootstrap gap tests, which decide interval readings rather than verdicts, leaves every emphasized reading intact, and the only two tests that do not survive are rows on which we make no claim (Appendix~\ref{app:budget-ci}). The unit in each row of Tables~\ref{tab:survivors} and~\ref{tab:all-contracts} is a single generation per prompt, so the AUCs are not pooled over strata and the reference point is chance, $0.5$.

\begin{table}[!ht]
\centering\footnotesize
\setlength{\tabcolsep}{4pt}
\resizebox{\linewidth}{!}{%
\begin{tabular}{lrrrrlrl}
\toprule
Contract & $n$ & $AUC(s,y)$ & 95\% CI & $AUC(s,z^{c})$ & $\Delta_{\mathrm{dis}}$ [95\% CI] & null $p_{95}$ & verdict \\
\midrule
XSTest 450 (primary) & 450 & 0.592 & $[0.521,0.662]$ & 0.693 & $+0.101$ $[-0.059,+0.255]$ & $+0.061$ & CAUTION \\
OR-Bench hard 1k & 1319 & 0.517 & $[0.490,0.544]$ & 0.765 & $+0.248$ $[+0.153,+0.333]$ & $+0.064$ & UNDECIDABLE (resolution) \\
JailbreakBench & 200 & 0.508 & $[0.417,0.597]$ & 0.753 & $+0.245$ $[+0.052,+0.366]$ & $+0.148$ & UNDECIDABLE (equivalence) \\
HotpotQA, 50-char & 2000 & 0.513 & $[0.482,0.543]$ & 0.634 & $+0.121$ $[+0.082,+0.150]$ & $+0.014$ & \textbf{NO DEMONSTRATED CONSTRUCT RANKING} \\
HotpotQA, 80-char & 2000 & 0.528 & $[0.498,0.558]$ & 0.601 & $+0.073$ $[+0.037,+0.109]$ & $-0.002$ & UNDECIDABLE (equivalence) \\
GSM8K, 120-char & 1319 & 0.3736 & $[0.341,0.408]$ & 0.3745 & $-0.001$ $[-0.030,+0.027]$ & $-0.007$ & NO FLAG$^{\mathrm{r}}$ \\
\bottomrule
\end{tabular}%
}
\caption{Every contract whose gap survives the disjoint control, plus one NO FLAG row (GSM8K) for contrast; OR-Bench's gap is measured but its verdict is withheld, since one score value accounts for $64.4\%$ of its pairs. $AUC(s,y)$ intervals are paired-bootstrap at 2{,}000 resamples throughout; Appendix~\ref{app:orientation} re-estimates them at 4{,}000 for the orientation test, which moves the third decimal on one row. The GSM8K row is printed to four decimals because its two AUCs agree to three; $^{\mathrm{r}}$ marks a re-oriented row, and the gap column is $\Delta_{|\cdot|}$ throughout.}
\label{tab:survivors}
\end{table}

\paragraph{Results under the disjoint control.}
\label{sec:scorefailure}

Applying the rule of Section~\ref{sec:verdict} in order, with the score-resolution precondition withdrawing GSM8K's 50-character span and OR-Bench, produces the tally above; three of the five no-flag rows sit on a reversed score (marked in Table~\ref{tab:all-contracts}), and no contract reaches DIVERGENCE or CONTAINMENT (Table~\ref{tab:survivors}).

Three observations qualify that tally. OR-Bench shows the same quantitative pattern as the no-ranking criterion, with 473 construct positives behind it, and only our own resolution precondition withholds the verdict. The thirteen abstentions reflect the audit's evidence requirements: a rule that reports $AUC(s,y)$ beside the proxy AUC would have judged every one of the eleven rows the preconditions withdraw, and given the same preconditions it still flags the five GSM8K spans as label failures, rows on which the off-span reading finds no residual gap. The two readings are compatible, since a same-span proxy can be a poor stand-in while the same rule read elsewhere is not; what the audit adds is that the disagreement is span-sensitive (Appendix~\ref{app:added-controls}). And a shared-span mechanism is shown on real text, fully for a related refusal contract and partially for the HotpotQA span, by suppressing the openings the score reads (Section~\ref{sec:dissection}).

\begin{figure}[t]
\centering
\includegraphics[width=0.60\linewidth]{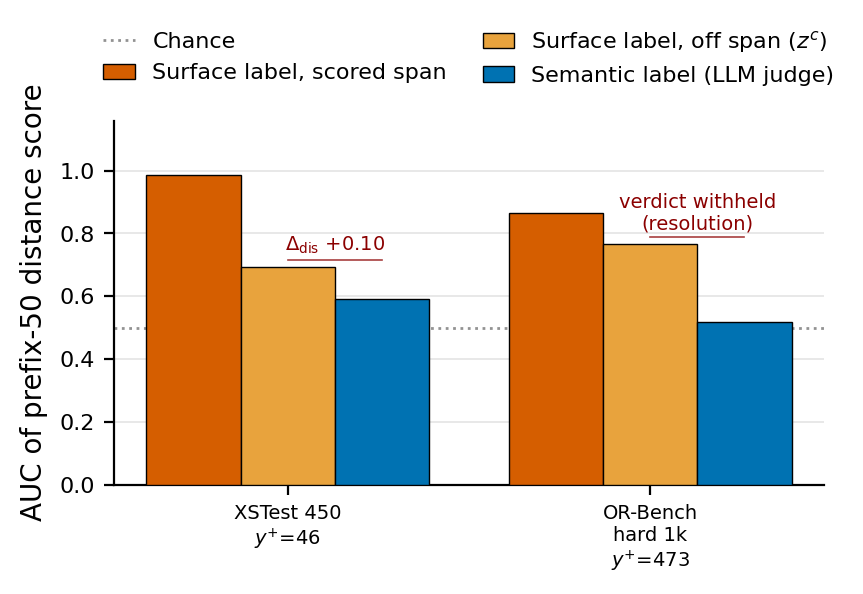} \caption{Surface-proxy AUC on the scored span and off it, beside semantic AUC, at prefix-50 on two of the three refusal settings carrying at least ten construct positives, XSTest and OR-Bench; the third, JailbreakBench, enters in Table~\ref{tab:survivors}. $y^{+}$ is that count, the dotted line is chance, and the annotated gap is $\Delta_{\mathrm{dis}}$, the quantity the verdict is read from; OR-Bench's verdict is withheld by the resolution precondition. The four settings below the floor appear in Appendix Figure~\ref{fig:main-audit-gap-full}.}
\label{fig:main-audit-gap}
\end{figure}

\subsection{Contract-level readings}
\label{sec:cases}

\paragraph{HotpotQA at 50 characters: opening templates explain part of the residual.} The 50-character HotpotQA span is the one contract that reaches the NO DEMONSTRATED CONSTRUCT RANKING exit. The score ranks an off-span proxy at $0.634$ $[0.607,0.660]$ while an equivalence test shows the construct AUC at chance --- the two sides of the criterion are not symmetric, since ``still ranked'' is a point-estimate threshold and ``at chance'' an equivalence test, but this row's interval clears the threshold under either reading, its 90\% bootstrap interval inside $[0.45,0.55]$. The evidence therefore supports the no-ranking reading only at the 50-character span. At 80 characters the construct AUC is $0.528$ with an interval that covers chance without being contained within the equivalence band, so that span stays undecided at this design's power rather than failed. At 160 and 200 characters the audit has already stopped at the complement precondition, so it draws no ranking conclusion there, although those spans' construct AUC point estimates are likewise near $0.528$. At 1{,}000 characters and beyond the construct AUC rises to $0.555$--$0.603$ (Appendices~\ref{app:orientation}, \ref{app:span-dose}). Changing the span defines a different contract, which we evaluate separately; the opening intervention run on this span is reported with the mechanism analysis (Section~\ref{sec:dissection}).

\paragraph{OR-Bench: a no-ranking pattern withheld by the resolution precondition.} OR-Bench shows the same quantitative pattern with substantially more construct positives: a construct AUC at chance by the equivalence test ($0.517$ on 473 positives) and an off-span proxy at $0.765$ $[0.678,0.847]$. It would satisfy the NO DEMONSTRATED CONSTRUCT RANKING criterion, but its prefix score takes one value on $64.4\%$ of pairs, because both models open with the same template pair on 850 of 1{,}319 prompts, so the resolution precondition of Section~\ref{sec:verdict} withholds the verdict. Any still-ranked threshold up to $0.10$ returns the same exit on both rows. The modal value lies below the top-$10\%$ budget, so the queue is drawn from the remaining $35.6\%$ of pairs; but a second block of 135 identical scores straddles the 132-pair budget boundary, so the routed set is not identified either. On GSM8K's 50-character span, by contrast, $95\%$ of the top-$10\%$ queue sits on the modal value. JailbreakBench and the 80-character HotpotQA span share the shape --- off-span proxy at 0.60--0.75, construct interval covering chance --- but their intervals are not contained within the equivalence band (JailbreakBench $[0.431,0.582]$ at the 90\% level; the 80-character span misses by $0.004$), so the audit reports them as undecided at this design's power rather than failed (Figure~\ref{fig:main-audit-gap}). The GSM8K row in Table~\ref{tab:survivors} is there for contrast and receives NO FLAG for the opposite reason: its off-span proxy largely coincides with the construct ($\kappa(z^{c},y){=}0.667$) and the disjoint gap is negative.

In the two cases whose construct AUC the equivalence test shows at chance while surface ranking survives --- the 50-character HotpotQA span, and OR-Bench's withheld pattern --- the surviving association is with refusal vocabulary or answer-string presence rather than with the target construct; Section~\ref{sec:repair} draws the practical consequence.

\paragraph{The motivating refusal contract.} Read on the continuous quantities first: the recorded judge labels give a residual of $+0.101$ after the disjoint control, above the contract's own null ($p_{95}{=}+0.061$, permutation $p{=}0.011$) but with a bootstrap interval, $[-0.059,+0.255]$, that covers zero. Among the rows above their null it is the only one whose score ranks the construct above chance. The contract is therefore statistically ambiguous. The two analyses answer different questions: the paired bootstrap asks how uncertain $\Delta_{\mathrm{dis}}$ itself is, and its interval covers zero; the permutation test asks whether the score's association with the off-span proxy exceeds what the construct explains, and it does. The verdict follows the permutation criterion, so the row is CAUTION; under the bootstrap criterion it would be CONTAINMENT (Appendix~\ref{app:null} states both null hypotheses), and the constant it is compared against is $0.10$. Both analyses support the narrower conclusion we draw: the reported $+0.393$ was not recovered under the disjoint control, and the estimated residual gap is $0.101$ with the interval above. The verdict summarizes the continuous metrics (Section~\ref{sec:verdict}); moving the CAUTION constant anywhere in $[0.02,0.10]$ leaves this row CAUTION and above $0.101$ makes it NO FLAG, and no other row's verdict depends on the constant (Appendix~\ref{app:cutoff}).

\label{sec:label-reliability}

In the motivating refusal contract the construct label is the weakest field; the question is how far the verdict depends on it. In a blind full-output check with two annotators and adjudication, resolved side labels agree with the judge at $\kappa{=}0.930$ and $0.715$ while pair-level disagreement agrees at $\kappa{=}0.266$, and the judge over-calls the pair-level event, firing on 15.6\% of resolved prompts against human adjudication's 2.8\%. Down-sampling judge positives to the rate human adjudication implies leaves the residual near its recorded value ($+0.106$ on average beside the recorded $+0.101$), but re-running the whole procedure on each draw returns UNDECIDABLE on 75.0\% of draws and CAUTION on 0.5\%: with 13 positives the contract is underpowered rather than re-classified. The sampling protocol, the threshold-crossing rates and the directional extremes are in Appendices~\ref{app:reliability-details} and~\ref{app:label-noise}.

\subsection{The template mechanism in the refusal contracts}
\label{sec:dissection}

The verdicts above say that an association survives the off-span re-read on some contracts and not others; they do not say what the shared evidence is. Identifying it requires contracts for which we control every field. We dissect two: OR-Bench, whose verdict the resolution precondition withholds but whose construct AUC is at chance by the equivalence test, so it gives the cleaner probe of the mechanism, and our own CAUTION case, in which the construct AUC remains above chance. The span at which both are scored instantiates partial-output scoring, as used directly in process reward model (PRM) guided search and in cascade settings whose routing decision is made from incomplete generated text (Section~\ref{sec:findings}). The score family, however, is not: no deployed system routes on a 50-character TF-IDF distance, and we use it only as a diagnostic instrument for isolating the shared-span mechanism, not as a deployable router (Appendices~\ref{app:strawman}, \ref{app:agcr}). The span effect is not an artifact of that instrument: replacing the lexical score with pretrained sentence encoders on the same contracts leaves the surface--construct split in place, although the XSTest categorical verdict changes from CAUTION to NO FLAG, its residual falling below the band while still exceeding its null (Table~\ref{tab:encoder-contracts}). Nor does shared-span exposure by itself produce a failure, since the same-span correctness contract of Appendix~\ref{app:frugal} receives NO FLAG on the check. Our controlled contract additionally places the cheap proxy on that same partial span; its fields are those of Table~\ref{tab:contract-instantiation}: Qwen3.5-2B \citep{qwenteam2026qwen35} and Gemma-4-E2B-it \citep{gemmateam2026gemma4} on XSTest \citep{rottger2024xstest}, a prefix TF-IDF distance computed with scikit-learn's default vectoriser \citep{pedregosa2011scikit}, a simple same-prefix keyword proxy (Appendix~\ref{app:strawman} compares it with a trained refusal classifier) and independent judge adjudication (reliability in Appendices~\ref{app:reliability-details}, \ref{app:inter-auditor}). Three runs recur: the \emph{primary run} diagnosed in Section~\ref{sec:generality}, a \emph{clean run} regenerating the same prompts under a protocol that lets a repair be tested, and a \emph{pair panel} across model pairs. The budget is $B{=}55$ on the first two, the top $10\%$ of the 548-prompt clean run, held fixed on the 450-prompt primary run, where it is the top $12\%$; the panel uses its own top $10\%$, $B{=}45$ (Appendix~\ref{app:budget-ci} sweeps it).

\label{sec:failure}

The score reaches AUC 0.985 against the proxy and 0.592 against the construct (primary run), so the proxy-only evaluation substantially overstates how strongly the same ranking tracks the declared construct. The
labels are nearly independent, $\kappa(z,y){=}0.024$ against a ceiling of $0.414$. Over the complete
output the prefix score still reaches $AUC(s,z^{+}){=}0.855$, and read strictly off the scored span the residual is the $+0.101$ of Section~\ref{sec:scorefailure}.

\label{sec:mechanism}The shared evidence is localized. All 13 prefix-keyword positives sit at the score's maximal value ($s{=}1.0$; the score itself takes 43 distinct values with modal mass $0.436$, so the sample stays rankable and the resolution precondition is not at issue), yet 11 of them have $y{=}0$: the proxy is a \emph{disagreement} rule, so these are proxy positives on pairs the judge calls agreed. Consistent with containment, they sit where the score is most extreme, because the same opening that trips the keyword rule also drives the prefix distance to its maximum. The concentration is consistent with a shared-template explanation but does not by itself distinguish it from label error tied to the same openings; the intervention below is what tests it. 

The intervention is a feature ablation on the realized generations, not a regeneration: we suppress those side-specific opening families with labels held fixed and re-score the edited text, comparing against deleting the same number of characters from the same responses at a random offset, over twenty seeds. OR-Bench provides the cleaner intervention setting. Its verdict is withheld and its mechanism is still readable, because the categorical verdict and the intervention answer different questions: the resolution precondition withholds a categorical verdict because the score does not order enough items uniquely, while the intervention is a paired comparison on the same items, measuring how the score's associations move when one text feature is removed. We use OR-Bench to localize a mechanism, not to assign a contract verdict. On OR-Bench hard 1k, scored by the same prefix-50 distance and with 473 construct positives, suppression lowers the score's agreement with the same-span keyword rule from $0.866$ to $0.596$ and with the off-span rule from $0.765$ to $0.447$, while the volume-matched deletion costs $0.042$ and $0.024$. The construct AUC, already at chance, changes negligibly ($0.517\to0.521$). The opening templates account for most of the score's association with refusal vocabulary, both within and beyond the scored span, while the construct AUC stays at chance. The intervention therefore isolates a surface association directly: it can be removed while the measured construct AUC stays where it was (Appendix~\ref{app:added-controls}). 

The motivating contract corroborates that mechanism rather than standing on its own. Suppression lowers its same-span keyword AUC from $0.985$ to $0.645$ against a drop of $0.019$ (sd $0.007$) under matched deletion, so the shared-span association is concentrated in the tested opening templates; but the intervention is not construct-neutral there. The drops are $0.285$ for the full-output keyword reading against $0.164$ for the construct, whose AUC falls to $0.428$ $[0.348,0.507]$, an interval that covers $0.5$ without establishing chance, and the comparison has a ceiling on one side, since the construct AUC had only $0.092$ above chance to lose and the intervention took all of it. The full three-target comparison, and the whole-family variant whose effect is a volume artifact, are in Appendix~\ref{app:added-controls}.

Run on the 50-character HotpotQA span, the same recipe finds a weaker and different mechanism. Suppressing each side's fixed opener (Qwen's visible ``thinking'' header and ``the user wants to'' phrase, Gemma's ``based on the provided passages'' family; 1{,}999 and 398 sides, 1.2\% of characters) lowers the same-span proxy AUC from $0.697$ to $0.666$ (paired change $-0.031$ $[-0.059,-0.002]$) and the off-span proxy AUC from $0.634$ to $0.604$ ($-0.030$ $[-0.051,-0.008]$). Matched random deletion leaves the first unchanged ($+0.000$ $[-0.003,+0.003]$) and moves the second by $-0.004$ $[-0.008,-0.001]$, an interval that excludes zero but is an order of magnitude smaller than the suppression's. The construct AUC's change, $+0.021$ $[-0.006,+0.047]$, is not distinguishable from zero, but on the stripped score its 90\% interval is $[0.508,0.560]$, outside the equivalence band, so that score is not certified at chance either. $\Delta_{\mathrm{dis}}$ falls from $+0.121$ to $+0.070$. The tested opener families therefore account for some of the off-span association but not all of it. A targeted lexical-overlap control does not explain the remainder either; its source remains unidentified (Appendix~\ref{app:added-controls}).

\section{Released Artifacts}
\label{sec:findings}

\begin{table}[!ht]
\centering\footnotesize
\setlength{\tabcolsep}{4pt}
\begin{tabular}{p{0.19\linewidth}p{0.20\linewidth}p{0.17\linewidth}p{0.17\linewidth}p{0.20\linewidth}}
\toprule
Released artifact & Contract & Off-span control & Algorithm~\ref{alg:rlcaudit} verdict & Same-span check (not a verdict) \\
\midrule
PRM800K & prefix score, five proxy rules & yes, released score & NO FLAG (four instantiable rules; one not instantiable) & --- \\
Math-Shepherd & rollout step labels vs.\ final correctness & yes & NO FLAG & provenance: consumption mismatch as step correctness (Appendix~\ref{app:prmcontainment}) \\
RouteLLM (gpt4\_dataset) & prefix distance, judge target & yes; complement empty on 16.9\% of sides & UNDECIDABLE (complement)$^{\ast}$ & same-span gap negative \\
\midrule
PRM800K & full-solution best-of-$n$ & no: score reads the complete solution & UNDECIDABLE (field missing) & within-problem AUC 0.854 at full span; no flag \\
Search-and-learn & partial-chain beam pruning & no: pruned beams not retained & UNDECIDABLE (field missing) & released metrics only \\
RouterBench, MMLU prof.\ law & response distance, 10\% budget & no: stored complete responses & UNDECIDABLE (field missing) & label agreement $\kappa{=}0.728$; no flag \\
RouterBench, MT-Bench & same & no & UNDECIDABLE (field missing) & ranking at chance against both label sets \\
RouterBench, GSM8K slice & same & no & UNDECIDABLE (not instantiable) & label cannot equal the stored response's correctness \\
Hybrid LLM / MixInstruct & BARTScore quality gap & no: complete responses & UNDECIDABLE (field missing) & label agreement $\kappa{=}0.426$, queue disagreement 13.8\%; no flag \\
UltraFeedback & preference selection target & no: complete completions & UNDECIDABLE (field missing) & label agreement $\kappa{=}0.631$; no flag \\
FrugalGPT-style$^{\dagger}$ & answer-reading scorer, our traces & no: scorer reads the whole answer & UNDECIDABLE (field missing) & paired gaps cover zero; no flag \\
\bottomrule
\end{tabular}
\caption{Every external contract, split by what the released files support: whether the off-span control can be run, the verdict Algorithm~\ref{alg:rlcaudit} returns, and the same-span check reported where it cannot run, which is not a verdict of the audit. Eleven contracts over eight released artifacts and one published recipe; full audits in Appendix~\ref{app:external}. $^{\ast}$RouteLLM's complement is empty on $16.9\%$ of sides, above the $15\%$ precondition, so the contract is undecidable from the released files (Appendix~\ref{app:routellm}). $^{\dagger}$The FrugalGPT-style row is instantiated on our own traces, since the released system publishes no per-example generations (Appendix~\ref{app:frugal}).}
\label{tab:verdict-tally}
\end{table}

The audit so far ran on contracts whose fields we control. The external cases answer two questions: whether the full off-span audit can be instantiated from released artifacts at all, and, where it cannot, what weaker same-span check the release still supports. The audit is applicable only where per-example generations, the scored span and a construct label are jointly released. For released artifacts, an UNDECIDABLE outcome often identifies missing audit fields rather than a property of the score. The audit reads only published files, so it returns verdicts on systems we did not build; the contract's fields say which files to ask for. Table~\ref{tab:verdict-tally} separates what each release supports: three contracts support the off-span control (two receive no coupling flag; RouteLLM is withheld by the complement precondition), and the other eight admit no off-span re-read on the released files, because the score reads the complete output or the pruned candidates are not retained. Under Algorithm~\ref{alg:rlcaudit} those eight are UNDECIDABLE with a required field missing, and the table reports for them only a same-span label-agreement check, which is not a verdict of the audit and whose absence of a flag is not endorsement (Section~\ref{sec:noflag}); Section~\ref{sec:partial-span} walks through the individual releases, and Appendix~\ref{app:external} audits each in full.

\subsection{Partial-span selection}
\label{sec:partial-span}

Test-time search steers generation with a process reward model scoring an unfinished chain. PRM800K
\citep{lightman2024verify} answers on released data whether that score is evidence about final-answer correctness: over 815{,}632 scored solutions, within-problem AUC rises monotonically with
the fraction of the solution scored, from 0.529 at the first step to 0.854 over the whole solution (Appendix Figure~\ref{fig:deployed-spans} plots this curve against the search pipelines').
Search-and-learn \citep{beeching2024scaling} decides at the left-hand end of that span curve, scoring partial chains of
Llama-3.2-1B \citep{dubey2024llama} solutions to MATH-500 \citep{hendrycks2021math}. Its released metrics
put the PRM near chance exactly where pruning acts (within-problem $AUC{=}0.445$ at the first scored step),
but the surviving beams were selected by the score under evaluation and the pruned ones are not retained, so
the contract is \emph{not decidable from the released files} and that reading is a released metric, not a
verdict. It is an instance of the release practice Section~\ref{sec:release} measures (Appendix~\ref{app:prmsearch}).

PRM800K's own dumps are not conditioned that way. On 300 problems and 493{,}030 sampled solutions --- a smaller cut of the same dump than the 500-problem, 815{,}632-solution curve above, set by the control's per-span recomputation cost --- with the proxy a cheap on-track rule, the orientation-robust gap under the disjoint control is negative
at both a quarter and a half of the solution, $-0.032$ $[-0.055,-0.008]$ and $-0.051$
$[-0.077,-0.024]$. Holding score and construct fixed and varying the proxy across five rules a practitioner might reach for leaves every instantiable verdict unchanged (one rule, a committed \verb|\boxed| answer, fires on no prefix at that span and is not instantiable), so the reading is unchanged across the tested instantiable rules, which is robustness within that family rather than independence of proxy choice in general.

\paragraph{Further released artifacts.}
\label{sec:routerbench}

PRM800K and Math-Shepherd are the two contracts that take no coupling flag under the complete off-span control, and RouteLLM \citep{ong2025routellm} is the one the complement precondition withholds. The remaining external contracts cannot instantiate the control at all, because they store complete outputs or omit the pruned candidates: RouterBench \citep{hu2024routerbench} on MMLU professional law \citep{hendrycks2021mmlu} and MT-Bench \citep{zheng2023judging} (Appendix~\ref{app:routerbench}), Hybrid LLM \citep{ding2024hybrid,jiang2023llmblender}, UltraFeedback (Appendix~\ref{app:ultrafeedback}), and FrugalGPT \citep{chen2023frugalgpt} on our own judged HotpotQA \citep{yang2018hotpotqa} and GSM8K \citep{cobbe2021training} traces, none of which shows a flag on the same-span check that remains. One row is not instantiable in a stronger sense: RouterBench's grade-school-math slice records fractional exact-match labels over a single stored response, so $99.2\%$ of them cannot equal that response's binary correctness.

\subsection{Auditability of the literature}
\label{sec:release}

The audit can be applied only where the required per-example artifacts are released. From a screened pool of routing, cascade and deferral papers we drew a fixed-seed
random sample of 40, read all of them, and coded two release conditions: whether the per-example generations are released (R1), and whether the proxy rule is reproducible from the text (R2); Table~\ref{tab:survey-flow} records the count at each screening stage. Of 38 in scope, none is found to satisfy R1 from the paper and its linked materials (95\% CI $[0.0,9.2]$) and 17 satisfy R2, so their conjunction is also 0. The artifacts audited in Section~\ref{sec:findings} are not counterexamples, because they were not drawn from this sample: they are reward datasets, benchmarks and routing releases reached by targeted search rather than by the random draw, and where one releases generations we could recompute a proxy rule over a span (PRM800K, RouteLLM). What the sample bounds is the routing literature's own release practice. A third coding bounds which papers the failure can apply to: 19 of 38 score the query alone, 13 read generated text, and 6 do not say, so between $34\%$ and $50\%$ of the sample can exhibit containment at all; within the 13, the only papers the audit could apply to, no release was identified either. The same survey codes how
often a paper checks its own label against the construct it claims --- 3 of 38, $7.9\%$ with CI
$[2.7,20.8]$, a reporting rate --- how often the check would fail is unmeasured (Appendix~\ref{app:prevalence}).

\begin{table}[!htbp]
\centering
\footnotesize
\begin{tabular}{lr}
\toprule
Stage & Papers \\
\midrule
Six fixed arXiv queries (routing, cascades, deferral, selective prediction; 2023--) & 178 \\
Abstract screen: makes a per-query decision and evaluates it against a label & 125 \\
Fixed-seed random sample, retrieved in full & 40 \\
In scope after full-text reading & 38 \\
\quad release per-example generations (R1) & 0 \\
\quad proxy rule reproducible from the text (R2) & 17 \\
\quad check their own label against the construct & 3 \\
\bottomrule
\end{tabular}
\caption{Flow of the release survey.}
\label{tab:survey-flow}
\end{table}

\section{Implications}
\label{sec:repair}

\paragraph{Implications for the motivating routing configuration.} A flagged contract can motivate a revised instantiation, while an undecidable contract instead identifies the additional artifact or statistical power needed for a defensible judgment. Relocating the span and adding a construct-bearing cue are the follow-up diagnostics the audit points to; Appendix~\ref{app:repair} illustrates both on the motivating setup, with the routed queue in Figure~\ref{fig:contract-overview}. We do not claim a routing method: the semantic gain's interval covers zero, and final-span TF-IDF is not evaluated as deployable.

\paragraph{Recommended follow-up analyses.} Table~\ref{tab:exits} lists the follow-up each pattern motivates; none is a routing method. No demonstrated construct ranking means the score is not construct evidence at this span; a containment-compatible pattern calls for relocating or intervening on the span, as above; a residual divergence calls for examining both the label and off-span spillover, since the pattern does not choose between them; an undecidable contract calls for the missing artifact or the missing power. The distinction matters in practice: without the off-span measurement, a no-ranking row reads as a proxy-label failure, and the prescription would be to repair the label --- which at scale means substituting another cheap proxy, since construct-level annotation is precisely what the cheap proxy exists to avoid. On the evidence of Section~\ref{sec:cases}, that repair could leave the score validated against the same surface feature it already tracks.

\paragraph{Absence of a flag is not endorsement.}
\label{sec:noflag}
The audit only ever flags failures, so a clean check licenses no positive claim. PRM800K's full-solution best-of-$n$ contract shows no flag under the same-span check, and does not support the off-span control, while showing the worst selection harm we measured: best-of-$n$ picks a wrong completion on $19.2\%$ of PRM800K problems whose pool holds a correct one, and on $51.5\%$ in the search-and-learn release, whose contract is undecidable as above (Appendix~\ref{app:prmsearch}).

\paragraph{Implications for reporting practice.} The auditability results of Section~\ref{sec:release} turn into a concrete reporting request. A proxy-based validation claim should declare the span the score reads and the span the proxy is computed from, report the construct AUC beside the proxy AUC at the stated selection budget, and release the per-example generations that let the same proxy rule be re-read off the scored span. None of the 38 sampled routing papers currently supports that re-read, so in practice the control is available to authors auditing their own systems, and to third parties only where a release includes the generations. Appendix~\ref{app:reporting-guidance} tabulates the minimum report for each audit component; the span fields cost a sentence to declare, and they are the precondition for every other row.

\section{Limitations}
\label{sec:discussion}

\paragraph{Identification.}
The exits are evidence patterns. The off-span re-read removes what a shared span can supply, but a gap that survives it can be a proxy--construct mismatch or prefix-to-continuation dependence that bypasses the construct, and the permutation null removes both, so DIVERGENCE does not identify a cause and licenses no unique repair (Sections~\ref{sec:limits}, \ref{sec:repair}). 

\paragraph{Statistics.}
\label{sec:limits2}
The categorical summaries depend on the constants in Appendix Table~\ref{tab:constants}; the continuous estimates are the primary results. On structured synthetic cases the rule abstains on much of what it should flag at $n{=}450$ and returns NO FLAG on 78\% of replicates of a case that mixes containment with mismatch even at $n{=}2000$ (Appendix~\ref{app:error-rates}). The permutation null and the paired bootstrap interval test different hypotheses (Appendix~\ref{app:null}); they disagree on five of eleven rows and move one verdict, our own: CAUTION under the null, CONTAINMENT under the interval, and both are reported. The motivating contract is not precise enough to separate a $0.15$ residual gap from zero under the paired-bootstrap reading, its estimated requirement being 51 construct positives against the 46 it has; the reported CAUTION follows the contract-specific permutation criterion instead, and both readings are reported. JailbreakBench and the 80-character HotpotQA span are underpowered enough that the equivalence step leaves them undecided. Five rules were added after the results were seen. Each addressed a specific failure of the instrument: a score constant on most of a span, a construct AUC below chance, a construct with three positives, a partial control that leaves shared evidence in, and a null that generated text violates. All five were then applied uniformly to every contract, and Table~\ref{tab:recount} reports the resulting sensitivity. Appendix~\ref{app:preregistration} reports what the frozen procedure returns on a held-out contract that informed none of these rules: it abstains at the equivalence step on a contract whose other three quantities have the no-ranking shape. Table~\ref{tab:prereg} records them with their cost, and Table~\ref{tab:recount} recounts the twenty contracts without the three preconditions; the resolution precondition, run on every contract, withholds OR-Bench, which would otherwise be a second NO DEMONSTRATED CONSTRUCT RANKING exit.

\paragraph{Construct labels.}
Each construct label is derived from one greedy generation per side, adjudicated by a judge that over-calls pair-level disagreement against human adjudication ($15.6\%$ against $2.8\%$ on 109 resolved prompts), with pair-level agreement of $\kappa{=}0.266$ on a sample enriched toward high-scoring prompts. Down-sampling the judge's positives to the human rate and re-running the procedure returns UNDECIDABLE on three quarters of draws and CAUTION on almost none (Section~\ref{sec:label-reliability}), so the CAUTION is a reading of the recorded labels, not a robust one; what several sampled draws per side would return is unmeasured (Appendix~\ref{app:singledraw}).

\paragraph{Generalization.}
The twenty controlled contracts are seven refusal settings and thirteen span variants of two correctness tasks on one model pair, not a sample of systems, and no deployed router is audited end to end; a ten-pair final-span panel (Appendix~\ref{app:pair-panel}), two correctness-routing contracts (Appendix~\ref{app:noncorrectness}) and the calibration and boundary controls (Appendix~\ref{app:calibration-controls}) bound how far the primary pair generalizes without extending the sample. Released artifacts containing the required fields are atypical: no release of per-example generations could be identified for any of 38 surveyed papers, so the absence of a statistical failure in the wild is bounded by where the audit could look rather than evidence that the failure is rare. DIVERGENCE and CONTAINMENT are not reached on this corpus under the reported rule, but both are reachable. On structured synthetic cases the rule returns CONTAINMENT on 93\% and DIVERGENCE on 90--100\% of replicates at $n{=}2000$ (Table~\ref{tab:error-rates}), and on this corpus the motivating contract exits as CONTAINMENT under the bootstrap reading, while OR-Bench would be a second NO DEMONSTRATED CONSTRUCT RANKING but for the resolution precondition (Table~\ref{tab:recount}).

\section{Conclusion}

Agreement between a score and a proxy measured on the same span of text does not by itself certify the construct the proxy stands for; the evidence span can be declared as a field of the validation contract, and the proxy then re-evaluated on evidence disjoint from the score's. Re-evaluating the proxy off the scored span turns a reported agreement into three continuous quantities with a contract-specific null, directionally validated on synthetic contracts with known containment and corroborated by targeted interventions on real text. In the one controlled contract meeting the no-ranking criterion, the score ranks an off-span surface proxy while failing to rank the construct. The appropriate follow-up is to stop treating the score at that span as reliable evidence for the adjudicated construct; replacing the proxy label alone does not address the demonstrated score--construct mismatch (Appendix~\ref{app:added-controls}). One further contract is CAUTION on the recorded judge labels, a reading that does not survive a label-sensitivity analysis, and the remaining contracts either show no residual discrepancy under the control or remain undecidable. Most of the released artifacts we examined cannot be audited this way at all, which is a finding about the surveyed reporting practice rather than about any one system. The report that would let a reader check any of it is short: both spans named, the construct agreement printed beside the proxy agreement, and the generations released.

\section*{Reproducibility Statement}

Every number in this paper is produced by a script from released or
packaged artifacts, and each result section names the script that produces it. The contract fields for
every audited configuration, the run map that distinguishes the primary, clean and panel runs, the
judge validation against human labels, and the packaged run identifiers are given in
Appendix~\ref{app:reproducibility}. The verdict procedure, including which branches are auditor
judgments and which rules were fixed before the results were seen, is stated in
Appendices~\ref{app:algorithm} and~\ref{app:preregistration}. Code and analysis outputs are released at \url{https://github.com/wdi1024/rlc-audit}. The released JSON outputs use earlier exit names: \texttt{ALIGNED} for NO FLAG and \texttt{SCORE\_FAILURE} for NO DEMONSTRATED CONSTRUCT RANKING.

\bibliographystyle{tmlr}

\bibliography{acl_arr_refs}

\appendix
\section{Containment: Proofs}
\label{app:proposition}

Proposition~\ref{prop:containment} is elementary, and we state it in full because the paper's central
claim --- that a proxy read from the scored span certifies a score for reasons that have nothing to do
with the construct --- is otherwise carried by prose alone.

\begin{proposition}\label{prop:containment}
Let $s = g(x_{<b})$, suppose $x_{<b}$ is independent of $x_{\ge b}$, and let every label below take each of its two values with positive probability. Then \emph{(i)} for a
span-measurable proxy $z = \mathbf{1}\{f(x_{<b}) \ge \tau\}$, $AUC(s,z)$ does not depend on that
independence and equals $1$ whenever $g$ is a strictly increasing transform of $f$; while \emph{(ii)}
for a disjoint proxy $z^{c} = h(x_{\ge b})$, $AUC(s,z^{c}) = \tfrac12$ necessarily.
\end{proposition}

\paragraph{Setup.} Fix a contract. Let $x$ be a model output, $b$ the budget in characters, and
$x_{<b}$ the prefix the score reads. The score is $s = g(x_{<b})$ for a measurable $g$, and the proxy
is $z = \mathbf{1}\{f(x_{<b}) \ge \tau\}$ for a measurable $f$ and threshold $\tau$. Write
$P_1$ for the law of $g(x_{<b})$ conditioned on $f(x_{<b}) \ge \tau$ and $P_0$ for the law conditioned
on $f(x_{<b}) < \tau$. With ties broken at one half,
\[
AUC(s,z) \;=\; \Pr\!\big[S_1 > S_0\big] + \tfrac{1}{2}\Pr\!\big[S_1 = S_0\big],
\qquad S_1 \sim P_1,\; S_0 \sim P_0 \text{ independent.}
\]

\paragraph{Claim 1 (a same-span agreement survives independence).} Suppose $x_{<b}$ is independent
of $x_{\ge b}$. Both $P_1$ and $P_0$ are pushforwards of the law of $x_{<b}$ under maps reading only
$x_{<b}$, so neither is touched by that assumption and $AUC(s,z)$ is whatever the pair of rules makes
it. If $g = \phi \circ f$ with $\phi$ strictly increasing on the support, $f(x_{<b}) \ge \tau$ implies
$g(x_{<b}) \ge \phi(\tau)$ and conversely, so $P_1$ sits on $[\phi(\tau),\infty)$, $P_0$ on
$(-\infty,\phi(\tau))$, every draw satisfies $S_1 > S_0$, and $AUC(s,z) = 1$. A same-span agreement can
therefore be perfect in a world where the scored span predicts nothing outside itself.

\paragraph{Claim 2 (a disjoint agreement cannot).} Let $z^{c} = h(x_{\ge b})$ under the same
independence. Then $s = g(x_{<b})$ and $z^{c}$ are independent, so the conditional law of $s$ given
$z^{c}=1$ equals its law given $z^{c}=0$; drawing $S_1$ and $S_0$ from the same distribution gives
$\Pr[S_1 > S_0] = \Pr[S_0 > S_1]$ and hence $AUC(s,z^{c}) = \tfrac12$ exactly. Any excess over chance
in the disjoint control is evidence that the scored span really does predict text beyond it, which is the
one reading the same-span proxy AUC cannot supply.

\paragraph{Claim 3 (what is \emph{not} special about containment).} It is tempting to argue instead
that a same-span agreement is uninformative because it fixes nothing about $AUC(s,y)$. That is true but proves too much, so the paper does not rest on it. Fix the contract.
If there is a $U$ measurable with respect to $x$, uniform and independent of $x_{<b}$, then any
function of $(s,U)$ is an admissible construct: $y_1 = \mathbf{1}\{s \ge m\}$ at a median $m$ gives
$AUC(s,y_1)=1$, $y_0 = \mathbf{1}\{U \ge \tfrac12\}$ gives $\tfrac12$, and mixing between them sweeps
the interval. But this argument never consults where $z$ was read: it applies verbatim to a disjoint
proxy. Lack of constraint on the construct is a generic property of proxy AUCs, not a property of
containment, and it does not by itself license re-reading the proxy off-span. Claims 1 and 2 do.
Two caveats: the mixing construction needs an atomless component, which greedy decoding over a finite
prompt set need not provide, and on a finite sample the swept set is finite rather than all of
$[0,1]$.

\paragraph{Claim 4 (when containment is legitimate).} Suppose the construct is also span-measurable,
$y = h(x_{<b})$. Then both terms of $\Delta_{\mathrm{AUC}}$ are fixed by the span, and a small gap is consistent with the proxy rule and the construct rule agreeing about the same evidence; it is not by itself proof of agreement, since different labels can share an AUC against one score. Nothing is wrong
with such a contract, and Appendix~\ref{app:frugal} audits one that takes no coupling flag. Representation--label
coupling is the asymmetric case, $z$ span-measurable and $y$ not, and that asymmetry is what makes
containment possible: Claim 1 shows that a large same-span agreement \emph{can} arise from shared-span
structure alone while the second term cannot be produced the same way. It does not imply that the first
term must be inflated in a given contract, nor that the gap must be large; whether either happens is
what the off-span control measures. Sharing a span is necessary for containment, not sufficient
(Appendix~\ref{app:frugal}).

The extended proxy $z^{+}$ is measurable with respect
to $\sigma(x) \supseteq \sigma(x_{<b})$, so Claim 1 still applies to the part of $z^{+}$ determined by
the prefix; $\Delta_{\mathrm{ext}}$ therefore reduces containment without eliminating it. The disjoint
proxy $z^{c}$ is measurable with respect to $\sigma(x_{\ge b})$, which is the complement, and Claim 1
does not apply. This is the formal content of the distinction drawn in Section~\ref{sec:contract} and
the reason the two controls disagree in Section~\ref{sec:split}.

\section{Calibration Controls}
\label{app:calibration-controls}
This appendix collects the calibration and boundary controls referenced from the main text: Table~\ref{tab:rlc-controls} reports prevalence and calibration checks for the headline diagnostic, Table~\ref{tab:boundary-controls} the span controls, and Table~\ref{tab:model-pair-boundary} the full model-pair and span rows, of which Section~\ref{sec:boundary} quotes a subset.

\begin{table}[!htbp]
\centering
\small
\resizebox{\linewidth}{!}{%
\begin{tabular}{llrrrrl}
\toprule
Control score on XSTest 450 & $\Delta_{\mathrm{AUC}}$ band & AUC $z$ & AUC $y$ & AP $z$ & AP $y$ & Role \\
\midrule
Raw prefix-50 TF-IDF & $\geq 0.15^{\dagger}$ & 0.985 & 0.592 & 0.500 & 0.127 & Audited score \\
Surface-label oracle & $\geq 0.15$ & 1.000 & 0.508 & 1.000 & 0.104 & Coupled control \\
Semantic-label oracle & negative gap & 0.527 & 1.000 & 0.031 & 1.000 & Semantic control \\
Judge-ensemble disagreement score & negative gap & 0.599 & 0.730 & 0.088 & 0.264 & Non-oracle usage example \\
\bottomrule
\end{tabular}%
}
\caption{Calibration controls for the RLC-Audit diagnostic on XSTest 450; bands are on $\Delta_{\mathrm{AUC}}$, the superseded rule, and the oracle rows are constructed, not audited contracts. These controls check the direction of the diagnostic bins, not only the audited router.}
\label{tab:rlc-controls}
\end{table}

$^{\dagger}$ These bands are computed from $\Delta_{\mathrm{AUC}}$, the quantity a standard report shows. Under the control that decides (Appendix~\ref{app:added-controls}) the motivating contract is
CAUTION; we print both because the difference between them is the paper's point.

The final row is a methodological calibration control. Its role is to show the direction of the
diagnostic: scores aligned with the semantic construct can receive NO FLAG, while scores aligned mainly with the surface proxy receive DIVERGENCE. A no-residual-flag exit narrows the coupling risk for that score; it does not certify routing utility.

The diagnostic status is insensitive to reasonable stricter cutoffs in the main audit. Requiring both AUC gap $\ge 0.30$ and $\kappa(z,y) \le 0.10$ still places the five primary and diagnostic ranking rows above the $0.15$ band under $\Delta_{\mathrm{AUC}}$ (superseded for the primary row by Appendix~\ref{app:added-controls}), because their AUC gaps are +0.310 to +0.433 and their per-model keyword-vs-judge kappas (side-level means) are at most 0.048; the motivating contract's pair-level $\kappa(z,y)$ is 0.024. The reported continuous AUC/AP/kappa values, not the categorical label alone, should be used when comparing audits. Table~\ref{tab:boundary-controls} collects the full set of validity and boundary controls with their outcomes.

\begin{table}[!htbp]
\centering
\scriptsize
\begin{tabular}{>{\raggedright\arraybackslash}p{0.20\linewidth}>{\raggedright\arraybackslash}p{0.34\linewidth}>{\raggedright\arraybackslash}p{0.34\linewidth}}
\toprule
Concern & Control & Outcome \\
\midrule
Lexical TF-IDF shortcut & Prefix MiniLM-L6-v2/e5-small-v2 & Surface/semantic AUC remains split, 0.943/0.610 and 0.962/0.602 through the full rule (Table~\ref{tab:encoder-contracts}). \\
Batch fitting artifact & HashingVectorizer and leave-one-setting-out TF-IDF & The split persists without fitting on the target batch. \\
Span mismatch & 512-token span and word-token prefixes & Matched prefix keyword stays high (0.991); prefix-vs-full keyword falls to 0.476; word-token-8 is 0.983/0.591. \\
Model-pair artifact & Same-family Qwen8B/Qwen9B; Qwen9B/Llama8B & Same-family has no prefix positives; 8B/9B prefix-50 clears (0.444/0.568), while raw full has a $\geq 0.15$ gap: 0.777/0.383 AUC, kappa -0.048, and 0/55 semantic disagreements in the top queue. \\
Statistical fluke & Primary/diagnostic ranking rows plus sparse-positive operating-point rows & Primary gap CI $[0.324,0.464]$, permutation $p<10^{-4}$ (Appendix~\ref{app:budget-ci}). A second refusal setting reproduces the split: JailbreakBench \citep{chao2024jailbreakbench}, a jailbreak benchmark whose released label is an LLM judge (Appendix Table~\ref{tab:prevalence-safety}), gives 0.818 against keywords and 0.508 against judges with 27 semantic positives. \\
\bottomrule
\end{tabular}
\caption{Validity and boundary controls. Each row targets a different explanation for the raw-prefix failure.}
\label{tab:boundary-controls}
\end{table}

\begin{table}[!htbp]
\centering
\scriptsize
\setlength{\tabcolsep}{4pt}
\resizebox{\linewidth}{!}{%
\begin{tabular}{lrrrrlccl}
\toprule
Contract & AUC $z$ & AUC $y$ & AUC $z^{c}$ & $\Delta_{\mathrm{AUC}}$ & $\Delta_{\mathrm{dis}}$ [95\% CI] & null $p_{95}$ / $p$ & modal & verdict \\
\midrule
XSTest 450 / MiniLM-L6-v2 & 0.943 & 0.610 & 0.673 & +0.333 & +0.063 $[-0.110,+0.259]$ & +0.054 / 0.038 & 0.280 & NO FLAG \\
XSTest 450 / e5-small-v2 & 0.962 & 0.602 & 0.692 & +0.360 & +0.090 $[-0.070,+0.257]$ & +0.058 / 0.017 & 0.436 & NO FLAG \\
OR-Bench hard 1k / MiniLM-L6-v2 & 0.884 & 0.531 & 0.781 & +0.353 & +0.250 $[+0.148,+0.339]$ & --- & 0.542 & UNDECIDABLE (resolution) \\
OR-Bench hard 1k / e5-small-v2 & 0.870 & 0.531 & 0.773 & +0.339 & +0.242 $[+0.148,+0.327]$ & --- & 0.644 & UNDECIDABLE (resolution) \\
\bottomrule
\end{tabular}%
}
\caption{Encoder-scored prefix-50 contracts through the full rule of Section~\ref{sec:verdict}, on the two settings with at least ten construct positives. The score is one minus the cosine between sentence embeddings of the two 50-character prefixes (all-MiniLM-L6-v2, \citealp{reimers2019sentencebert}; e5-small-v2 with its query prefix, \citealp{wang2022textembeddings}); proxy, construct and preconditions are those of the TF-IDF rows, which the script reproduces first. \emph{modal} is the share of pairs at the score's most common value. Generated by \texttt{\detokenize{scripts/encoder_contracts_full_rule.py}}.}
\label{tab:encoder-contracts}
\end{table}

The model-pair controls make the scope claim sharper. Existing packaged results support a pair/span-dependent RLC claim, not the stronger claim that every heterogeneous pair mismatches. The mismatch appears when the score span and proxy label share artifact-bearing evidence; the same pair can take no flag on one span contract and fail another. Token-matched 512-token cross-pair rows additionally show that the effect is strongest for Qwen/Gemma, weaker for Qwen/Llama, and still present for Gemma/Llama under the same cached prompt set. Because these rows are banded under the superseded rule rather than judged, no row in Table~\ref{tab:model-pair-boundary} contradicts the main-text tally.

\begin{table}[!htbp]
\centering
\scriptsize
\resizebox{\linewidth}{!}{%
\begin{tabular}{lllrrrll}
\toprule
Pair & Score span & $z$/$y$ n+ & AUC $z$ & AUC $y$ & Gap & $\Delta_{\mathrm{AUC}}$ band (superseded) & Role \\
\midrule
Qwen3.5-2B / Gemma-4-E2B-it & raw prefix-50 & 13/46 & 0.985 & 0.592 & +0.393 & $\geq 0.15$$^{\dagger}$ & primary \\
Qwen3.5-2B seed1 / seed2 & raw prefix-50 & 0/41 & -- & 0.527 & -- & -- & no proxy positives \\
Llama-3.2-1B / Qwen3-1.7B & raw prefix-50 & 147/77 & 0.530 & 0.571 & -0.041 & below $0.10$ & scope check \\
Qwen3.5-9B / Llama-3.1-8B & raw prefix-50 & 281/50 & 0.444 & 0.568 & -0.124 & at chance & scope check \\
Qwen3.5-9B / Llama-3.1-8B & raw full trace & 270/50 & 0.777 & 0.383 & +0.394 & $\geq 0.15$ & span control \\
Qwen3.5-2B / Gemma-4-E2B-it & raw prefix-50, 512-token matched & 15/44 & 0.991 & 0.567 & +0.424 & $\geq 0.15$ & token-matched \\
Qwen3.5-2B / Llama-3.2-3B & raw prefix-50, 512-token matched & 167/58 & 0.589 & 0.475 & +0.114 & $0.10$--$0.15$ & token-matched \\
Gemma-4-E2B-it / Llama-3.2-3B & raw prefix-50, 512-token matched & 154/52 & 0.655 & 0.486 & +0.169 & $\geq 0.15$ & token-matched \\
\bottomrule
\end{tabular}%
}
\caption{Model-pair and span boundary summary from packaged analyses. These rows sit outside the twenty audited contracts and they are placed in the magnitude bands of $\Delta_{\mathrm{AUC}}$, the superseded rule, rather than given exits of the final taxonomy. $^{\dagger}$ computed from
$\Delta_{\mathrm{AUC}}$; under the deciding control this row is CAUTION
(Appendix~\ref{app:added-controls}). Two Gemma models appear in this paper: Gemma-4-E2B-it is the dissected contract's model, while ``Gemma9B'' in these scope tables is
Gemma-2-9B-it. A dash means the proxy target has no positives, so no verdict is computable. The role
column records why each row was run. Panel models: Qwen3 \citep{yang2025qwen3}, Gemma \citep{gemmateam2024gemma}, Mistral \citep{jiang2023mistral}, Phi-3.5 \citep{abdin2024phi3} and Llama~3 \citep{dubey2024llama}.
\textbf{Caveat.} These verdicts are computed from point estimates: the packaged analyses do not carry
paired-bootstrap intervals. The \emph{at chance} mark is a point-estimate scope check (interval covering $0.5$), not the main pipeline's NO DEMONSTRATED CONSTRUCT RANKING verdict, which requires the equivalence band of Section~\ref{sec:verdict}. The distinction between the NO FLAG and at-chance prefix-50 scope rows (AUC $y$ of $0.571$ and $0.568$) is therefore not
established by this table, and we do not rest any claim on it; the rows are reported to show that the same metric family takes no flag on one contract and fails another as pair and span change, which the two extremes of the band column support on their own. Generated by
\texttt{\detokenize{scripts/analyze_model_pair_boundary.py}} and
\texttt{\detokenize{scripts/analyze_token_matched_prefix_crosspairs.py}}.}
\label{tab:model-pair-boundary}
\end{table}

\section{Multi-Pair Final-Span Panel}
\label{app:pair-panel}

The panel in Appendix Table~\ref{tab:pair-panel-full} audits one shared contract instantiation across model pairs. The contract fields are: score $s$ is final-span TF-IDF cosine distance on the parsed \texttt{<final>} channel of tagged \texttt{<reasoning>}/\texttt{<final>} generation (thinking disabled, greedy decoding, 768 max new tokens); proxy $z$ is same-span refusal-keyword disagreement; semantic $y$ is final-channel judged refusal-decision disagreement (OpenAI GPT-4o-mini final-channel adjudication); and the budget is the top-10\% queue ($B{=}45$ of 450). This is deliberately a different contract from the primary raw-prefix contract in Table~\ref{tab:contract-instantiation}: under the audit methodology, the generation protocol, score span, and judge are contract fields, so the panel rows are comparable to each other but are not re-runs of the motivating contract.

The panel judge differs from the Haiku judge. We treat this as a contract field and report it explicitly, so labels are never mixed across tables. As a cross-judge consistency check, we re-adjudicated the final channels of nine of the ten panel pairs with the Haiku judge model (Claude Haiku 4.5); the same-family control pair (Qwen8B/Qwen9B) was not re-adjudicated. Side labels are shared across pairs, so this needed one re-adjudication per model and phase rather than per pair. Pair-level semantic labels agree at 0.895 on average across the nine pairs ($\kappa$ 0.604 mean, 0.518--0.665), and the panel proxy--semantic gap spans $[-0.015,+0.128]$ under the panel judge against $[-0.043,+0.101]$ under the Haiku judge. No pair reaches the DIVERGENCE cutoff under either labeling, and the one pair whose gap changes sign (Qwen/Mistral, $+0.018$ versus $-0.043$) does so around zero, far from any verdict boundary. The separation from the motivating contract's gap ($+0.393$, CI $[0.324,0.464]$) therefore holds under both judges. Per-model refusal labels agree closely across judges: raw agreement 0.922 with $\kappa{=}0.843$ (Qwen3.5-9B), 0.918 with $\kappa{=}0.832$ (Gemma-2-9B-it), and 0.898 with $\kappa{=}0.782$ (Mistral-7B-Instruct-v0.3, \citealp{jiang2023mistral}), each over all 450 prompts. Pair-level semantic-disagreement labels, which compound noise from both sides, agree at 0.873--0.884 raw ($\kappa$ 0.56--0.65). Full per-pair numbers are in \texttt{\detokenize{analysis_results/cross_judge_panel_full.json}} (\texttt{\detokenize{scripts/cross_judge_panel_full.py}}). The panel's headline quantity is judge-robust: the final-span score's semantic AUC is 0.664 vs.\ 0.673 (Qwen/Gemma), 0.649 vs.\ 0.710 (Qwen/Mistral), and 0.707 vs.\ 0.710 (Gemma/Mistral) under the GPT-4o-mini and Claude labels respectively, so no panel row changes verdict when re-labeled by the Haiku judge.

Table~\ref{tab:pair-panel-full} reports the full panel including parse coverage. Five pairs involving Llama-3.1-8B are excluded from the main-text panel because the tagged parse succeeds exactly on only 8/450 generations, and 80 generations fall back to using the raw trace as the final span. Under the audit methodology this is not a data-cleaning preference: when the extracted span is not the final-answer span, the final-span score--proxy--construct contract is simply not instantiated, so reporting those rows as final-span evidence would itself be a span-contract violation. The excluded rows are retained here for completeness with their parse accounting.

\begin{table}[!htbp]
\centering
\scriptsize
\resizebox{\linewidth}{!}{%
\begin{tabular}{llrrrrrrl}
\toprule
Pair & exact parse coverage & $\kappa$ & AUC kw & AUC sem & $\Delta_{\mathrm{AUC}}$ & 95\% CI & sem@B & Row \\
\midrule
Qwen9B/Gemma9B & 385/450; 449/450 & 0.296 & 0.650 & 0.664 & $-$0.015 & $[-0.092, +0.062]$ & 17/45 & panel \\
Qwen9B/Mistral7B & 361/450; 226/450 & 0.267 & 0.667 & 0.649 & $+$0.018 & $[-0.060, +0.097]$ & 13/45 & panel \\
Qwen9B/Phi3.5 & 385/450; 336/450 & 0.090 & 0.654 & 0.526 & $+$0.128 & $[+0.027, +0.232]$ & 6/45 & panel \\
Qwen8B/Phi3.5 & 421/450; 336/450 & 0.168 & 0.697 & 0.579 & $+$0.118 & $[+0.029, +0.207]$ & 9/45 & panel \\
Gemma9B/Mistral7B & 449/450; 226/450 & 0.343 & 0.700 & 0.707 & $-$0.006 & $[-0.068, +0.060]$ & 18/45 & panel \\
Gemma9B/Qwen8B & 449/450; 421/450 & 0.382 & 0.600 & 0.557 & $+$0.042 & $[-0.030, +0.115]$ & 8/45 & panel \\
Gemma9B/Phi3.5 & 449/450; 336/450 & 0.299 & 0.662 & 0.647 & $+$0.015 & $[-0.059, +0.086]$ & 11/45 & panel \\
Mistral7B/Qwen8B & 226/450; 421/450 & 0.355 & 0.764 & 0.696 & $+$0.068 & $[+0.002, +0.136]$ & 22/45 & panel \\
Mistral7B/Phi3.5 & 226/450; 336/450 & 0.227 & 0.721 & 0.667 & $+$0.054 & $[-0.024, +0.131]$ & 11/45 & panel \\
Qwen8B/Qwen9B & 421/450; 361/450 & 0.230 & 0.537 & 0.631 & $-$0.094 & $[-0.189, +0.003]$ & 8/45 & panel \\
Qwen9B/Llama8B & 385/450; 8/450 (raw-as-final 80) & 0.186 & 0.623 & 0.559 & $+$0.064 & --- & 4/45 & excluded \\
Gemma9B/Llama8B & 449/450; 8/450 (raw-as-final 80) & 0.367 & 0.642 & 0.633 & $+$0.009 & --- & 13/45 & excluded \\
Mistral7B/Llama8B & 226/450; 8/450 (raw-as-final 80) & 0.309 & 0.764 & 0.701 & $+$0.063 & --- & 18/45 & excluded \\
Qwen8B/Llama8B & 421/450; 8/450 (raw-as-final 80) & 0.337 & 0.716 & 0.664 & $+$0.052 & --- & 11/45 & excluded \\
Llama8B/Phi3.5 & 8/450 (raw-as-final 80); 336/450 & 0.164 & 0.686 & 0.598 & $+$0.087 & --- & 7/45 & excluded \\
\bottomrule
\end{tabular}%
}
\caption{Full tagged final-span panel with parse coverage and bootstrap 95\% CIs on the gap; exact-parse counts are per model, in pair order. All ten included CIs lie below the primary prefix contract's CI $[0.324,0.464]$. Coverage is reported per pair, not per model, because the panel draws on three tagged generation phases and Qwen3.5-9B was regenerated in each; it therefore appears as 385/450 in the pairs drawn from the first phase and 361/450 in those drawn from the others. The five rows marked excluded involve Llama-3.1-8B, whose tagged final spans mostly fail to parse, so its final-span contract is not instantiated; those rows are printed here for completeness but are not part of the ten-pair panel the main text summarizes and carry no weight in any claim. Non-exact coverage for other models is truncation of an otherwise-parsed final channel, which preserves the span type. Short names in this table denote Qwen3.5-9B, Gemma-2-9B-it, Mistral-7B-Instruct-v0.3, Qwen3-8B, Phi-3.5-mini-instruct and Llama-3.1-8B-Instruct; the 2B/E2B pair of Section~\ref{sec:dissection} does not appear here. Generated by \texttt{\detokenize{scripts/analyze_larger_pair_final_repair.py}} and \texttt{\detokenize{scripts/build_panel_tables.py}}.}
\label{tab:pair-panel-full}
\end{table}

\section{External Audits of Released Artifacts}
\label{app:external}

Eleven contracts over eight released artifacts and one published recipe. Ten are instantiated from files we did not produce; the FrugalGPT-style contract re-uses its scorer and string-match label on our own judged traces, since that release carries no per-example generations. Each names the same fields; what differs is how much of the contract the release actually supports.

\subsection{UltraFeedback's Selection Target}
\label{app:ultrafeedback}

UltraFeedback \citep{cui2024ultrafeedback}\footnote{Ganqu Cui et al.; a different group from the OR-Bench authors of \citet{cui2025orbench}.} releases, per instruction, several completions carrying two
separately produced quality signals: a scalar \texttt{overall\_score}, and four fine-grained axis ratings with
their own rationales. Downstream work selects and trains on the scalar while claiming the property the axes
measure, which is the contract shape this paper audits. We take the scalar as the proxy $z$ and each of
truthfulness and the fine-grained mean as the construct $y$, over 11{,}375 directional pairs from 4{,}000
instructions.

The completions are complete outputs, so no off-span re-read is possible and Algorithm~\ref{alg:rlcaudit} returns UNDECIDABLE with a required field missing; what the release supports is the same-span check, and it shows no flag. Against truthfulness the target agrees on 0.818 of pairs with $\kappa(z,y){=}0.631$ and
ranks it at AUC 0.883; against the fine-grained mean, 0.839, $\kappa{=}0.672$, AUC 0.904. The length check does
not show the asymmetry either: length ranks the target at AUC 0.682 and truthfulness at 0.644, a gap of 0.038,
and against the fine-grained mean the construct is the more length-coupled of the two (0.694 versus 0.682).

UltraFeedback is one of the external artifacts on which only the same-span check can run, and it shows no flag; MT-Bench, under the same check, ranks at chance against both label sets. Taken together the external audits establish that the procedure returns verdicts on files we did not produce where the release supports the off-span control, and that where it does not, the same-span check flags nothing.
They do not establish that coupling is common, and we do not claim it. Reproduction:
\texttt{\detokenize{scripts/audit_ultrafeedback.py}}.

\subsection{The Null Distribution of the Disjoint Gap}
\label{app:null}

Proposition~\ref{prop:containment} gives $AUC(s,z^{c})=1/2$ under independence of the scored span and
its complement. That premise is false for generated text and the paper says so, which leaves
$\Delta_{\mathrm{dis}}$ a statistic whose null we had asserted rather than measured while applying fixed cutoffs to it. This appendix measures it. Two null hypotheses appear in the paper and are not competitors: the paired bootstrap interval addresses $H_0^{\mathrm{boot}}\!:\ \Delta_{\mathrm{dis}}=0$, the sampling uncertainty of the gap itself, while the permutation test below addresses $H_0^{\mathrm{perm}}\!:\ s \perp z^{c} \mid y$, whether the score carries off-span association the construct does not explain; the verdict is read from the second and both are reported.

\paragraph{Which null.} Permuting $z^{c}$ freely would break its relationship with the construct as
well, and a score that ranks the construct will rank anything correlated with it, so that null is too
permissive and would make almost every contract significant. The question is whether the score is
associated with the off-span proxy \emph{beyond} what the construct explains, so we permute $z^{c}$ within strata of $y$, a conditional permutation in the spirit of the conditional randomization test \citep{candes2018panning}. This holds $P(z^{c}\mid y)$ and all marginals fixed, leaves $AUC(s,y)$ exactly
invariant, and removes only the excess association. Two thousand permutations per contract give the
null of the orientation-robust gap and a permutation $p$-value.

\begin{table}[!htbp]
\centering
\footnotesize
\setlength{\tabcolsep}{4pt}
\begin{tabular}{lrrrrr}
\toprule
contract & $n$ & observed $\Delta_{\mathrm{dis}}$ & null mean & null $p_{95}$ & $p$ \\
\midrule
XSTest 450 (primary run) & 450 & $+0.101$ & $-0.029$ & $+0.061$ & $0.0105$ \\
OR-Bench hard 1k & 1319 & $+0.248$ & $+0.016$ & $+0.064$ & $0.0005$ \\
JailbreakBench & 200 & $+0.245$ & $+0.058$ & $+0.148$ & $0.0010$ \\
GSM8K, 50-char span & 1319 & $-0.033$ & $-0.046$ & $-0.025$ & $0.1574$ \\
GSM8K, 80-char span & 1319 & $-0.032$ & $-0.050$ & $-0.025$ & $0.1094$ \\
GSM8K, 120-char span & 1319 & $-0.001$ & $-0.031$ & $-0.007$ & $0.0170$ \\
GSM8K, 160-char span & 1319 & $-0.019$ & $-0.036$ & $-0.013$ & $0.1174$ \\
GSM8K, 200-char span & 1319 & $-0.027$ & $-0.034$ & $-0.013$ & $0.3058$ \\
GSM8K, 300-char span & 1319 & $-0.045$ & $-0.036$ & $-0.013$ & $0.7216$ \\
GSM8K, 600-char span & 1319 & $-0.083$ & $-0.039$ & $-0.015$ & $0.9980$ \\
HotpotQA, 50-char span & 2000 & $+0.121$ & $-0.002$ & $+0.014$ & $0.0005$ \\
HotpotQA, 80-char span & 2000 & $+0.073$ & $-0.017$ & $-0.002$ & $0.0005$ \\
HotpotQA, 120-char span & 2000 & $+0.009$ & $-0.028$ & $-0.013$ & $0.0015$ \\
HotpotQA, 160-char span & 2000 & $+0.010$ & $-0.018$ & $-0.003$ & $0.0020$ \\
HotpotQA, 200-char span & 2000 & $+0.026$ & $-0.018$ & $-0.002$ & $0.0005$ \\
HotpotQA, 300-char span & 2000 & $+0.044$ & $-0.026$ & $-0.011$ & $0.0005$ \\
HotpotQA, 600-char span & 2000 & $+0.038$ & $-0.023$ & $-0.004$ & $0.0005$ \\
\bottomrule
\end{tabular}
\caption{Each contract's own null for the disjoint gap: $z^{c}$ permuted within strata of $y$, 2{,}000 permutations. The set is diagnostic and includes rows later removed by preconditions and the 80-character GSM8K span, which lies outside the twenty contracts (Appendix~\ref{app:null}).}
\label{tab:null-main}
\end{table}

\noindent The null is contract-specific. Its 95th percentile is below zero on most rows, so there the
constants are conservative, but it reaches $+0.148$ on JailbreakBench --- a contract whose gap sits at the $0.15$ cutoff. A fixed constant is defensible only where it sits above the null of every
contract it judges, and ours does not. We report the constants because they are what the audit was
pre-registered with, and the per-contract null beside them.

The audit uses the per-contract null as its significance bar. The alternative bar, ``the paired bootstrap interval excludes zero'', is worth stating separately because the two do not always agree: over the eleven rows that carry the disjoint control they part on five, and on exactly one of those the verdict moves. Four of the five are settled before any significance test --- GSM8K at 50 characters by the resolution precondition, at 120 characters by orientation, and at 300 and 600 characters by a negative gap that no band can reach --- so they are unchanged (\texttt{\detokenize{scripts/significance_bar_comparison.py}}). The fifth is our own motivating contract, and it is the row where the choice matters: its gap exceeds its null ($p{=}0.011$) while its paired interval covers zero ($[-0.059,+0.255]$). Under the null bar it is CAUTION. Under the bootstrap bar the gap does not survive, and a reported gap of $+0.393$ then exits as CONTAINMENT, the reading under which the reported agreement is compatible with having been produced at this span. Both readings are reported. The tables carry CAUTION, the weaker claim in that it does not assert the reported agreement was produced and so the reading less favorable to our thesis; the CONTAINMENT the paired interval returns is stated beside it in Section~\ref{sec:scorefailure}, and neither is forced on the reader. On the other ten rows the verdicts do not depend on the independence premise: OR-Bench clears its null at $p{\le}0.001$ with its construct AUC shown at chance and would be a NO DEMONSTRATED CONSTRUCT RANKING row under either bar, though the resolution precondition withholds its verdict before either applies; JailbreakBench clears its null at the same level but its construct AUC cannot be certified at the equivalence band, so it is undecided under either.

\paragraph{Open ends of the null.} Permutation destroys spillover along with the association of
interest, so a gap above this null is evidence of a proxy--construct mismatch \emph{or} of
prefix-to-continuation dependence that does not run through the construct. $\kappa(z^{c},y)$ characterizes how far the two labels coincide and bounds what a residual can mean; it does not identify which of the two produced the residual (Section~\ref{sec:limits}). The HotpotQA rows are the visible case: every span
sits above its null, including spans whose gap is $+0.009$, which is a statement about direction only. Reproduction:
\texttt{\detokenize{scripts/disjoint_null_distribution.py}}.

\subsection{Calibrating the Control, and the Power to Use It}
\label{app:calibration}

\paragraph{Does $\Delta_{\mathrm{dis}}$ recover the earned share?} The disjoint control is an
estimator, and nothing in the audited data shows it returns what it claims, because there the truth is
what we are trying to learn. We therefore generate contracts whose containment fraction $\alpha$ is
set by construction. Three latents are needed: a construct latent drives $y$; an artifact latent lives
only in the scored span, standing for an opening template; and a vocabulary latent is expressed in
both the span and the remainder, standing for genuine predictability from the span about later text
that is \emph{not} the construct. The proxy rule fires on a mixture, $\alpha$ of it artifact and
$1-\alpha$ vocabulary; read on the span it sees both, read off the span only the vocabulary survives. Table~\ref{tab:alpha-sweep} in the main text lists the sweep.

$\Delta_{\mathrm{dis}}$ falls monotonically with the containment fraction (correlation $-0.958$), from
$+0.158$ where the agreement is entirely earned to $-0.036$ where it is entirely containment.
$\Delta_{\mathrm{ext}}$ does not track $\alpha$ ($+0.178$) and still reports $+0.273$ at $\alpha{=}1$,
where a control that had removed containment would report zero. On data where the answer is known, one
of these tracks $\alpha$ in rank and the other does not; neither is an estimator of the earned share on an absolute scale. $\alpha$ governs containment, not validity: at every $\alpha$ the proxy fires on latents other than the construct --- the artifact, the vocabulary latent, or a mixture of the two --- so the construct is a different latent throughout, so the large $\Delta_{\mathrm{dis}}$ at $\alpha{=}0$ is a
correct mismatch verdict on a contract with no containment, not a false alarm.

\paragraph{The ordering claim across a grid, not a line.} Six points on one line is thin evidence for
an instrument, so we repeat the construction over a grid: $\alpha\in\{0,\dots,1\}$ crossed with
spillover strength $\rho_v$ (how strongly the shared vocabulary latent is expressed off-span), sample
size $n\in\{200,800,4000\}$ and construct prevalence $\in\{0.10,0.30,0.50\}$, five repetitions per
cell, 36 slices and 1{,}080 simulated contracts in total.

\begin{center}\footnotesize
\begin{tabular}{rrrrrrr}
\toprule
$\rho_v$ & slices & median $\rho(\alpha,\Delta_{\mathrm{dis}})$ & orders & median $\rho(\alpha,\Delta_{\mathrm{ext}})$ & $\Delta_{\mathrm{dis}}$ at $\alpha{=}0$ & at $\alpha{=}1$ \\
\midrule
0.0 & 9 & $+0.029$ & 1/9 & $+0.429$ & $-0.024$ & $-0.027$ \\
0.3 & 9 & $-0.886$ & 9/9 & $+0.257$ & $+0.027$ & $-0.034$ \\
0.6 & 9 & $-1.000$ & 9/9 & $+0.257$ & $+0.110$ & $-0.028$ \\
0.9 & 9 & $-1.000$ & 9/9 & $+0.086$ & $+0.186$ & $-0.021$ \\
\bottomrule
\end{tabular}
\end{center}

\noindent ``orders'' counts slices with $\rho(\alpha,\Delta_{\mathrm{dis}})\le-0.7$. Two things hold
across the grid. Wherever the span predicts any off-span signal, $\Delta_{\mathrm{dis}}$ orders cells by
containment in 27 of 27 slices at every sample size and prevalence tested, including $n{=}200$. Where
it carries none, at $\rho_v{=}0$, the control is flat near zero at every $\alpha$ --- which is not a
failure but the behavior Proposition~\ref{prop:containment} predicts in the one regime where its
premise holds, and a useful check that the estimator is not manufacturing signal.
$\Delta_{\mathrm{ext}}$ never orders; its median correlation with $\alpha$ is positive in every $\rho_v$ group ($+0.43$ to $+0.09$), the wrong direction. Reproduction:
\texttt{\detokenize{scripts/synthetic_grid.py}}.

\paragraph{What the design can detect.} A paired interval separates a gap of $0.15$ from zero only when its half-width is smaller than that; the ``$y^{+}$ needed'' column scales each observed count by $(\text{half-CI}/0.15)^{2}$. This is a precision extrapolation at the observed effect size and prevalence, not a prospective power analysis.

\begin{center}\small
\begin{tabular}{lrrrrl}
\toprule
Contract & $y^{+}$ & $\Delta_{\mathrm{dis}}$ & half-CI & $y^{+}$ needed & powered \\
\midrule
XSTest 450 (primary run) & 46 & +0.101 & 0.157 & 51 & no \\
AdvBench 520 & 3 & +0.040$^{\ast}$ & 0.147 & --- & excluded on power \\
SimpleSafety 100 & 8 & +0.056 & 0.181 & --- & excluded on power \\
XSTest 100 / 512tok & 9 & -0.058 & 0.130 & --- & excluded on power \\
AdvBench 100 / 512tok & 3 & +0.015 & 0.169 & --- & excluded on power \\
OR-Bench hard 1k & 473 & +0.248 & 0.090 & 171 & yes \\
JailbreakBench & 27 & +0.245 & 0.157 & 30 & no \\
\bottomrule
\end{tabular}
\end{center}

Sixteen of the twenty contracts clear the minimum-positive-count precondition, and of those, fourteen clear this test;
the two that do not are both refusal settings. Our motivating contract needs 51 construct positives and
has 46, so its half-width of $0.157$ exceeds the cutoff it is judged against: the CAUTION it receives
is partly a statement about the design and not only about the contract. JailbreakBench fails the same
test, 27 positives against 30 needed. Among the refusal contracts only OR-Bench, at 473 positives and a half-width of $0.090$, is comfortably powered
for the verdict it receives (the two HotpotQA spans are comfortably powered as well, at half-widths
$0.034$ and $0.036$), and a reader wanting a single adequately powered NO DEMONSTRATED CONSTRUCT RANKING should read the 50-character HotpotQA row; OR-Bench, the powered public row, is withheld by the resolution precondition. Reproduction:
\texttt{\detokenize{scripts/power_and_calibration.py}}.

\subsection{Anchoring the Constants on Real Generations}
\label{app:semisynthetic}

\begin{figure}[!htbp]
\centering
\includegraphics[width=0.96\linewidth]{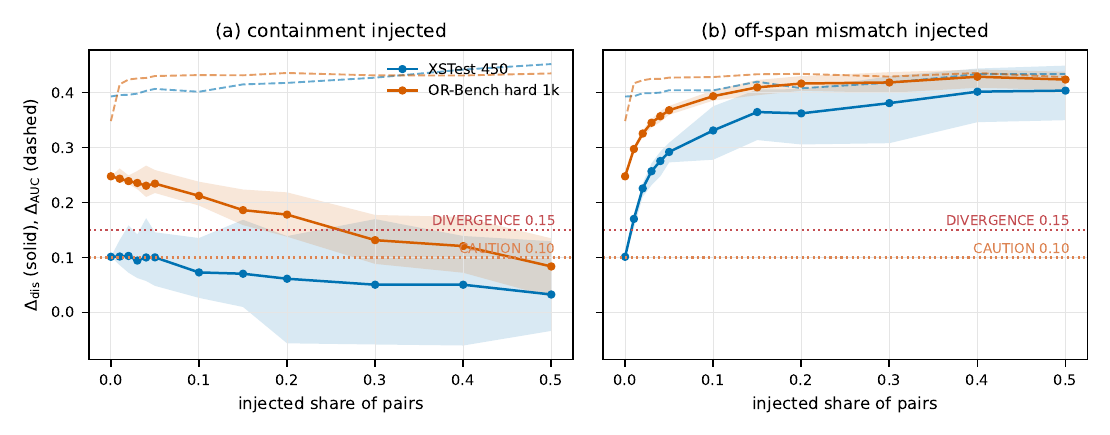}
\caption{$\Delta_{\mathrm{dis}}$ (solid, with the spread over twenty seeds) and $\Delta_{\mathrm{AUC}}$ (dashed) against the share of pairs altered. (a) A refusal opening on one side and a compliance opening on the other are written into the first 50 characters, so the same-span proxy fires and the prefix distance is large from evidence the construct does not carry; off-span text is untouched. (b) The same openings, plus a refusal sentence appended beyond the span on one side, so the off-span rule fires on pairs the score also separates. Dotted lines are the two constants. $^{\ast}$Quoted here for the power computation only; Table~\ref{tab:all-contracts} withholds it because the row fails the minimum-positive-count precondition.}
\label{fig:semisynthetic}
\end{figure}

The synthetic grid of Appendix~\ref{app:calibration} validates the disjoint control up to ordering and cannot say what a $\Delta_{\mathrm{dis}}$ of $0.10$ or $0.15$ amounts to on real text. This experiment injects known amounts of two things into the real XSTest and OR-Bench generations, with the judge labels held fixed, and reads what the controls return (Figure~\ref{fig:semisynthetic}; \texttt{\detokenize{scripts/semisynthetic_calibration.py}}). The injected pairs are drawn independently of the construct label, so $AUC(s,y)$ stays within $0.06$ of its recorded value throughout; twenty seeds per point, intervals are the 2.5--97.5 percentiles over seeds, and the zero-rate point reproduces the recorded rows.

\paragraph{Containment injected (a).} Writing template openings into up to half of the pairs raises the reported gap on both settings ($+0.39\to+0.45$ on XSTest, $+0.35\to+0.44$ on OR-Bench) and the partial control follows it, while $\Delta_{\mathrm{dis}}$ never rises: it is flat within the seed spread up to a $5\%$ injection and then falls, because the altered pairs carry no off-span positive and so dilute $AUC(s,z^{c})$. On real text, then, produced span evidence cannot push a contract toward the bands under the deciding control, whatever it does to the reported agreement.

\paragraph{Off-span mismatch injected (b).} A proxy positive the score separates but the construct does not is what the bands are meant to name, and injecting it moves $\Delta_{\mathrm{dis}}$ immediately. On XSTest the increment over the recorded $+0.101$ reaches $0.10$ at $1.6\%$ of pairs (seven of 450) and $0.15$ at $2.8\%$ (thirteen); on OR-Bench, whose recorded gap is already $+0.248$, at $3.2\%$ (42 of 1{,}319) and $11.3\%$ (149). Both curves saturate near $+0.40$--$0.43$ once the altered pairs dominate the off-span positives. The constants are therefore small on this scale: a few percent of pairs whose off-span proxy positive is driven by something only the score sees is enough to cross both, which is the sense in which they name a magnitude and cannot serve as a significance test, and is why each contract's own null carries that job.

\subsection{Error Rates on Structured Cases, and a Same-Precondition Baseline}
\label{app:error-rates}

The grid of Appendix~\ref{app:calibration} validates ordering. This experiment asks what the final rule returns on synthetic contracts whose structure is set by construction, and what a baseline returns that reports $AUC(s,y)$ and $\kappa(z,y)$ under the same preconditions, the same equivalence test and the same intervals but has no off-span proxy AUC. Seven cases are generated from three latents as in Appendix~\ref{app:calibration}: a valid proxy that reads the construct with noise on both spans; containment only, where the proxy fires on a span-only opening the score also reads; mismatch only, where the proxy measures a latent that is not the construct and the score tracks both; spillover only, where the proxy measures a vocabulary latent expressed on and beyond the span; a no-ranking pattern, where the score tracks the opening and the vocabulary but not the construct; the high-$\kappa$ counterexample, where $z^{c}$ agrees with $y$ on $80\%$ of items and the score predicts $z^{c}$ exactly; and a mixed case with half of the proxy's signal in the artifact and half in a mismatched latent. Each case is drawn 60 times at $n{=}450$ and $n{=}2000$ and passed through the rule of Section~\ref{sec:verdict} with 200 bootstrap and 150 permutation resamples (Table~\ref{tab:error-rates}).

\begin{table}[!htbp]
\centering
\scriptsize
\setlength{\tabcolsep}{3pt}
\resizebox{\linewidth}{!}{%
\begin{tabular}{llllll}
\toprule
 & & \multicolumn{2}{c}{final rule} & \multicolumn{2}{c}{same-precondition baseline} \\
Case & intended exit & $n{=}450$ & $n{=}2000$ & $n{=}450$ & $n{=}2000$ \\
\midrule
valid proxy & NO FLAG & NF 100 & NF 100 & none 100 & none 100 \\
containment only & CONTAINMENT & NF 5, CT 52, UD 43 & NF 7, CT 93 & UD 48, label 52 & label 98, not-ev. 2 \\
mismatch only & DIVERGENCE & NF 3, CA 18, DV 62, SF 2, UD 15 & CA 10, DV 90 & UD 13, label 83, not-ev. 3 & label 100 \\
spillover only & DIVERGENCE & CA 7, DV 37, SF 2, UD 55 & DV 100 & UD 48, label 47, not-ev. 5 & label 100 \\
no-ranking case & NO DEMONSTRATED CONSTRUCT RANKING & CA 5, DV 2, SF 7, UD 87 & DV 3, SF 97 & UD 87, label 12, not-ev. 2 & label 2, not-ev. 98 \\
high-$\kappa$ counterexample & DIVERGENCE & CA 2, DV 98 & DV 100 & none 100 & none 100 \\
half containment, half mismatch & CAUTION/DIVERGENCE & NF 47, CA 13, SF 2, UD 38 & NF 78, CA 22 & UD 40, label 57, not-ev. 3 & label 100 \\
\bottomrule
\end{tabular}%
}
\caption{Exit frequencies (\%) over 60 replicates per cell on synthetic contracts whose structure is set by construction; NF = NO FLAG, CA = CAUTION, DV = DIVERGENCE, CT = CONTAINMENT, SF = NO DEMONSTRATED CONSTRUCT RANKING, UD = UNDECIDABLE. The baseline reports $AUC(s,y)$ and $\kappa(z,y)$ under the same preconditions, equivalence test and intervals and has no off-span proxy AUC. Spillover is intended to exit as DIVERGENCE because the rule does not separate it from mismatch. Generated by \texttt{\detokenize{scripts/diagnostic_error_rates.py}}.}
\label{tab:error-rates}
\end{table}

Four things are read from the table. First, at $n{=}2000$ the rule returns the intended exit on $90$--$100\%$ of replicates for the valid, containment, mismatch, spillover and no-ranking cases, and spillover exits as DIVERGENCE, which is the non-identification Section~\ref{sec:limits} states. Second, at $n{=}450$, the motivating contract's size, abstention dominates: $43\%$ of containment cases, $55\%$ of spillover cases and $87\%$ of no-ranking cases are UNDECIDABLE, almost all at the equivalence step, which is the power limit Section~\ref{sec:discussion} reports on the real motivating contract. Third, the high-$\kappa$ counterexample is flagged DIVERGENCE on $98$--$100\%$ of replicates by the final rule and cleared on $100\%$ by the rule with the retired $\kappa$ exit, which is why that exit was retired; the baseline never flags it. Fourth, the mixed case is the rule's weak spot: at $n{=}2000$ it returns NO FLAG on $78\%$ of replicates and flags CAUTION on $22\%$, because the artifact half of the proxy is removed off-span and the mismatched half alone sits near the CAUTION constant, while the baseline flags every replicate as a label problem without saying which half is at fault. The baseline flags the containment case as a label problem on $98\%$ of replicates at $n{=}2000$, the prescription the off-span proxy AUC exists to prevent.

\paragraph{The same baseline on the twenty contracts.} Run on the twenty controlled contracts with the audit's preconditions (\texttt{\detokenize{scripts/baseline_same_preconditions.py}}), the baseline flags seven rows: the two the audit flags, XSTest 450 and the 50-character HotpotQA span, and the five GSM8K spans from 120 to 600 characters as label failures, since on the scored span the answer-string rule has not yet seen the answer and $\kappa(z,y)$ is $0.00$--$0.02$. Read beyond the span the same rule agrees with the construct at $\kappa(z^{c},y)$ of $0.62$--$0.67$ and the disjoint gap is negative, so the audit returns no coupling flag on them. What the off-span control adds on this corpus is therefore not additional flags but five withheld ones, together with the split between NO DEMONSTRATED CONSTRUCT RANKING and CONTAINMENT and the reversal of the partial control's reading on the two short HotpotQA spans (Appendix~\ref{app:added-controls}).

\subsection{Deployed Scores Under the Containment Control}
\label{app:prmcontainment}

Every other contract the control runs on has a score we wrote. Two released artifacts carry a deployed
partial-span score, the generations it scored, and an independent construct together, so the control
can run on them without our supplying any of the three: PRM800K \citep{lightman2024verify} and
Math-Shepherd \citep{wang2024mathshepherd}. On the Math-Shepherd side, the label's downstream uses are
what make the reading matter: \citet{sun2025freeprm} train against it as ground-truth process supervision,
\citet{yuan2024implicitprm} position their method as obtaining process rewards without such labels, and models
trained on it are scored against human first-error annotations \citep{zheng2024processbench}. We propose no
better label; perturbation stress tests and automated shortcut discovery are complementary lenses on the same
question \citep{shihab2026estprm,ravuru2026unmask}. We report both settings because they differ in whether
the failure this paper formalizes ever appears in the field, and this is the closest we can get to an
answer. The search space is narrow for a structural reason rather than an incidental one: containment
requires a score that reads part of the output, so a full-output score has an empty complement and
cannot instantiate the control at all. That rules out the artifacts with the richest releases
(MixInstruct's BARTScore and rank labels, RouterBench's per-model performance) and leaves the
step-level verifier family.

\texttt{scored-test-samples.jsonl} releases, per sampled solution to a MATH
test problem, the solution's steps as text, the process reward model's per-step rating distribution,
and sympy-graded final-answer correctness. The score is the deployed PRM's prefix score over steps
$1..k$, aggregated as the product of $P(\text{rating}{=}{+}1)$, the aggregation that best tracks the
released \texttt{prm\_score} (correlation 0.92; Appendix~\ref{app:prmsearch}). The construct is
\texttt{is\_correct}. The proxy is ours: the gold answer string appearing inside the
scored prefix, a cheap ``is this partial solution on track'' rule of the kind a practitioner could
compute, read off-span over steps $k{+}1..L$ for the disjoint control. AUCs are computed within problem
and averaged, because correctness varies within a problem and pooling would confound difficulty.

\begin{center}\footnotesize
\setlength{\tabcolsep}{4pt}
\begin{tabular}{rrrrrrl}
\toprule
span fraction & problems & $AUC(s,y)$ & $AUC(s,z)$ & $AUC(s,z^{c})$ & $\Delta_{\mathrm{dis}}$ & verdict \\
\midrule
0.25 & 300 & 0.602 & 0.331 & 0.570 & -0.032 $[-0.055,-0.008]$ & NO FLAG \\
0.50 & 300 & 0.701 & 0.373 & 0.650 & -0.051 $[-0.077,-0.024]$ & NO FLAG \\
\bottomrule
\end{tabular}
\end{center}

On 300 problems and 493{,}030 solutions the deployed prefix score predicts final correctness better than it predicts the on-track proxy, at both spans, and the orientation-robust gap under the disjoint control is negative at both; the AUC values differ from the 500-problem curve of Appendix~\ref{app:prmsearch} because the cuts differ. The contract takes no coupling flag.

\paragraph{The clear is not an artifact of our proxy.} One rule is an audit of the rule, not of the
deployed score, so we repeated the control holding the score and the construct fixed and varying only
the proxy across five rules a practitioner might reach for: the gold answer inside the prefix; a
\verb|\boxed| answer already committed; backtracking markers (``wait'', ``actually'', ``let me try'');
above-median arithmetic density; and above-median length.

\begin{center}\footnotesize
\setlength{\tabcolsep}{4pt}
\begin{tabular}{lrrrl}
\toprule
proxy rule & prevalence & $AUC(s,z^{c})$ & $\Delta_{\mathrm{dis}}$ & verdict \\
\midrule
gold in prefix    & 0.128 & 0.570 & $-0.032$ $[-0.055,-0.008]$ & NO FLAG \\
\verb|\boxed| committed & 0.000 & --- & --- & UNDECIDABLE (not instantiable) \\
backtracking      & 0.004 & 0.374 & $+0.023$ $[-0.003,+0.048]$ & NO FLAG \\
arithmetic density & 0.500 & 0.499 & $-0.102$ $[-0.114,-0.078]$ & NO FLAG \\
length            & 0.500 & 0.352 & $+0.046$ $[+0.024,+0.068]$ & NO FLAG \\
\bottomrule
\end{tabular}
\end{center}

\noindent Span fraction 0.25; the 0.50 rows are NO FLAG throughout, with $\Delta_{\mathrm{dis}}$
between $-0.051$ and $-0.026$. No rule reaches the caution band. Two rows carry information beyond
the verdict. The \verb|\boxed| rule has no positives at a partial span, because in this corpus an
answer is committed only at the end: the precondition of Section~\ref{sec:verdict} fires and the
contract is not instantiable, which is the intended behavior rather than a clear. And length, the
crudest surface rule, is the one row whose disjoint interval excludes zero. Its gap survives the
control at $+0.046$, less than half the caution threshold, so the verdict is unchanged --- and the control does not return zero by construction even on a contract that clears. We therefore state the negative result at its measured strength: the verdict is unchanged across the four instantiable rules at the quarter span, so it is not produced by our choice of proxy.

\paragraph{Math-Shepherd: separating span validity from construct mismatch.} Math-Shepherd's per-step labels are the training signal behind much of the released process reward model literature, and they are consumed as step-local quality. They are not produced that way: each label is estimated by completing the solution from that step and asking how often the completion reaches the right answer. The disjoint control nevertheless does not flag the span, which is what the estimand implies once it is written out; the paragraphs after the table locate the mismatch where it is.

The score is the mean of the released step labels over steps $1..k$; the construct is final-answer
correctness graded against GSM8K gold rather than against the last released label, so it is
independent of the annotation that produced the score. On 400 problems and 8{,}497 solutions the last
released label agrees with independent grading on 1.000 of solutions, confirming the release is what
it claims.

\begin{center}\footnotesize
\setlength{\tabcolsep}{4pt}
\begin{tabular}{llrrrl}
\toprule
span & proxy rule & $AUC(s,y)$ & $AUC(s,z^{c})$ & $\Delta_{\mathrm{dis}}$ & verdict \\
\midrule
0.25 & backtracking       & 0.678 & 0.383 & $-0.061$ $[-0.174,+0.112]$ & UNDECIDABLE \\
0.25 & arithmetic density & 0.678 & 0.481 & $-0.160$ $[-0.183,-0.135]$ & NO FLAG \\
0.25 & length             & 0.678 & 0.520 & $-0.158$ $[-0.181,-0.134]$ & NO FLAG \\
0.50 & backtracking       & 0.836 & 0.315 & $-0.151$ $[-0.319,+0.059]$ & UNDECIDABLE \\
0.50 & arithmetic density & 0.836 & 0.475 & $-0.312$ $[-0.337,-0.286]$ & NO FLAG \\
0.50 & length             & 0.836 & 0.534 & $-0.302$ $[-0.328,-0.276]$ & NO FLAG \\
\bottomrule
\end{tabular}
\end{center}

\noindent The backtracking rule fires on 7 and 4 off-span sides respectively, below the minimum-positive-count precondition, so it returns no verdict --- the precondition binds on the disjoint proxy exactly as it
binds on the construct. The two powered rules clear decisively, and they clear because $AUC(s,y)$ is
high, not because $AUC(s,z^{c})$ is at chance: a value function estimated from completions should
predict final correctness well, and this one does.

\paragraph{Rollout estimation does not constitute a span mismatch.} A label attached to step $k$ is estimated by completing the solution from step $k$, which invites the reading that the label's span is mis-declared. The estimand is $V(x_{<k})=\Pr[\text{correct}\mid x_{<k}]$, a function of the observed prefix, and rollout sampling is Monte Carlo integration of that quantity rather than evidence that the score reads the continuation. The same reasoning applies to any value-function estimate, and the high $AUC(s,y)$ reported above is the behaviour expected of a well-estimated value function. The mismatch is elsewhere, between the rollout-value proxy and the step-correctness construct the label is subsequently used for.

\paragraph{The construct mismatch, measured.} The proxy reports the probability that \emph{some}
completion from the prefix reaches the right answer; consumers read it as ``step $k$ is correct.'' On
127{,}482 parsed GSM8K solutions, 95{,}694 reach a wrong final answer. Within those, 120{,}661 of
310{,}097 non-final steps (38.9\%) still carry a positive label, and 16{,}697 solutions (17.4\%) are
positive at every non-final step, so the labels do not flag failure before the final step. The two interpretations diverge on wrong-but-recoverable steps: there the value reading makes the positive label correct---the prefix was recoverable and this continuation did not recover it---while the step-correctness reading makes it a mislabel. The 38.9\% therefore bounds the set of steps on which the two interpretations may disagree rather than measuring that disagreement, since a step that is correct on its own terms keeps its positive label under both readings and the released files do not say which steps those are. No containment control addresses it, because the proxy genuinely predicts the construct we audited against. Reproduction:
\texttt{\detokenize{scripts/mathshepherd_estimand_vs_construct.py}}.

\paragraph{An estimate from the flagged subset.} For a step whose asserted numerical equality is false under the arithmetic convention stated below, the step is incorrect under the step-correctness interpretation no matter what follows, while its positive rollout label can still stand under the value interpretation whenever the mistake is recoverable. Each such step is therefore a disagreement between the two interpretations. The natural external cross-check --- ProcessBench's human first-error annotations \citep{zheng2024processbench} --- cannot supply this: its GSM8K solutions come from Qwen2/2.5 and Llama-3 generators, none of them Math-Shepherd's, so no solution carries both label types. Mechanical verification of the equations inside Math-Shepherd's own steps can. Equations are parsed conservatively: numeric chains only, calculator annotations parsed once, and a chain counts as correct if either standard precedence or left-to-right evaluation matches its stated result, since generators write both; the pre-registered kill gate was coverage below $30\%$. Coverage is $87.4\%$ --- 105{,}498 of the 120{,}661 positive non-final steps in failing solutions contain a parseable equation --- and 4{,}690 of those ($4.4\%$ of covered, $3.9\%$ of all positive steps) are flagged as arithmetically false. The same check flags $2.2\%$ in succeeding solutions, an upper bound on parser artifact since those flags include wrong steps the solution recovered from. At the solution level, 4{,}334 failing solutions carry at least one flagged positive step, including 1{,}259 of the 16{,}697 never-warned solutions ($7.5\%$). An earlier version of the checker flagged 6{,}228 steps; a hand audit of 200 of its flags (seed 20260905) found 53 parser artifacts (fraction targets such as ``$=1/5$'' read as 1, percentages, leading-dot decimals, chained equalities, tails of larger expressions, algebra) and 39 steps false only beyond the precision the model displayed, and the checker was rewritten to close those classes. A second audit of 200 flags from the rewritten checker finds 157 false as written (128 genuine errors, 23 sign errors, 6 other), 9 false only beyond displayed precision and 34 unit, time or percentage shorthand the parser still misreads: precision $78.5\%$, Wilson 95\% interval $[72\%,84\%]$ \citep{wilson1927probable}, so the precision-adjusted count is roughly 3{,}700 with an interval of $3{,}400$--$3{,}900$. Both audits are recorded beside the counts (\texttt{\detokenize{analysis_results/mathshepherd_arithmetic_lower_bound_v2.json}}; \texttt{\detokenize{scripts/mathshepherd_arithmetic_lower_bound_v2.py}}). The construct mismatch is therefore measured from both sides: roughly 3{,}700 precision-adjusted false steps as a lower-end estimate from the flagged subset, and $38.9\%$ of non-final steps in failing solutions as an upper-end count. Reproduction: \texttt{\detokenize{scripts/mathshepherd_arithmetic_lower_bound.py}}.

Across both artifacts, then: a released PRM score and a released step-label aggregate used as a score, two spans each, twelve proxy--span cells that pass
their preconditions (eight in PRM800K, four here) and four that do not. No cell reaches a mismatch, so
what we cannot say is that the \emph{control} caught a deployed contract exhibiting the failure. The
contract's fields did, in the artifact above, and that asymmetry is the finding. \noindent Reproduction, in \texttt{\detokenize{scripts/}}:\\
\texttt{\detokenize{audit_prm800k_containment.py}},
\texttt{\detokenize{audit_prm800k_proxy_family.py}},
\texttt{\detokenize{audit_mathshepherd_containment.py}}.

\subsection{RouterBench's Released Artifacts}

\label{app:routerbench}

\paragraph{What the grade-school-math labels describe.} The not-instantiable verdict on this slice
follows from its arithmetic (fractional labels in quarters over a single stored response), but the size of the resulting error is a measurement, so we made it. Matching each of the 7{,}450 rows to its
GSM8K source recovers the gold answer for all of them; scoring the single stored response by exact
match then gives Table~\ref{tab:routerbench-gsm8k}.

\begin{table}[!htbp]
\centering
\small
\begin{tabular}{lrrrr}
\toprule
Model & released label & stored response EM & gap & label $=$ EM \\
\midrule
claude-v2 & 0.663 & 0.894 & $-$0.231 & 0.0024 \\
claude-v1 & 0.651 & 0.857 & $-$0.207 & 0.0007 \\
gpt-4-1106-preview & 0.659 & 0.844 & $-$0.185 & 0.0008 \\
claude-instant-v1 & 0.627 & 0.772 & $-$0.145 & 0.0016 \\
gpt-3.5-turbo-1106 & 0.605 & 0.749 & $-$0.144 & 0.0023 \\
zero-one-ai/Yi-34B-Chat & 0.548 & 0.580 & $-$0.032 & 0.0030 \\
WizardLM/WizardLM-13B-V1.2 & 0.506 & 0.544 & $-$0.038 & 0.0032 \\
mistralai/mixtral-8x7b-chat & 0.519 & 0.542 & $-$0.023 & 0.0051 \\
meta/llama-2-70b-chat & 0.523 & 0.533 & $-$0.010 & 0.0034 \\
meta/code-llama-instruct-34b-chat & 0.457 & 0.405 & $+$0.051 & 0.0031 \\
mistralai/mistral-7b-chat & 0.412 & 0.343 & $+$0.069 & 0.0044 \\
\bottomrule
\end{tabular}
\caption{RouterBench grade-school-math, 7{,}450 rows. ``Released label'' is the mean stored
performance value; ``stored response EM'' is the exact-match correctness of the one response the
resource actually stores. The two agree on $0.3\%$ of rows, and the gap ranges over $0.300$ across
models. Reproduction:
\texttt{\detokenize{scripts/routerbench_gsm8k_label_provenance.py}}.}
\label{tab:routerbench-gsm8k}
\end{table}

Three consequences, in decreasing order of what they license. The label does not describe the released
response, so a contract whose score reads that response is validated against something else; this is
the not-instantiable verdict, now with a magnitude. The released
\texttt{\detokenize{oracle_model_to_route_to}} column inherits it, marking no row of this slice
unsolvable against 242 of 10{,}042 on hellaswag, so an oracle bound computed over all of RouterBench
is one in which a fifth of the benchmark cannot fail, yet whose per-row maximum is fractional almost everywhere (99.2\% of its labels are non-binary), so the slice pulls the mean of per-row maxima down; dropping it therefore raises that bound from $0.912$ to $0.954$. And ranking the eleven models by the released label and by the responses gives different
orders (four of eleven move, Spearman $0.955$).

What this does not show is that a published leaderboard is wrong. Aggregate rankings over the whole
benchmark barely change when the slice is dropped (Spearman $0.991$; the two models that move are
adjacent and separated by $0.0003$), and the top model is the same under either scoring. We report the
robustness next to the defect, because a reader is entitled to know that the cost of this particular
not-instantiable verdict falls on response-conditional uses of one slice rather than on the
benchmark's headline ordering.

\begin{table}[!htbp]
\centering
\scriptsize
\begin{tabular}{lrrrrrl}
\toprule
RouterBench slice & n & $\kappa(z,y)$ & AUC $z$ & AUC $y$ & sem@B & Outcome \\
\midrule
MMLU professional law (letters) & 1534 & 0.728 & 1.000 & 0.904 & 111/153 & no flag \\
MT-Bench (stored GPT-4 vs.\ independent judge) & 80 & 0.074 & 0.448 & 0.463 & 2/8 & at chance, both labels \\
Grade-school math & 7450 & -- & -- & -- & -- & contract not instantiable \\
\bottomrule
\end{tabular}
\caption{External RLC-Audit of RouterBench released artifacts (pair claude-instant-v1 / Mixtral-8x7B; $z$ = released-label disagreement, $y$ = independent-adjudication or ground-truth disagreement; sem@B = construct disagreements in the top-10\% routed queue).}
\label{tab:routerbench-audit}
\end{table}

Section~\ref{sec:routerbench} summarizes the external audit of RouterBench \citep{hu2024routerbench}, which releases prompts, per-model responses, and per-model labels for eleven models; Table~\ref{tab:routerbench-audit} reports the three slices. The score $s$ is TF-IDF cosine distance between the pair's released responses, the proxy $z$ is disagreement under the benchmark's released labels, and the construct $y$ is disagreement under independent adjudication or ground truth, with a $B{=}10\%$ routed queue. This appendix records the per-slice mechanics and caveats.

\paragraph{NO FLAG: MMLU professional law.}
Released responses are bare option letters, so the response-distance score is essentially the letter-disagreement proxy, and the released exact-match labels are ground truth. Proxy AUC is trivially 1.000; construct AUC is 0.904 and $\kappa(z,y){=}0.728$; the routed queue is majority construct disagreements (111/153). The residual gap has a fully mechanical explanation---pairs where both models are wrong with different letters are proxy-positive but construct-negative---so proxy and construct evidence move together and the audit clears the contract.

\paragraph{Undecidable at the label stage: MT-Bench.}
RouterBench stores continuous GPT-4 single-answer grades for its 80 MT-Bench conversations. We binarize the stored grades at $0.7$ and obtain an independent binary adequacy adjudication from a different judge (claude-haiku-4-5) on the same released responses. Per-response side labels agree moderately (raw agreement 0.738/0.713, $\kappa$ 0.497/0.444 for the two models), but the pair-level disagreement events a router would target barely agree: $\kappa(z,y){=}0.074$ (0.018 when binarizing at $0.5$). The response-distance score obtains no proxy evidence to begin with (AUC 0.448 for $z$, 0.463 for $y$, both near chance, on full and opening-200 spans). The audit therefore flags this contract before any score comparison: a routing paper that validated a quality-disagreement router against the stored labels on this slice would be measuring an event with near-zero agreement across judges, so the resulting AUC could not be read as construct evidence---regardless of whether it were high or low. This is a label-stage analogue of the main-text failure: the issue is not that the stored labels are wrong, but that the disagreement construct is not judge-robust on this slice, and only the $\kappa(z,y)$ field of the contract reveals it.

\paragraph{Not instantiable: grade-school math.}
Every one of the 7450 grade-school-math rows carries a fractional label in $\{0.25, 0.5, 0.75\}$ for at least one model (e.g., claude-instant-v1: 1439 rows at 0.25, 746 at 0.5, 5232 at 0.75, only 33 at 0 or 1), while exactly one response string is stored per model per row. A fractional exact-match label must aggregate multiple generations, so the released label cannot equal the binary exact-match outcome of the released response. No score--proxy--construct contract over the released artifacts can be instantiated, and any audit---or any router trained or validated on (response text, label) pairs from this slice---inherits a label--evidence mismatch that is visible from the released files alone.

Two caveats bound these claims. The independent MT-Bench adjudication is itself an LLM judge, so the flag establishes judge-instability of the disagreement event rather than which judge is correct; and $n{=}80$ with 17--23 positives makes the AUC estimates coarse, which is why the load-bearing field is $\kappa(z,y)$, not the AUCs. Every number above was computed from RouterBench's released files plus one independent adjudication pass, so the audit can be applied to released routing benchmarks without rerunning generation.

\subsection{Hybrid LLM and MixInstruct}
\label{app:hybridllm}

Hybrid LLM \citep{ding2024hybrid} routes on a predicted BARTScore quality gap. That gap is the object its router is trained and validated against, so the contract question is whether it is evidence for the adequacy judgment the router is sold as serving. MixInstruct \citep{jiang2023llmblender} releases, per instruction, the text of twelve candidate responses, per-candidate automatic scores including \texttt{bartscore}, and \texttt{cmp\_results}: pairwise adequacy verdicts elicited separately. The two labels are therefore available over the same responses without any new generation or judging.

We take every directional verdict (``A is better''/``B is better''; ties carry no direction) for which both candidates have a released BARTScore, giving 253{,}519 comparisons over 5{,}000 test instructions. Writing $z$ for the sign of the BARTScore gap and $y$ for the independent verdict, pooled agreement is 0.713 with $\kappa(z,y){=}0.426$ (bootstrap over comparisons $[0.423,0.430]$; over the 4{,}771 instructions as clusters $[0.421,0.433]$; the 5{,}000 test instructions reduce to 4{,}771 because 229 contribute no directional comparison inside the model pairs with at least 200 jointly labeled rows) and $\mathrm{AUC}(\text{gap},y){=}0.778$; per-pair $\kappa$ ranges from 0.167 to 0.402 across the model pairs with at least 200 jointly labeled rows. The target and the construct agree well above chance and well below identity. The $\kappa$ value is not itself the verdict---the ten-pair panel of Section~\ref{sec:boundary} clears contracts with lower $\kappa(z,y)$ because their proxy and semantic evidence move together---which is why the length asymmetry below, not the agreement level, is what this audit reports.

The residual is not symmetric noise. The BARTScore gap correlates with the response-length difference at $r{=}0.239$, while the independent verdict correlates with it at $r{=}0.079$; length alone ranks the proxy at AUC 0.609 and the construct at 0.542, a difference whose instruction-cluster interval is $[0.063,0.072]$ (\texttt{\detokenize{scripts/hybridllm_cluster_bootstrap.py}}). A surface property that barely moves the construct moves the deployed target substantially, which is the coupling direction this paper's contract is designed to expose.

The asymmetry is specific to length, and we report the bound rather than the headline. Widening the surface channel to eleven content-blind features of the two responses---character and word count, mean word and sentence length, type-token ratio, sentence count, list markers, digit and uppercase share, terminal punctuation, and bigram self-repetition---and scoring an out-of-fold logistic model on each target separately recovers the deployed proxy at AUC 0.704 and the construct at 0.681 (paired bootstrap on the difference $[+0.021,+0.025]$), positive in 88 of 132 model pairs with at least 200 rows (median $+0.012$). The length family carries the whole asymmetry: word count $+0.074$, characters $+0.067$, sentence count $+0.061$, while type-token ratio ($-0.056$) and mean word length ($-0.022$) run the other way. The proxy is more length-coupled than the construct without being generally more surface-coupled; and the independently elicited construct is itself substantially recoverable from content-blind form, which is a property of automatic adequacy labels rather than of this router. We therefore state the finding as a length asymmetry on a deployed target, and claim nothing about its proxy being surface-driven overall. We report this as a measured length asymmetry in the ranking, not as a coupling failure: the next paragraph shows the asymmetry does not reach the operating point, so the same-span check shows no flag. The responses are complete, so the off-span control cannot run and the contract's Algorithm~\ref{alg:rlcaudit} verdict is UNDECIDABLE with a required field missing (Table~\ref{tab:verdict-tally}); everything in this appendix is that supplementary check, not a verdict. The second elicitation is itself automatic (Section~\ref{sec:routerbench} discusses the
same caveat).

The audit's own frame asks about the operating point, so we build the queue a router using this target would act on. Ranking the 253{,}519 comparisons by $|$gap$|$ and taking the most decisive 10\%, the target contradicts the independent verdict on 0.138 of that queue ($[0.134,0.142]$ over comparisons; $[0.129,0.145]$ resampling instructions as clusters, with the queue re-ranked on every resample), against 0.287 overall; the same holds at 1\% (0.134) and degrades smoothly to 0.188 at 50\%. Mean absolute length difference inside the 10\% queue is 1.30 times the overall value, and the queue shares only 20.4\% of its examples with a queue ranked by length alone, which itself contradicts the verdict on 0.231. In the observed queue, higher confidence is associated with higher agreement with the construct. This narrows the finding: the length asymmetry is real in the ranking, and the queue's lower disagreement and its 20.4\% overlap with a length-ranked queue are observed enrichments at the operating point, not a matched or stratified test that length plays no role there. We report it because the operating point is the unit this paper argues for, and it does not support reading Hybrid LLM as a field instance of coupling.

{\raggedright\sloppy Reproduction: \texttt{\detokenize{scripts/hybridllm_routed_composition.py}}, \texttt{\detokenize{scripts/audit_hybridllm_mixinstruct.py}}, and \texttt{\detokenize{scripts/audit_hybridllm_surface_axes.py}} recompute every number above from the public split; results in \texttt{\detokenize{analysis_results/hybridllm_mixinstruct_audit.json}} and \texttt{\detokenize{analysis_results/hybridllm_surface_axes.json}}.\par}

\subsection{PRM-Guided Search}

\begin{figure}[!htbp]
\centering
\includegraphics[width=0.62\linewidth]{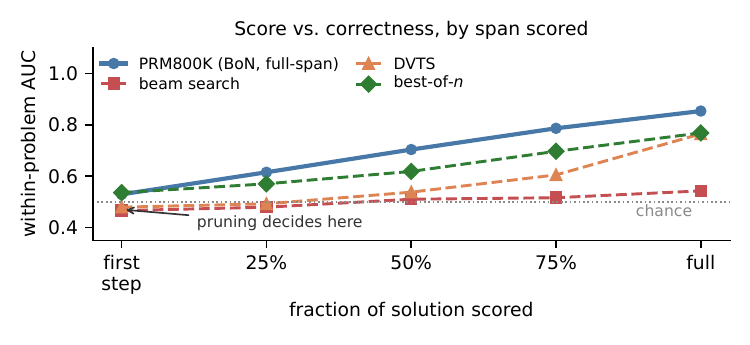}
\caption{Within-problem AUC of each deployed score against final correctness, by the fraction of the
solution scored (MATH-500, Llama-3.2-1B, 8B PRM; PRM800K, 815{,}632 solutions). All four contracts are
measured at the same five fractions and the search arms are averaged over pool sizes. Selection-harm
rates are discussed in the text rather than plotted, because the four pools have different correct
base rates (0.43, 0.32, 0.23, 0.53) and the rates are not comparable across contracts.}
\label{fig:deployed-spans}
\end{figure}

\label{app:prmsearch}

\paragraph{Normalising the selection-harm rates.} Normalizing for pool composition does not rescue the reading that partial-span selection is the
expensive kind (the harm rates themselves are in Section~\ref{sec:noflag}), and this is the arithmetic. Harm is already conditional on a correct answer being in
the pool, but the pools differ in how many correct answers they hold: beam search 0.43, DVTS 0.32,
best-of-$n$ 0.23. Dividing each harm rate by the share of incorrect candidates available to be chosen,
$h/(1-\text{base rate})$, gives $56.6\%$ for beam search and $66.8\%$ for best-of-$n$ at $n{=}256$, so
the full-span contract remains the worse of the two under the normalization as well as before it. We
therefore do not treat these rates as a cross-contract comparison in either form.

The preceding external artifacts verify completed outputs. This section audits released resources in which the deployed decision is taken on a partial span: a process reward model (PRM) scores an unfinished chain, and pruning or selection acts on that score while the construct---final-answer correctness---depends on evidence the span does not yet contain. These are the field contracts closest in shape to the paper's primary one, and both are auditable from released files alone, with no new generation and no judge calls.

\textbf{Search-and-learn.} The search-and-learn pipeline \citep{beeching2024scaling} generates Llama-3.2-1B solutions to MATH-500, scores each partial chain step-by-step with an 8B PRM \citep[RLHFlow Deepseek-data, last-step aggregation;][]{rlhflow2024prm}, prunes beams during generation, and returns the completion with the highest final score (\texttt{pred\_naive}). The released dumps retain, for every surviving completion, the full per-step score trajectory, the completion text, and the gold answer. We grade every completion against gold (Hendrycks-style normalization with a sympy equivalence fallback), take the per-step score at a span fraction as the proxy at that span, and audit twenty released configurations: beam search at $n\in\{16,64,256\}$ and diverse verifier tree search (DVTS) at $n{=}256$, five seeds each.

\begin{table}[!htbp]
\centering
\scriptsize
\begin{tabular}{lrrrrr}
\toprule
Configuration & harm@pool & acc naive & acc majority & AUC first step & AUC full \\
\midrule
beam $n{=}16$  & 0.154 $\pm$ 0.005 & 46.4 & 48.2 & 0.492 & 0.566 \\
beam $n{=}64$  & 0.260 $\pm$ 0.023 & 48.3 & 50.9 & 0.461 & 0.532 \\
beam $n{=}256$ & 0.325 $\pm$ 0.014 & 48.7 & 51.9 & 0.445 & 0.529 \\
DVTS $n{=}256$ & 0.412 $\pm$ 0.018 & 49.7 & 53.3 & 0.479 & 0.764 \\
best-of-$n$ $n{=}256$ & 0.515 $\pm$ 0.006 & 42.5 & 43.7 & 0.536 & 0.768 \\
\bottomrule
\end{tabular}
\caption{Search-and-learn released dumps, MATH-500, mean $\pm$ sd over five seeds; best-of-$n$ is the same pipeline's sampling-only release, a same-generator full-span control. harm@pool: fraction of problems whose pool contains a correct completion but whose deployed selection (score argmax) returns a wrong one. acc: released selection metrics under our grader (\%). AUC: mean within-problem AUC of the per-step score against final correctness, at the first scored step and over the full trajectory.}
\label{tab:prmsearch}
\end{table}

Table~\ref{tab:prmsearch} gives the result. Within problems, the deployed score is at or below chance at the first scored step, the point where pruning begins, in every pruning configuration (best-of-$n$ prunes nothing and is the full-span control, at $0.536$), and reaches only 0.53--0.76 over the full trajectory. These are selection-conditional descriptions of the surviving pool, not a measurement of pruning quality on the candidates that were pruned. The released selection metrics record what that costs at the operating point: with a correct completion in the pool, the score's argmax picks a wrong one on 15.4\% of problems at $n{=}16$, rising monotonically to 32.5\% at $n{=}256$ and 41.2\% for DVTS, and proxy-argmax selection trails the pipeline's own majority vote at every pool size. The surface check shows the familiar sign structure: completion length correlates with the score at $r{\approx}{-}0.51$ to ${-}0.53$ and with correctness at ${-}0.36$ to ${-}0.38$ (DVTS, which releases no token counts, gives ${-}0.34$ versus ${-}0.30$ on character length).

\textbf{PRM800K.} The scored test samples of \citet{lightman2024verify} release 815{,}632 model solutions to 500 MATH problems with per-step PRM score distributions, an outcome-model score, and correctness. Taking the product-aggregated prefix score at a span fraction as the proxy, within-problem AUC against correctness rises monotonically across spans: 0.529 at the first step, 0.615 at a quarter, 0.704 at half, 0.786 at three quarters, and 0.854 over the full solution. The span this PRM is deployed on---full-solution best-of-$n$---is therefore comparatively aligned, and we report it as such; what the audit reads is the curve
across spans, not the endpoint. Selection harm grows with pool size here too (9.3\% of solvable problems at $N{=}4$, 17.5\% at $N{=}64$, 19.2\% over the full pool of ${\sim}1{,}600$), and the score is more length-coupled than the construct ($r{=}{-}0.51$ versus ${-}0.44$).

\textbf{Scope and limits of these audits.} The partial-span contracts show a proxy--construct link that is near chance exactly where the deployed pruning decision is taken, with the cost visible in the pipeline's own released metrics. A same-generator best-of-$n$ control bounds the comparison: its independent samples produce much sparser pools (base rate 0.229 against beam's 0.427) and the highest selection harm of all three contracts (51.5\%), so harm rates index pool composition and are not comparable across contracts---beam search's richer pools are themselves the product of its unmeasured pruning stage. The cross-contract claim is carried by the span at which each score decides (within-problem AUC 0.77 for best-of-$n$ at full span, 0.45 for beam at its first pruning step), not by harm levels. Three caveats bound the claim. First, as with Appendix~\ref{app:hybridllm}, this is a failing proxy--construct link on field systems, not a demonstration of the shared-surface mechanism of Section~\ref{sec:mechanism}; the length asymmetry is consistent with it but not a dissection. Second, our grader is conservative: on the eight configurations whose released accuracy files survive, it scores 1.2--2.4 points below the pipeline's own lenient parser (mean $-2.0$), so harm rates may be slightly overstated; the ordering naive $<$ weighted $\approx$ majority and every monotonic trend match the released metrics. Third, pruned beams are not retained, so the beam-search counterfactual is indirect; and PRM800K's released aggregate score is not exactly reproducible from the per-step fields (product aggregation correlates with it at 0.92 and gives full-span AUC 0.922 versus 0.932 for the released score, both pooled over all solutions rather than the within-problem average of Section~\ref{sec:partial-span}), so the span sweep uses the product reconstruction. Reproduction: \texttt{\detokenize{scripts/audit_searchandlearn_beam.py}} and \texttt{\detokenize{scripts/audit_prm800k_scored.py}}; results in \texttt{\detokenize{analysis_results/searchandlearn_beam_audit.json}} and \texttt{\detokenize{analysis_results/prm800k_audit.json}}.

\section{The Keyword Proxy Is Not a Straw Man}
\label{app:strawman}

Trained refusal classifiers exist, so it is fair to ask whether the motivating contract fails only because its
proxy is a twelve-term keyword list. We substituted a public refusal classifier
(\texttt{\detokenize{protectai/distilroberta-base-rejection-v1}}) for that list, holding the score at the
prefix-50 TF-IDF distance, and audited the result on the clean run.

The substitution cannot be made at that span. Read over the same 50 characters the score reads, the classifier
never fires: its rejection probability has median 0.000 and maximum 0.002 across all 548 responses (550 prompts less the two whose final-channel judge output fails to parse on one side), so the
proxy has no positives and the contract is not instantiable. Read over the full response instead, the same
classifier fires on 52.7\% of responses and the contract returns DIVERGENCE again (gap $+0.349$,
$\kappa(z,y){=}{-}0.004$), because it is then a different span from the score's.

The instructive numbers are the construct validities. At the prefix span the keyword label predicts the judged
refusal decision at AUC 0.505 and the classifier at 0.366, both at or below chance; the same classifier over
the full response reaches 0.726, and over the parsed final answer 0.871 (Appendix~\ref{app:agcr}). Fifty
characters do not contain a refusal decision for either instrument. The keyword list is usable there only
because it emits rare surface matches, which is also what makes it visible to a score reading the same
characters. So the contract is not a straw man built from a weak proxy: at the span practitioners actually score, only the keyword-style proxy produced positive labels under the classifier and threshold we tested, and the audit's verdict is that such a proxy does not license a semantic routing claim. Reproduction:
\texttt{\detokenize{scripts/validated_proxy_audit.py}}.

\paragraph{The score is a controlled instrument, not a deployed router.} Showing the proxy is not a straw
man does not help if the \emph{score} is one, and no deployed system routes on a 50-character TF-IDF
cosine distance. The TF-IDF score is a controlled diagnostic instrument rather than a deployed router: it is built to isolate a
mechanism, not a router anyone ships, and Section~\ref{sec:contract-revision} says explicitly that the
repair validates a less-coupled contract and not a deployable router.

Scoring a partial span is not exotic: deciding from an unfinished generation is the normal case in three deployed families. Cascades and routers commit before the expensive model runs, which is the whole source of their savings; that timing is not itself a partial span, since such a system can also read a cheap model's completed answer, and the span the scorer reads is what matters here \citep{chen2023frugalgpt,ding2024hybrid,
aggarwal2024automix,gupta2024language}. Process reward models score partial chains and prune beams
mid-generation, so the operative span is a prefix by construction
\citep{lightman2024verify,beeching2024scaling}; Section~\ref{sec:partial-span} audits exactly that
setting and finds the released PRM near chance at the span where pruning acts. Speculative and
early-exit decoding decide on partial output for latency. In all three the decision is taken over less
text than the construct needs, which is the structural condition
Proposition~\ref{prop:containment} describes.

 The partial-output \emph{decision structure} is representative of
deployed practice; the 50-character cutoff and the particular \emph{score family} are diagnostic choices, and we make no claim that a TF-IDF
distance is what anyone would deploy. The results that depend on the specific score are confined to
Section~\ref{sec:dissection}, and Appendix~\ref{app:calibration-controls} reproduces the split under two neural prefix encoders, so the mechanism is not an artifact of lexical matching.

\subsection{RouteLLM's Deployed Routing Target}
\label{app:routellm}

Every other contract the containment control runs on is one we built, for the reason
Section~\ref{sec:release} measures: the control needs per-example generations and a proxy rule, and
the sampled papers publish neither together. RouteLLM \citep{ong2025routellm} is an exception we found
outside that sample, and we report it because it is the only case where the control reaches a
deployed routing target's own artifacts.

\paragraph{The contract.} \texttt{routellm/gpt4\_dataset} releases 119{,}101 prompts with the full
response of gpt-4-1106-preview and of mixtral-8x7b-instruct-v0.1, together with a 1--5 GPT-4 judge
score of the weak response; 109{,}101 rows carry all three. RouteLLM's routing target is defined by
thresholding that score, so we take the construct to be $y = \mathbf{1}\{\text{score} \le 3\}$,
prevalence $0.137$. The score is a 200-character prefix TF-IDF cosine distance between the two
responses, the partial-span configuration Section~\ref{sec:partial-span} finds deployed elsewhere. The
proxy is a same-prefix surface rule: whether the two prefixes agree on three format features (list
markers, code markers, length above forty words). We audit a fixed-seed subsample of 20{,}000 rows.

$AUC(s,y) = 0.663$ and $AUC(s,z) = 0.601$: the prefix score predicts RouteLLM's
routing target \emph{better} than it predicts the same-span surface rule, so the orientation-robust
gap is negative and there is nothing for containment to account for. The three arms give
$\Delta_{|\cdot|} = -0.062$ $[-0.077,-0.048]$ same-span, $-0.050$ $[-0.064,-0.037]$ extended, and
$-0.093$ $[-0.106,-0.078]$ disjoint. The disjoint gap is reported but flagged: $16.9\%$ of complements
are empty at this span, above the $15\%$ limit of Section~\ref{sec:verdict}, and by our own
precondition it does not count. At a $10\%$ budget the queue holds 716 rows the judge called
inadequate against a base rate of $13.7\%$, a lift of $2.6\times$.

It shows the audit is not restricted to contracts we construct: on a deployed routing target's own artifacts it instantiates all three arms. The verdict is withheld by our own complement precondition, and the negative same-span gap is reported as an observation, not a clear. It does not show that RouteLLM's router is free of containment, because that router scores
the \emph{query} and never reads either response: the configuration is immune to the failure mode by
construction, a property the contract's fields make visible. Nor does it
escape the caveat of Section~\ref{sec:label-reliability}, since the construct is an automatic judge
score. Reproduction: \texttt{\detokenize{scripts/audit_routellm_deployed.py}}.

\section{What Was Fixed Before the Results Were Seen}
\label{app:preregistration}

Five of this paper's rules were added after we had looked at the rows they affect --- the resolution, orientation and minimum-count preconditions, the disjoint control and the per-contract significance bar (Table~\ref{tab:prereg}); the disjoint control followed an external objection to the first design's partial control. One reading was also withdrawn: an earlier draft read Math-Shepherd's rollout labels as a mis-declared span, which would have applied equally to every value-function method; the disjoint control cleared the span, and the paper now reports the mismatch where it is, between the released label and the construct it is consumed as (Section~\ref{sec:findings}). These changes were not pre-registered, so we record the order explicitly rather than let the finished procedure imply it was planned.

\subsection{A prospective check: a held-out contract under frozen rules}
\label{sec:holdout}
 The procedure was developed while looking at the twenty controlled contracts of Section~\ref{sec:apply}, and five of its rules were added after results were seen (Section~\ref{sec:limits2}). To measure what the frozen procedure returns on a contract that informed none of it, we fixed Algorithm~\ref{alg:rlcaudit} and every constant, pre-registered the selection rule and the contract fields, and then ran the audit once on a model pair from the panel of Appendix~\ref{app:pair-panel}, none of which contributed to any threshold. The selected contract is Qwen3.5-9B \citep{yang2025qwen3} / Gemma-2-9B-it \citep{gemmateam2024gemma} on XSTest at the same 50-character span ($n{=}450$, 79 construct positives); it passes all three preconditions. 

Its same-span agreement is large, $AUC(s,z){=}0.759$ against a construct AUC of $0.501$, so $\Delta_{\mathrm{AUC}}{=}{+}0.258$; the score ranks the off-span proxy at $0.585$; the residual is $\Delta_{\mathrm{dis}}{=}{+}0.085$ $[-0.027,+0.124]$ and it exceeds the contract's own null ($p_{95}{=}{+}0.059$, permutation $p{=}0.005$). Under the frozen rule the contract is UNDECIDABLE (equivalence): the 90\% interval on the construct AUC, $[0.438,0.562]$, does not satisfy the equivalence criterion. The abstention stands as the outcome of the held-out evaluation. Its point estimates are consistent with the no-ranking pattern, but the frozen rule does not certify it at this sample size (protocol in \texttt{\detokenize{notes/holdout_preregistration_20260908.md}}). The run is evidence about what the frozen procedure returns on one contract that shaped none of its rules; it is not a prospective validation of the procedure itself.

\begin{table}[!htbp]
\centering
\footnotesize
\setlength{\tabcolsep}{4pt}
\begin{tabular}{p{0.40\linewidth}p{0.24\linewidth}p{0.28\linewidth}}
\toprule
Element & Fixed & Effect on our own rows \\
\midrule
Contract fields, four metrics, verdict cutoffs & before any audit & --- \\
Budget $B{=}55$ and its definition & before any audit & --- \\
Survey coding scheme (A/B/C) & before reading the sample & --- \\
Survey sample (fixed seed, $n{=}40$) & before reading the sample & --- \\
Score-resolution precondition & after seeing GSM8K prefix-50; run on every contract only in the current revision & withdrew 2 rows: GSM8K at 50 characters, and OR-Bench, which had carried a NO DEMONSTRATED CONSTRUCT RANKING while the rule was applied by hand \\
Orientation precondition & after seeing GSM8K 120--200 & withdrew the raw reading of 5 of the 17 orientation-audited rows (6 rows in Appendix~\ref{app:orientation} sit below $0.5$; the sixth, GSM8K at 80 characters, is a diagnostic row outside the twenty and carries no verdict) \\
Minimum-positive-count precondition ($y^{+}{\ge}10$) & after the disjoint control & excluded 4 settings \\
Disjoint control $z^{c}$ & after review of $z^{+}$ & moved our motivating contract to non-surviving \\
Per-contract permutation null as the significance bar & after the disjoint control & one verdict depends on it, our own: CAUTION under the null, CONTAINMENT under the paired interval (Appendix~\ref{app:null}) \\
\bottomrule
\end{tabular}
\caption{Which rules were fixed before the results were seen and which were added after, with the effect of each on our own rows (Appendix~\ref{app:preregistration}).}
\label{tab:prereg}
\end{table}

\begin{table}[!htbp]
\centering
\footnotesize
\begin{tabular}{lrrrrrr}
\toprule
Rules applied & CA & SF & CT & DV & NF & UD \\
\midrule
all rules & 1 & 1 & 0 & 0 & 5 & 13 \\
no power & 1 & 1 & 0 & 0 & 7 & 11 \\
no complement & 1 & 1 & 0 & 0 & 8 & 10 \\
no resolution & 1 & 2 & 0 & 0 & 6 & 11 \\
none of the three & 1 & 2 & 1 & 0 & 11 & 5 \\
\bottomrule
\end{tabular}
\caption{The twenty contracts recounted with each post-hoc precondition switched off; the column codes are those of Table~\ref{tab:error-rates}: CA = CAUTION, SF = NO DEMONSTRATED CONSTRUCT RANKING, CT = CONTAINMENT, DV = DIVERGENCE, NF = NO FLAG, UD = UNDECIDABLE (\texttt{\detokenize{scripts/recount_without_preconditions.py}}).}
\label{tab:recount}
\end{table}

Each rule added after the fact removed evidence we had wanted to use; the permutation bar is the one whose effect is two-sided, since it turns the bootstrap reading's CONTAINMENT on the primary row into CAUTION. The additions were not pre-registered. Table~\ref{tab:recount} therefore recounts the twenty contracts with each of the three switched off, making their effect directly inspectable. Dropping all three lets two exits fire that the rules withhold, OR-Bench as NO DEMONSTRATED CONSTRUCT RANKING and XSTest 100 at 512 tokens as CONTAINMENT, and clears six rows the rules decline to judge; no verdict the paper reports changes.

\section{Making the Orientation Check a Test}
\label{app:orientation}

The orientation exit compares $AUC(s,y)$ with $0.5$, and a threshold on a point estimate fires on
$0.499$ as readily as on $0.290$. Since every row it fires on is a row whose reported gap we then
decline to use, the check has to be a test rather than a comparison. We put a bootstrap interval
(4{,}000 resamples, examples i.i.d.\ at the original $n$) on $AUC(s,y)$ for every verdict-bearing row
and require the interval to lie wholly below $0.5$.

\begin{center}\small
\begin{tabular}{lrrrl}
\toprule
Contract & $n$ & $y^{+}$ & $AUC(s,y)$ & 95\% CI \\
\midrule
AdvBench 520 & 520 & 3 & 0.290 & $[0.194, 0.451]$ \\
AdvBench 100 / 512tok & 100 & 3 & 0.296 & $[0.190, 0.403]$ \\
GSM8K, 80-char span & 1319 & 320 & 0.295 & $[0.262, 0.330]$ \\
GSM8K, 120-char span & 1319 & 320 & 0.374 & $[0.341, 0.407]$ \\
GSM8K, 160-char span & 1319 & 320 & 0.347 & $[0.314, 0.381]$ \\
GSM8K, 200-char span & 1319 & 320 & 0.358 & $[0.325, 0.391]$ \\
\bottomrule
\end{tabular}
\end{center}

\begin{figure}[t]
\centering
\includegraphics[width=0.92\linewidth]{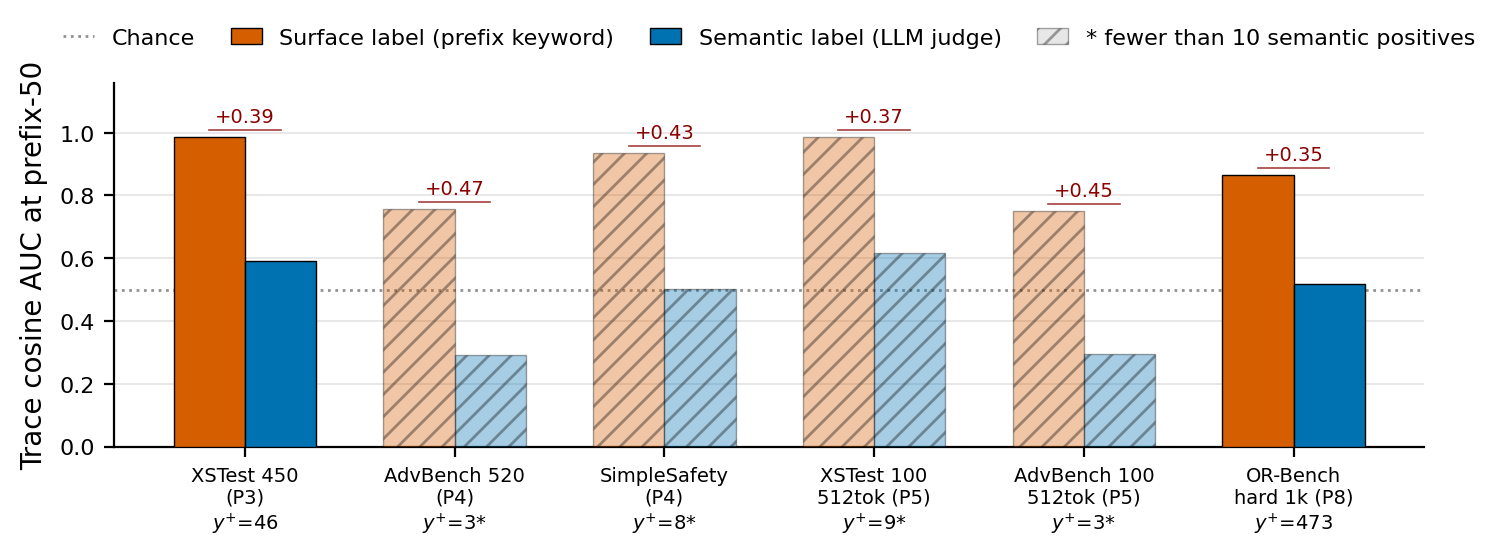}
\caption{The six refusal settings audited at prefix-50 before JailbreakBench was added (the seventh in Table~\ref{tab:all-contracts}), including the four whose semantic-positive counts fall below the floor of ten (hatched); the main text draws only the two countable settings (Figure~\ref{fig:main-audit-gap}). The hatched bars are shown here because their AdvBench rows are the gaps that invert under this appendix's orientation check. The settings draw on XSTest \citep{rottger2024xstest}, AdvBench \citep{zou2023advbench} and SimpleSafetyTests \citep{vidgen2023simplesafetytests}.}
\label{fig:main-audit-gap-full}
\end{figure}

All six rows with a point estimate below $0.5$ clear the stricter rule, so no verdict in this paper
depends on the difference. Four of the remaining rows sit on the other side of the question and are recorded for the same reason: HotpotQA at 50, 80, 160 and 200 characters returns $0.513$, $0.5277$, $0.5278$ and $0.5283$ (four decimals where three would collide) with intervals covering $0.5$, so none is shown to rank the construct. The 50-character row is the one whose off-span proxy stays ranked, the combination Table~\ref{tab:survivors} reports as a NO DEMONSTRATED CONSTRUCT RANKING exit. The two AdvBench rows clear the interval test but carry three construct positives, and Section~\ref{sec:split} excludes them on power regardless; we would not rest an orientation claim on them (their inverted bars are visible in Figure~\ref{fig:main-audit-gap-full}).

 Reproduction:
\texttt{\detokenize{scripts/orientation_significance.py}}.

\section{Verdict Cutoffs and Operating-Point Penetration}
\label{app:cutoff}

\paragraph{Sensitivity to the two thresholds.} The bands that decide a verdict apply to $\Delta_{\mathrm{dis}}$, and exactly one contract reaches them, the primary one at $\Delta_{\mathrm{dis}}{=}{+}0.101$. Over a grid crossing the DIVERGENCE band $[0.05,0.40]$ and the CAUTION band $[0.02,0.20]$ (points with the CAUTION constant above the DIVERGENCE one excluded), it is CAUTION on $50\%$ of the grid, NO FLAG on $42\%$ and DIVERGENCE on $8\%$, so the constants we fixed give this contract its \emph{least} favorable verdict on half the grid and the strongest one almost nowhere (\texttt{\detokenize{scripts/band_sweep_disjoint_nokappa.py}}; the retired $\kappa$ exit never fired on this row, so the shares equal those of the earlier three-way grid). No other verdict in Table~\ref{tab:survivors} depends on a band at all. The equivalence margin gets the same treatment: writing $\delta$ for the half-width of the band $[0.5-\delta,\,0.5+\delta]$ and testing the 90\% construct-arm intervals of the four at-chance candidates, the 50-character HotpotQA span certifies from $\delta{=}0.038$ (OR-Bench, whose verdict is withheld, from $0.040$), the 80-character span only from $0.054$, and JailbreakBench only from $0.082$. The demonstrated failure is therefore stable over $\delta\in[0.038,0.10]$, and the declared $\delta{=}0.05$ is the conservative choice that admits neither marginal row (\texttt{\detokenize{analysis_results/equivalence_band_tost.json}}). Reproduction: \texttt{\detokenize{scripts/verdict_cutoff_sensitivity.py}}.

\paragraph{When a length asymmetry reaches the queue.}
The pooled Hybrid LLM audit found coupling in the ranking but not at the operating point, which raises the
question of whether some model pairs behave differently. Splitting the 253{,}519 comparisons into the 132 pairs
with at least 400 directional judgments and repeating the queue analysis per pair, the routed queue is worse
than the pair's own base rate in 1 of 132, and that one by $+0.009$. Median penetration is $-0.132$. The
mechanism is nonetheless visible in the right direction: a pair's length asymmetry, measured as
$\mathrm{AUC}(\text{length},z)-\mathrm{AUC}(\text{length},y)$, correlates with its penetration at $r{=}0.407$.
Coupling reaches the operating point in a way that tracks the asymmetry in this data ($r{=}0.407$, one system, no causal claim), and in this deployed system the asymmetry
is not large enough for the queue to suffer. That is a quantitative statement about when the failure would
matter, and it is the form a prevalence claim would have to take. Reproduction:
\texttt{\detokenize{scripts/hybridllm_penetration_sweep.py}}.

\section{Label-Noise Robustness}
\label{app:label-noise}
Table~\ref{tab:label-noise} reports how many semantic-label changes are needed to erase the diagnostic gap, under both adversarial and random noise.

\begin{table}[!htbp]
\centering
\scriptsize
\resizebox{\linewidth}{!}{%
\begin{tabular}{lrrrrrr}
\toprule
Setting & AUC $z$ & AUC $y$ & target AUC $y$ & min pair-label flips & flip rate & random 20\% AUC $y$ \\
\midrule
XSTest 450 & 0.985 & 0.592 & 0.835 & 36/450 & 0.080 & 0.526 \\
AdvBench 520* & 0.757 & 0.290 & 0.607 & 4/520 & 0.008 & 0.494 \\
SimpleSafetyTests 100 & 0.934 & 0.501 & 0.784 & 8/100 & 0.080 & 0.499 \\
XSTest100 512 & 0.985 & 0.617 & 0.835 & 10/100 & 0.100 & 0.531 \\
AdvBench100 512* & 0.750 & 0.296 & 0.600 & 4/100 & 0.040 & 0.479 \\
OR-Bench hard 1k & 0.866 & 0.517 & 0.716 & 162/1319 & 0.123 & 0.510 \\
\bottomrule
\end{tabular}%
}
\caption{Semantic-label noise stress test. ``OR-Bench hard 1k'' is the released split name; it contains 1{,}319 prompts. The adversarial swap changes one low-score semantic positive into a negative and one high-score semantic negative into a positive, preserving semantic-positive prevalence and maximally helping the router. The target semantic AUC is $AUC_{\mathrm{surf}}-0.15$, matching the default AUC-gap screening bin. Random 20\% flips summarize 1,000 trials. Starred rows have fewer than five semantic positives (AdvBench 520: 3; AdvBench100 512: 3), so their semantic AUC and flip counts are reported for completeness and are not stable estimates.}
\label{tab:label-noise}
\end{table}

The sparse AdvBench rows are sensitive under prevalence-preserving swaps because they contain only three semantic positives; as in the main text, we use them for operating-point composition only, never as standalone semantic-ranking evidence. The better-powered XSTest and OR-Bench checks require many score-aligned label changes to erase the default AUC-gap diagnosis, while random, label-independent noise drives semantic AUC toward chance (score- or class-dependent noise need not).

\begin{table}[t]
\centering
\small
\resizebox{\linewidth}{!}{%
\begin{tabular}{llrrrr}
\toprule
Question & Check & AUC$(s,z_{\mathrm{prefix}})$ & AUC$(s,y)$ & AP $y$ & top-budget semantic \\
\midrule
Failure (primary run) & Raw prefix TF-IDF & 0.985 & 0.592 & 0.127 & 6/55 \\
Mechanism & Fixed-label opening strip & 0.645 & -- & -- & artifact signal reduced \\
Matched baseline (clean run) & Raw prefix TF-IDF & 0.956 & 0.622 & 0.198 & 11/55 \\
Intervention (clean run) & Final-span TF-IDF & 0.533 & 0.675 & 0.299 & 20/55 \\
Intervention (clean run) & Audited composite score & 0.532 & 0.715 & 0.402 & 28/55 \\
\bottomrule
\end{tabular}%
}
\caption{Mechanism and actionability summary. The proxy column is measured against the
\emph{original} artifact-bearing proxy $z_{\mathrm{prefix}}$ (prefix-50 keyword disagreement) for
every row, including the rows whose score reads a final span, so the rows are comparable. Appendix
Table~\ref{tab:marker-strip} scores the same final-span rows against their \emph{own} same-span
proxy instead, where the revised score reads 0.604; the two columns answer different questions and
are not interchangeable.}
\label{tab:mechanism-action}
\end{table}

\section{Composite Score Decomposition}

\subsection{The Repair, in Full}
\label{app:repair}
\label{sec:contract-revision}

The mismatch sits in artifact-bearing openings, so we score final-answer spans instead of prefixes and
re-run the measurement. At the same budget the matched raw-prefix baseline routes 11/55 semantic
disagreements, final-span scoring 20/55, and a composite---final-span TF-IDF plus an observable final
refusal-marker cue---28/55. A paired bootstrap over the 548 items puts the composite gain at $+17$
($[+7,+26]$, permutation $p<10^{-4}$) against $+9$ ($[0,+18]$) for the final-span step alone, so the
composite carries the repair claim (Appendix Table~\ref{tab:mechanism-action}). Measured against the original artifact-bearing prefix proxy, coupling collapses from 0.956 to 0.532 across the same rows while semantic alignment rises.

\paragraph{What the repair is and is not.} The claim is confined to routed composition: the paired
semantic AUC delta covers zero ($+0.093$ $[-0.011,+0.194]$), so this validates a less-coupled contract
and not a deployable router. The marker cue is itself a keyword feature, so the repair absorbs a
better-aligned proxy into the score, as the contract predicts. It cannot do the job alone: 133 of 548
items tie at its top value against 55 budget slots, so its routed count is a draw from the tie
(20.7 expected, $[15,26]$).

Auditing the repaired contract settles the objection that it is just another keyword score. Against
the original artifact-bearing proxy it is decoupled: prefix-keyword alignment falls from $0.956$ to
chance, semantic alignment rises from $0.622$ to $0.715$, and $\Delta_{\mathrm{AUC}}$ moves from
$+0.334$ to $-0.183$. Our own rule has a consequence here: this baseline's reported agreement is itself
mostly containment ($\Delta_{\mathrm{ext}}{=}{-}0.033$ on the clean run), so it clears before the
repair is applied, and the repair claim is about routed composition, not about turning a DIVERGENCE into
an NO FLAG. The same instrument marks the repair's limit: the marker is now a component of the score, so scored against
it the composite returns $\Delta_{\mathrm{AUC}}{=}0.000$ with $\Delta_{\mathrm{AP}}{=}+0.195$, a CAUTION-sized gap on the average-precision reading; the deciding rule of Section~\ref{sec:verdict} reads $\Delta_{\mathrm{dis}}$ against the contract's null instead. Absorbing an indicator uses it up as an auditing target. The mirror intervention separates the two spans in the other direction as well: stripping marker sentences leaves proxy alignment intact ($-0.008$) whereas stripping openings collapsed it by $0.340$ (Appendix~\ref{app:agcr}).

\label{app:agcr}

The composite score is meant as evidence of actionability, so Table~\ref{tab:agcr-ablation} decomposes the components summarized in the main text. On the clean run, final-span TF-IDF implements the audit's span relocation step; the observable final-marker signal implements the audit's cheap construct-indicator step; the fixed composite default combines the two without fitting to semantic outcomes. The learned logistic row is an out-of-fold diagnostic upper bound. Table~\ref{tab:agcr-sensitivity} reports the corresponding fixed-weight sensitivity band. The default $\lambda{=}0.7$ was fixed before those numbers were read and is not a tuned value: $\lambda{=}0.5$ scores better on every reported quantity (AUC $0.744$ against $0.715$, $31/55$ routed against $28/55$), so tuning would have produced a different default.

\begin{table}[!htbp]
\centering
\scriptsize
\resizebox{\linewidth}{!}{%
\begin{tabular}{lrrrl}
\toprule
Score & AUC & AP & top-10\% sem & Role \\
\midrule
Raw prefix TF-IDF & 0.622 & 0.198 & 11/55 & artifact-bearing baseline \\
Final TF-IDF only & 0.675 & 0.299 & 20/55 & construct-bearing span \\
Final marker only & 0.728 & 0.291 & 20.7 [15, 26]$^\dagger$ & cheap observable indicator \\
Composite, $\lambda=0.7$ & 0.715 & 0.402 & 28/55 & audited composite \\
Learned logistic OOF & 0.741 & 0.421 & 29/55 & upper-bound diagnostic \\
\bottomrule
\end{tabular}%
}
\caption{Audited composite score decomposition on the clean run (Qwen3.5-2B/Gemma-4-E2B-it); the learned row is out-of-fold and diagnostic only. $^\dagger$The marker score is binary and leaves 133 items tied for $B{=}55$ slots, so its routed count is an expectation with a 95\% tie interval over $20{,}000$ tie-breaks rather than a single draw (\texttt{\detokenize{scripts/marker_tie_sensitivity.py}}).}
\label{tab:agcr-ablation}
\end{table}

\begin{table}[!htbp]
\centering
\scriptsize
\resizebox{\linewidth}{!}{%
\begin{tabular}{lccc}
\toprule
Pair & Final TF-IDF AUC/AP, top sem & Composite $\lambda\in[0.6,0.8]$ AUC/AP range & Composite top sem range \\
\midrule
Qwen2B/GemmaE2B (clean run) & 0.675 / 0.299, 20/55 & 0.701--0.730 / 0.380--0.422 & 25--31/55 \\
Qwen9B/Gemma9B & 0.664 / 0.316, 17/45 & 0.687--0.716 / 0.342--0.371 & 17--19/45 \\
Qwen9B/Mistral7B & 0.649 / 0.234, 13/45 & 0.674--0.701 / 0.296--0.335 & 15--19/45 \\
Gemma9B/Mistral7B & 0.707 / 0.376, 18/45 & 0.738--0.766 / 0.419--0.454 & 20--22/45 \\
Qwen9B/Llama8B & 0.559 / 0.121, 4/45 & 0.580--0.602 / 0.145--0.165 & 7--10/45 \\
Gemma9B/Llama8B & 0.633 / 0.283, 13/45 & 0.667--0.711 / 0.338--0.386 & 16--22/45 \\
Mistral7B/Llama8B & 0.701 / 0.320, 18/45 & 0.719--0.736 / 0.357--0.372 & 19/45 \\
Qwen8B/Llama8B & 0.664 / 0.226, 11/45 & 0.692--0.719 / 0.273--0.313 & 16--20/45 \\
Qwen8B/Qwen9B & 0.631 / 0.181, 8/45 & 0.647--0.664 / 0.301--0.396 & 11--14/45 \\
\bottomrule
\end{tabular}%
}
\caption{Fixed composite weight sensitivity from cached final-channel runs, over $\lambda\in[0.6,0.8]$. The first row is the 2B/E2B contract of Section~\ref{sec:dissection}; the rest are 9B-family panel pairs, named as in Appendix Table~\ref{tab:pair-panel-full}. The Llama rows sit outside the ten-pair panel, since their final-span contract is not instantiated (Appendix~\ref{app:pair-panel}).}
\label{tab:agcr-sensitivity}
\end{table}

\paragraph{Symmetric span intervention on the revised contract.}
The motivating contract is diagnosed by suppressing openings with labels held fixed. The revised contract is
tested the same way: refusal-marker sentences are removed from the final span, all labels stay fixed, and the
score is recomputed. Because that strip also removes text (77.2\% of characters survive; 11.2\% of responses
become empty), a volume control removes randomly chosen sentences to the same character budget, averaged over
20 seeds. Table~\ref{tab:marker-strip} reports both.

\begin{table}[!htbp]
\centering
\small
\resizebox{\linewidth}{!}{%
\begin{tabular}{lrrrl}
\toprule
Score on the clean run ($n{=}548$) & AUC proxy & AUC semantic & sem@$B{=}55$ & Role \\
\midrule
Final-span TF-IDF & 0.604 & 0.675 [0.606, 0.741] & 20/55 & revised contract \\
Final-span TF-IDF, marker sentences removed & 0.595 & 0.601 [0.535, 0.667] & 8/55 & span intervention \\
Final-span TF-IDF, random sentences removed & -- & 0.548 [0.538, 0.560] & 1/55 & volume control, 20 seeds \\
Final marker alone & 1.000 & 0.728 [0.668, 0.783] & 20.7 [15, 26]$^\dagger$ & cheap observable indicator \\
\bottomrule
\end{tabular}%
}
\caption{Marker-strip intervention on the revised contract, with a volume-matched random-strip control (AUC 95\% bootstrap intervals in brackets; sem@$B$ = construct disagreements in the top-55 queue). $^\dagger$Tie-aware, as in Table~\ref{tab:agcr-ablation}: the binary marker leaves 133 items tied for 55 slots, so its count is reported as an expectation with a 95\% tie interval rather than as a single draw.}
\label{tab:marker-strip}
\end{table}

The strip is proxy-neutral: proxy alignment moves by $-0.008$, where suppressing openings in
the motivating contract moved it by $-0.340$. And the semantic loss it causes ($-0.074$) is smaller than the loss
from deleting an equal volume at random ($-0.127$), so the final-span score's semantic signal is distributed
across the answer rather than concentrated in marker sentences. This does not make the marker uninformative (alone it reaches semantic AUC 0.728), but it does separate the two spans by measurement rather than assertion:
the opening span fails the intervention on the proxy side, the final span does not. Re-running the judge on the marker-stripped responses leaves its decisions almost unchanged (side-level agreement $0.980$, $\kappa{=}0.961$; pair-level $0.970$), so the decision does not live in the marker sentence and the marker is a symptom of a construct-bearing span rather than an artifact the score exploits. One exclusion is not neutral: stripping empties 123 of 1{,}096 sides, exactly the shortest refusals where the marker is the whole response, so the claim covers the 89\% that retain text. \paragraph{Keyword versus span in the repair.}
The composite absorbs a keyword cue, so we tested whether the final span carries construct evidence outside that vocabulary. Removing the twelve refusal terms from the final-answer text and fitting an out-of-fold logistic model on the masked text recovers each model's judged refusal decision at AUC 0.982 and 0.987. The span therefore carries the decision without the keyword vocabulary.

Substituting the masked cue for the marker raises the routed count to $34/55$ and the masked cue alone reaches $40/55$, but both are fitted to the judge labels they are scored against, so we report them as diagnostic upper bounds and do not use them.

The circularity is removed by taking the cue from an instrument we neither built nor trained. A public refusal classifier (\texttt{\detokenize{protectai/distilroberta-base-rejection-v1}}) scores each final answer, and the cue is the absolute difference between the two scores. It is a working instrument here, recovering the judged side decision at AUC 0.871 and 0.968, and it is not a restatement of the keyword: the two cues correlate at $r{=}0.259$. Substituted into the composite it routes 29/55, against 28/55 for the marker cue and 20/55 for the final span alone, with semantic AUC 0.748. The repair therefore reproduces with a cue that is independent of our labels, which is what the circularity objection asks for. A first attempt using a public NLI model instead failed as an instrument rather than as a hypothesis---entailment probabilities were near zero throughout and predicted the side decision at AUC 0.535 and 0.547, against 0.835 for the bare keyword, and we report it here. Reproduction: \texttt{\detokenize{scripts/marker_strip_symmetry.py}}, \texttt{\detokenize{scripts/judge_marker_stripped.py}}, \texttt{\detokenize{scripts/nli_cue_repair.py}}, and \texttt{\detokenize{scripts/external_refusal_cue.py}}, with results under \texttt{\detokenize{analysis_results/}}.

\section{Auditing a Published Recipe (FrugalGPT-style Correctness Routing)}
\label{app:frugal}

FrugalGPT's scorer reads the query and the generated answer, and the correctness label it is validated against
is a string match on that same answer. Score and proxy therefore share a span by design, which is the stated
precondition for representation-label coupling. We test whether the precondition is enough. Table~\ref{tab:frugal-audit} gives the per-contract numbers.

The instantiation uses our judged HotpotQA and GSM8K runs (Qwen3.5-2B and Gemma-4-E2B-it pooled, $n{=}500$ per
task). The score is an out-of-fold logistic model over TF-IDF features of the question and the generated answer,
trained to predict the string-match label; the gold answer is never shown to the scorer. The proxy $z$ is the
recipe's own label, gold-answer-string-in-trace. The construct $y$ is independent judge adjudication of
correctness on the same trace. Escalation routes the lowest-scoring 10\%.

\begin{table}[!htbp]
\centering
\small
\resizebox{\linewidth}{!}{%
\begin{tabular}{lrrrrrl}
\toprule
Recipe instantiation & $n$ & $\kappa(z,y)$ & AUC $z$ & AUC $y$ & routed wrong & Outcome \\
\midrule
HotpotQA multi-hop & 500 & 0.529 & 0.754 & 0.706 & 35/50 (base 0.268) & no flag \\
GSM8K reasoning & 500 & 0.747 & 0.894 & 0.902 & 50/50 (base 0.380) & no flag \\
\bottomrule
\end{tabular}%
}
\caption{RLC-Audit of a FrugalGPT-style contract, in which score and proxy read the same span by construction (routed wrong = judged-incorrect answers in the 10\% escalation queue, with the dataset base rate in parentheses). The scorer reads the whole answer, so no off-span re-read exists and Algorithm~\ref{alg:rlcaudit} returns UNDECIDABLE with a required field missing; the outcome column is the same-span check, not a verdict.}
\label{tab:frugal-audit}
\end{table}

Neither task shows the failure. The paired AUC gaps are $+0.048$ $[-0.009,+0.103]$ and $-0.008$
$[-0.033,+0.016]$, both covering zero, and the escalation queues are strongly enriched for judged-incorrect
answers relative to base rate. This does not clear the released FrugalGPT system, which we did not run, and it
does not weaken the refusal result, which is a different contract. What it establishes is a boundary the paper needs: a shared span is
necessary for coupling but not sufficient. Correctness depends on the answer text, so a proxy computed
from the answer text is a coarse but real measurement of the construct, whereas a refusal-keyword
proxy computed on an opening prefix is not, and the audit separates the two cases---the distinction
Section~\ref{sec:contract-revision} measures with span interventions. Reproduction: \texttt{\detokenize{scripts/audit_frugalgpt_recipe.py}}; results in
\texttt{\detokenize{analysis_results/frugalgpt_recipe_audit.json}}.

\section{The Audit Procedure and Its Declared Branches}
\label{app:algorithm}

\begin{figure}[!t]
\centering
\includegraphics[width=\linewidth]{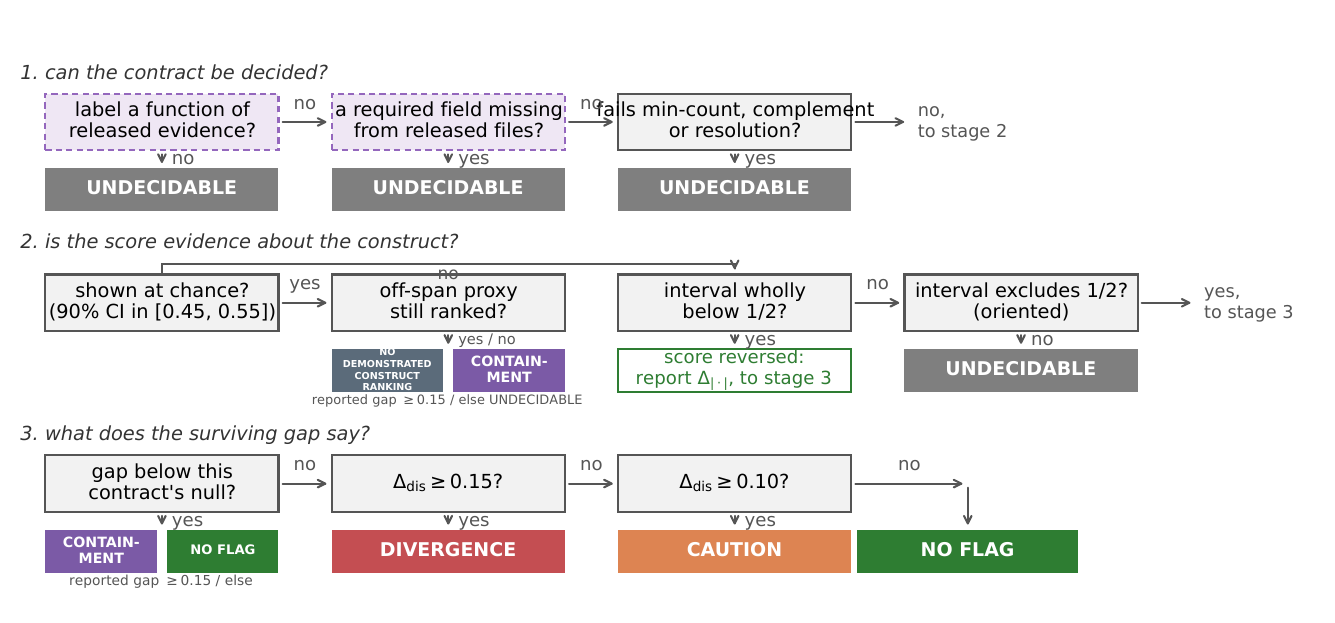}
\caption{Overview of the audit logic. The three stages are: can the contract be decided, is the score evidence about the construct, and what the surviving gap says. Solid boxes are computations; the two dashed boxes are auditor judgments. Algorithm~\ref{alg:rlcaudit} evaluates the stages in this order.}
\label{fig:verdict-tree}
\end{figure}

This appendix states the decision procedure Section~\ref{sec:verdict} summarizes.
Off-span agreement $\kappa(z^{c},y)$ is reported beside every verdict and decides none: a high value says the off-span proxy and the construct largely coincide, which bounds what a residual gap can mean, and nothing more.

\begin{table}[!ht]
\centering
\footnotesize
\setlength{\tabcolsep}{4pt}
\begin{tabular}{p{0.19\linewidth}p{0.40\linewidth}p{0.15\linewidth}p{0.20\linewidth}}
\toprule
Evidence pattern & What the AUCs show & Exit & Follow-up \\
\midrule
Containment-compatible: the observed gap is removed by the disjoint control & $\Delta_{\mathrm{AUC}}$ large while $\Delta_{\mathrm{dis}}$ falls below the contract's null; or the construct AUC is at chance, $z^{c}$ is not ranked either, and $\Delta_{\mathrm{AUC}}\geq 0.15$ & CONTAINMENT & move the span \\
Residual divergence, consistent with proxy--construct mismatch or non-construct off-span spillover & $\Delta_{\mathrm{dis}}$ above the null and $\geq 0.15$ ($\geq 0.10$: CAUTION), with the construct AUC ranked & DIVERGENCE, CAUTION & inspect the proxy label and off-span spillover; no unique repair identified \\
No demonstrated construct ranking & construct AUC at chance (90\% interval inside $[0.45,0.55]$) while $z^{c}$ is still ranked & NO DEMONSTRATED CONSTRUCT RANKING (at this span) & stop treating the score as evidence at this span \\
No flag under this audit & $\Delta_{\mathrm{dis}}$ below the contract's null, or above it but under the CAUTION constant; $\kappa(z^{c},y)$ reported beside it & NO FLAG & none \\
Cannot be decided & a precondition fails, the construct interval covers $0.5$ without being contained within the equivalence band, or neither target shows a signal & UNDECIDABLE & release more, or re-instantiate \\
\bottomrule
\end{tabular}
\caption{Three flagged evidence patterns and two non-flag outcomes, the evidence that identifies each, the exit it takes and the follow-up it motivates. The exit names are summaries of this audit's own measurements; NO DEMONSTRATED CONSTRUCT RANKING in particular is a statement about the span tested, not about the score at every span.}
\label{tab:exits}
\end{table}

The exits name evidence patterns, not verified causes: DIVERGENCE names a residual proxy--construct divergence consistent with a label mismatch or with off-span spillover that does not pass through the construct, and Section~\ref{sec:limits} says why the control cannot separate that from spillover. The two at-chance exits differ in what the off-span proxy shows. If the score still ranks the proxy read beyond its span, it is tracking a surface feature that spills past the span while its construct AUC lies inside the declared equivalence margin: NO DEMONSTRATED CONSTRUCT RANKING. The exit is about this score's ordering within that margin, not a claim that the score carries no information about the construct. It is also a stop rule for the tested span rather than a claim about other spans, since ranking $z^{c}$ shows that this span predicts later surface text and not that no span could rank the construct (Section~\ref{sec:scorefailure}). If it ranks neither the off-span proxy nor the construct, and the same-span reported gap was large ($\Delta_{\mathrm{AUC}}\geq 0.15$), the pattern is compatible with a reported agreement produced by this span alone, and the contract can be re-instantiated at another: CONTAINMENT. Without a large reported gap there is nothing for containment to have produced, and the contract is UNDECIDABLE (no signal). ``Still ranked'' is read at $|AUC(s,z^{c})-0.5|\geq 0.10$ on the point estimate; Section~\ref{sec:scorefailure} reports the intervals behind the two rows it applies to.

Two exits in Figure~\ref{fig:verdict-tree} are auditor judgments rather than computations, drawn dashed (their reproducibility is bounded only weakly, on eight items, Appendix~\ref{app:inter-auditor}) --- whether the released label can be a function of the released evidence, and whether a required contract field is missing from the released files; neither can move a verdict toward DIVERGENCE. Table~\ref{tab:prereg} records which rules were fixed before the results were seen and which five were added after.

\begin{table}[!t]
\centering
\footnotesize
\setlength{\tabcolsep}{4pt}
\begin{tabular}{p{0.30\linewidth}p{0.24\linewidth}p{0.13\linewidth}p{0.25\linewidth}}
\toprule
Constant & Value & Fixed & Sensitivity \\
\midrule
Equivalence band, ``at chance'' & 90\% interval in $[0.45,0.55]$ & before audit & Appendix~\ref{app:cutoff} \\
Magnitude bands on $\Delta_{\mathrm{dis}}$ & CAUTION $0.10$; \mbox{DIVERGENCE} $0.15$ & before audit & Appendices~\ref{app:cutoff}, \ref{app:semisynthetic} \\
Off-span agreement $\kappa(z^{c},y)$ & reported; no longer an exit & before audit (as an exit, retired here) & Appendix~\ref{app:error-rates} (high-$\kappa$ case) \\
Off-span proxy ``still ranked'' & $|AUC(s,z^{c})-0.5|\geq 0.10$ & after the control & Section~\ref{sec:verdict} \\
Significance bar & contract's permutation null, 95th pct. & after the control & Appendix~\ref{app:null} \\
Minimum-positive-count precondition & $\geq 10$ positives on any binary AUC target (construct, and each proxy reading) & after the control & Appendix Table~\ref{tab:recount} \\
Complement precondition & non-empty on $\geq 85\%$ of sides & after the control & Table~\ref{tab:recount}; Section~\ref{sec:findings} \\
Resolution precondition & modal score share $\leq 0.5$ & after GSM8K-50 & Table~\ref{tab:recount} \\
\bottomrule
\end{tabular}
\caption{Every constant the summaries depend on, when it was fixed relative to the results, and where its sensitivity is reported.}
\label{tab:constants}
\end{table}

\begin{algorithm}[!htbp]
\centering
\begin{minipage}{0.95\linewidth}
\small
\begin{enumerate}
\setlength{\itemsep}{0pt}
\item \textbf{Align the contract.} Join scores, score spans, proxy labels, semantic labels, model
pair, and route budget on the same examples.
\item \textbf{Check that the score can rank, and which way it points.} If one score value carries over half the items the contract is withdrawn under the resolution precondition, a scope restriction rather than a proof that the score cannot rank (Section~\ref{sec:verdict}) (GSM8K's 50-character span: twelve distinct values over 1{,}319 items, modal mass $0.634$, its eight proxy positives on one value). Proxy positives sharing a value is not by itself degeneracy: on the motivating contract all 13 proxy positives sit at the score's maximum while the score takes 43 distinct values with modal mass $0.436$, which is the containment signature Section~\ref{sec:mechanism} dissects, not a failure to rank. If the bootstrap interval on $AUC(s,y)$ lies wholly
below $0.5$ the score is reversed with respect to the construct rather than uninformative about it;
report $\Delta_{|\cdot|} = |AUC(s,z)-0.5| - |AUC(s,y)-0.5|$ and let the verdict follow the smaller
gap. ``Ranks no better than chance'' must be shown, not assumed: the NO DEMONSTRATED CONSTRUCT RANKING exit requires the 90\% interval on $AUC(s,y)$ to sit inside the equivalence band $[0.45,0.55]$ (TOST at $\alpha{=}0.05$) with the off-span proxy still ranked; an interval that covers $0.5$ without being contained within the equivalence band leaves the contract undecided at the design's power (Appendix~\ref{app:orientation}).
\item \textbf{Measure both targets, and the containment controls.} Compute AUC/AP and budget
precision against $z$ and $y$, and recompute the proxy away from the scored span to obtain
$\Delta_{\mathrm{ext}}$ and $\Delta_{\mathrm{dis}}$.
\item \textbf{Measure construct separation.} Report prevalence and $\kappa(z,y)$, together with the
largest $\kappa$ those two margins permit.
\item \textbf{Inspect routed composition.} Count semantic disagreements, surface-only disagreements,
agreed refusals, and agreed compliances in $R_B(s)$.
\item \textbf{Localize and re-audit.} Stress artifact-bearing spans; any intervention is a new
contract, including any follow-up intervention motivated by the audit.
\item \textbf{Return a verdict.} Apply the exits of Figure~\ref{fig:verdict-tree} in order. As one line: a contract failing a precondition is UNDECIDABLE; else if the 90\% interval on $AUC(s,y)$ lies inside $[0.45,0.55]$, NO DEMONSTRATED CONSTRUCT RANKING when $|AUC(s,z^{c})-0.5|\geq 0.10$, CONTAINMENT when it is not and $\Delta_{\mathrm{AUC}}\geq 0.15$, and UNDECIDABLE (no signal) otherwise; else if the 95\% interval on $AUC(s,y)$ lies wholly below $0.5$, re-orient onto $\Delta_{|\cdot|}$ and continue; else if the 95\% interval covers $0.5$, UNDECIDABLE; else if $\Delta_{\mathrm{dis}}$ does not exceed the contract's null 95th percentile, CONTAINMENT when $\Delta_{\mathrm{AUC}}\geq 0.15$ and NO FLAG otherwise; else DIVERGENCE if $\Delta_{\mathrm{dis}}\geq 0.15$, CAUTION if $\geq 0.10$, NO FLAG below.
\end{enumerate}
\end{minipage}
\caption{RLC-Audit, the procedure whose exits Figure~\ref{fig:verdict-tree} draws.}
\label{alg:rlcaudit}
\end{algorithm}

\paragraph{A documentation-consistency check on the two declared branches.}

\label{app:inter-auditor}

Two exits in Figure~\ref{fig:verdict-tree} require auditor judgment. We have not measured human inter-auditor reliability. As a documentation-consistency check only, we wrote a neutral description of each artifact from its released documentation, paired it with the branch question, and put eight pre-specified decisions to an LLM auditor (\texttt{gpt-4.1-mini}) that never sees our verdicts or this paper; it reproduced seven of the eight, and the remaining disagreement identified an error in our key, which corrected gives eight of eight. Both numbers are reported, since the correction was made after seeing the answer. We treat this only as evidence that the branch questions are answerable from the released documentation, not as a reliability estimate. The sampling design, the constant-answer baseline, the per-item results and all raw responses are in \texttt{\detokenize{scripts/inter_auditor_reliability.py}}.

\section{The Containment Controls, Contract by Contract}
\label{app:span-independent}

Table~\ref{tab:all-contracts} lists all twenty controlled contracts with the precondition that removes each and the verdict it receives; the tables below then give both containment controls for every contract that supports them.

\begin{table}[!htbp]
\centering\footnotesize
\setlength{\tabcolsep}{3pt}
\begin{tabular}{lrrrll}
\toprule
Contract & $n$ & $y^{+}$ & $\Delta_{\mathrm{dis}}$ & precondition & verdict \\
\midrule
XSTest 450 (primary run) & 450 & 46 & $+0.101$ & --- & CAUTION \\
AdvBench 520 & 520 & 3 & --- & min-count & UNDECIDABLE \\
SimpleSafety 100 & 100 & 8 & --- & min-count & UNDECIDABLE \\
XSTest 100 / 512tok & 100 & 9 & --- & min-count & UNDECIDABLE \\
AdvBench 100 / 512tok & 100 & 3 & --- & min-count & UNDECIDABLE \\
OR-Bench hard 1k & 1319 & 473 & --- & resolution & UNDECIDABLE \\
JailbreakBench & 200 & 27 & $+0.245$ & --- & UNDECIDABLE (equivalence) \\
\midrule
GSM8K, 50-char & 1319 & 320 & --- & resolution & UNDECIDABLE \\
GSM8K, 120-char & 1319 & 320 & $-0.001$ & --- & NO FLAG$^{\mathrm{r}}$ \\
GSM8K, 160-char & 1319 & 320 & $-0.019$ & --- & NO FLAG$^{\mathrm{r}}$ \\
GSM8K, 200-char & 1319 & 320 & $-0.027$ & --- & NO FLAG$^{\mathrm{r}}$ \\
GSM8K, 300-char & 1319 & 320 & $-0.045$ & --- & NO FLAG \\
GSM8K, 600-char & 1319 & 320 & $-0.083$ & --- & NO FLAG \\
\midrule
HotpotQA, 50-char & 2000 & 436 & $+0.121$ & --- & \textbf{NO DEMONSTRATED CONSTRUCT RANKING} \\
HotpotQA, 80-char & 2000 & 436 & $+0.073$ & --- & UNDECIDABLE (equivalence) \\
HotpotQA, 120-char & 2000 & 436 & --- & complement & UNDECIDABLE \\
HotpotQA, 160-char & 2000 & 436 & --- & complement & UNDECIDABLE \\
HotpotQA, 200-char & 2000 & 436 & --- & complement & UNDECIDABLE \\
HotpotQA, 300-char & 2000 & 436 & --- & complement & UNDECIDABLE \\
HotpotQA, 600-char & 2000 & 436 & --- & complement & UNDECIDABLE \\
\bottomrule
\end{tabular}
\caption{All twenty audited contracts with the precondition that removes each and the verdict it receives. $n$ is the number of paired prompts and $y^{+}$ the construct positives. $\Delta_{\mathrm{dis}}$ is quoted only where the preconditions pass; intervals for the surviving rows are in Table~\ref{tab:survivors}. The equal $n{=}1{,}319$ on the OR-Bench and GSM8K rows is a coincidence of the two sources. $^{\mathrm{r}}$~The construct AUC is below chance, so the row is re-oriented and its gap is $\Delta_{|\cdot|}$ (Section~\ref{sec:verdict}). Every gap column in this table and in Table~\ref{tab:survivors} is orientation-robust.}
\label{tab:all-contracts}
\end{table}

Sections~\ref{sec:split} and~\ref{sec:failure} report $\Delta_{\mathrm{ext}}$ for individual contracts;
Table~\ref{tab:span-independent} gives every contract the control can be instantiated on that passes the resolution precondition. It holds the
score, the construct, the model pair and the budget fixed and changes only where the proxy rule is
evaluated: $z$ is the proxy as the contract defines it, on the span the score reads, and $z^{+}$
is the same rule applied to the complete output. The full-output proxy includes the scored span and is therefore only a partial control; $y$ is read over the complete output. The residual under the disjoint reading is the part of the agreement that survives once the proxy is read strictly off the scored span. The comparison holds the score fixed and varies the measurement: the re-read proxy is a different measurement from $z$ even under the same rule, which is why its null is estimated on the contract rather than assumed.

\begin{table}[!htbp]
\centering
\small
\setlength{\tabcolsep}{4pt}
\begin{tabular}{lrrrrrl}
\toprule
Contract & $n$ & $y^{+}$ & $\Delta_{|\cdot|}$ & $\Delta_{\mathrm{ext}}$ & $\Delta_{\mathrm{dis}}$ & 95\% CI on $\Delta_{\mathrm{dis}}$ \\
\midrule
XSTest 450 (primary run) & 450 & 46 & +0.393 & +0.263 & +0.101 & $[-0.059,+0.255]$ \\
AdvBench 520\textsuperscript{p} & 520 & 3 & +0.047 & +0.071 & +0.040 & $[-0.077,+0.217]$ \\
SimpleSafety 100\textsuperscript{p} & 100 & 8 & +0.433 & +0.341 & +0.056 & $[-0.080,+0.281]$ \\
XSTest 100 / 512tok\textsuperscript{p} & 100 & 9 & +0.369 & -0.054 & -0.058 & $[-0.168,+0.091]$ \\
AdvBench 100 / 512tok\textsuperscript{p} & 100 & 3 & +0.045 & -0.161 & +0.015 & $[-0.156,+0.182]$ \\
OR-Bench hard 1k & 1319 & 473 & +0.348 & +0.252 & +0.248 & $[+0.153,+0.333]$ \\
JailbreakBench & 200 & 27 & +0.310 & +0.246 & +0.245 & $[+0.052,+0.366]$ \\
\midrule
GSM8K, 120-char span & 1319 & 320 & -0.058 & -0.001 & -0.001 & $[-0.030,+0.027]$ \\
GSM8K, 160-char span & 1319 & 320 & -0.144 & -0.019 & -0.019 & $[-0.047,+0.007]$ \\
GSM8K, 200-char span & 1319 & 320 & -0.116 & -0.027 & -0.027 & $[-0.057,+0.002]$ \\
GSM8K, 300-char span & 1319 & 320 & -0.055 & -0.046 & -0.045 & $[-0.073,-0.015]$ \\
GSM8K, 600-char span & 1319 & 320 & -0.102 & -0.059 & -0.083 & $[-0.113,-0.050]$ \\
HotpotQA, 50-char span & 2000 & 436 & +0.184 & +0.016 & +0.121 & $[+0.082,+0.150]$ \\
HotpotQA, 80-char span & 2000 & 436 & +0.116 & -0.008 & +0.073 & $[+0.037,+0.109]$ \\
HotpotQA, 120-char span\textsuperscript{c} & 2000 & 436 & +0.015 & -0.007 & +0.009 & $[-0.032,+0.048]$ \\
HotpotQA, 160-char span\textsuperscript{c} & 2000 & 436 & +0.013 & -0.011 & +0.010 & $[-0.029,+0.045]$ \\
HotpotQA, 200-char span\textsuperscript{c} & 2000 & 436 & +0.023 & -0.020 & +0.026 & $[-0.012,+0.061]$ \\
HotpotQA, 300-char span\textsuperscript{c} & 2000 & 436 & +0.017 & -0.018 & +0.044 & $[+0.006,+0.082]$ \\
HotpotQA, 600-char span\textsuperscript{c} & 2000 & 436 & +0.030 & -0.004 & +0.038 & $[-0.002,+0.078]$ \\
\bottomrule
\end{tabular}
\caption{What survives when the proxy moves off the scored span, under a partial control
($\Delta_{\mathrm{ext}}$, proxy re-read over the complete output) and a disjoint one
($\Delta_{\mathrm{dis}}$, proxy read only beyond the scored span). Upper block: the seven audited
refusal settings, scored by a prefix-50 TF-IDF distance against a refusal-keyword proxy. Lower block: the correctness contracts at each span passing the resolution precondition. On the shorter GSM8K spans the two controls coincide because the gold-answer proxy fires on the same 237 items whether it is read over the complete output or only beyond the span; they separate at 600 characters, where the span begins to contain the answer. $y^{+}$ counts construct
positives. \emph{All three} gap columns are orientation-robust, $|AUC(s,\cdot)-0.5|-|AUC(s,y)-0.5|$, so a
construct ranking below chance cannot inflate a gap; intervals are paired bootstraps at 2{,}000
resamples. \textsuperscript{p} marks rows with fewer than ten construct positives and
\textsuperscript{c} rows whose complement is empty on more than 15\% of sides; both are shown and both
are excluded from every count in the text. Reproduction:
\texttt{\detokenize{scripts/complement_span_proxy_control.py}}.}
\label{tab:span-independent}
\end{table}

\subsection{Two Further Controls}
\label{app:added-controls}

\paragraph{Question overlap does not appear to be what remains.} We tested question-overlapping lexical content as the most natural candidate for the residual: replacing every content word a response shares with its question by a same-length filler, so the 50-character boundary is untouched, changes the off-span proxy AUC by $-0.016$ $[-0.040,+0.007]$ against $+0.016$ $[-0.002,+0.033]$ when the same number of non-question content words is masked instead. The masking does not explain the remaining association; its source remains unidentified, and residual template text, topic and difficulty, and answer length are among the candidates these designs do not separate. The same-span proxy AUC \emph{rises} under the masking, $+0.045$ $[+0.021,+0.070]$ against $+0.005$ $[-0.018,+0.028]$ for the control, which is what one expects if the paraphrase is content the two prefixes share rather than the feature that separates them. Both HotpotQA interventions are reproduced by \texttt{\detokenize{scripts/opening_strip_hotpot.py}} (opening suppression) and \texttt{\detokenize{scripts/question_paraphrase_intervention.py}} (question-overlap masking).
\begin{table}[!t]
\centering
\small
\resizebox{\linewidth}{!}{%
\begin{tabular}{lrrll}
\toprule
Contract & $\Delta_{\mathrm{ext}}$ & $\Delta_{\mathrm{dis}}$ & under $\Delta_{\mathrm{ext}}$ & under $\Delta_{\mathrm{dis}}$ \\
\midrule
XSTest 450 (primary run) & $+$0.263 & $+$0.101 & DIVERGENCE & \textbf{CAUTION} \\

JailbreakBench & $+$0.246 & $+$0.245 & DIVERGENCE & UNDECIDABLE (equivalence) \\
HotpotQA, 50-char span & $+$0.016 & $+$0.121 & NO FLAG & \textbf{NO DEMONSTRATED CONSTRUCT RANKING} \\
HotpotQA, 120--600-char spans (5 rows) & ${\approx}0$ & --- & NO FLAG & UNDECIDABLE \\
\bottomrule
\end{tabular}%
}
\caption{Verdict under the partial control and under the disjoint one, on every contract where the verdict differs. $\Delta_{\mathrm{ext}}$ is the proxy re-read over the complete output,
$\Delta_{\mathrm{dis}}$ read only beyond the scored span; a dash means the contract does not reach
that control. Bold marks the verdicts the paper reports as its own results
(Appendix~\ref{app:added-controls}; \texttt{\detokenize{scripts/verdict_rule_and_ablation.py}}).}
\label{tab:which-control}
\end{table}

\paragraph{What only the disjoint control does.} Reporting $AUC(s,y)$ and $\kappa(z,y)$ beside the proxy AUC already detects a poor stand-in, and under the same preconditions that cheap rule flags the rows we flag plus five GSM8K spans (Appendix~\ref{app:added-controls}). It cannot say which pattern produced the gap. Run without the preconditions it would also keep every row the minimum-count and orientation checks withdraw. It cannot separate the two exits whose construct AUC is identical, NO DEMONSTRATED CONSTRUCT RANKING and CONTAINMENT, whose prescriptions are opposite: the 50-character HotpotQA span is the case in point, where the cheap rule raises a label failure and the repair that licenses (substitute another proxy) would keep a score that ranks answer strings in an opening and not correctness. 

Against the partial control the disjoint one does a further thing: it reverses a reading. On the 50- and 80-character HotpotQA rows, gaps that $\Delta_{\mathrm{ext}}$ dismisses as containment survive under $\Delta_{\mathrm{dis}}$. On the 50-character row that takes the verdict from NO FLAG to NO DEMONSTRATED CONSTRUCT RANKING, and on the 80-character row the reading reverses while the verdict is set at the equivalence step either way. On this corpus, then, the disjoint control demotes our own contract, reverses the reading of one HotpotQA span, and assigns the exits. DIVERGENCE and CONTAINMENT are not observed among the twenty, for reasons of corpus rather than taxonomy (Section~\ref{sec:discussion}).

\paragraph{Power, and what it can certify.} A paired interval separates a $0.15$ gap from zero only if its half-width is smaller, and on two of the sixteen powered contracts it is not --- the motivating contract
and JailbreakBench, both at $0.157$. Of the two contracts carrying a non-clear verdict, the 50-character HotpotQA span (436 construct positives, gap half-width $0.034$) is comfortably powered and the motivating contract is not; OR-Bench, withheld by resolution, would have been (473 positives, $0.090$). The equivalence step turns the remaining power shortfalls into explicit non-verdicts: JailbreakBench and the 80-character span carry surviving gaps but construct AUCs the band cannot certify, and their undecided status measures the design's power; Appendix~\ref{app:cutoff} sweeps the margin that decides it. Prefix containment is not specific to our model pair:\label{sec:boundary} under the same prefix-50 score, token-matched cross-pair rows outside the twenty run from NO FLAG to a DIVERGENCE-sized $\Delta_{\mathrm{AUC}}$ as the pair changes, reported gaps only, since the deciding control was not run on those rows (Appendix Table~\ref{tab:model-pair-boundary}). Nor is it automatic. A ten-pair panel over five models from four vendor families, instantiated at the \emph{final} span rather than a prefix and therefore a different contract, puts two rows' reported gaps past the CAUTION constant and none at DIVERGENCE (Appendix Table~\ref{tab:pair-panel-full}). The audit is also not specific to safety: applied to two correctness-routing contracts it returns aligned controls (Appendix~\ref{app:noncorrectness}). Nor is the instrument the finding. Replacing the TF-IDF distance on the motivating contract with a MiniLM or e5 prefix-50 encoder shrinks the reported gap the same way, from $+0.33$ and $+0.36$ to $+0.06$ and $+0.09$ under the disjoint control, both above their own null ($p{=}0.039$ and $0.017$), but leaves the residual below the CAUTION constant, so both encoder contracts clear; on OR-Bench both encoders fail the resolution precondition as the TF-IDF score does (Appendix Table~\ref{tab:encoder-contracts}).

\paragraph{Composition of the surviving agreement.} The three-target comparison of Section~\ref{sec:mechanism}, on the primary run with labels held fixed and the published strip patterns reused:

\begin{center}\small
\begin{tabular}{lrrr}
\toprule
Target & intact & stripped & drop \\
\midrule
$z$, prefix-50 keyword (same span) & 0.985 & 0.645 & $-$0.340 \\
$z^{+}$, full-output keyword & 0.855 & 0.570 & $-$0.285 \\
$y$, adjudicated construct & 0.592 & 0.428 & $-$0.164 \\
\bottomrule
\end{tabular}
\end{center}

The strip keeps 88.2\% of characters and empties 1.3\% of sides. Its cost to the construct is real: after stripping, the construct AUC is $0.428$ (95\% interval $[0.348,0.507]$), a point estimate below chance whose interval does not establish below-chance ranking. What the comparison licenses is the asymmetry---the opening is worth $0.285$ to
the full-output keyword reading, which includes the span, and $0.164$ to the construct---not a claim that the strip is
construct-neutral. On OR-Bench, stripping the longer opening-template family from both sides removes a third of all characters and the random-deletion control falls almost as far; that change is a volume effect, and is why the volume-matched recipe is the test. Reproduction: \texttt{\detokenize{scripts/opening_strip_span_independent.py}}.

\paragraph{Correcting the judge's over-call.} The label-noise study of
Appendix~\ref{app:label-noise} injects error into $y$, which raises its prevalence; the over-call of
Section~\ref{sec:label-reliability} is the opposite problem and needs the opposite test. We
down-sample judge positives from 46 to the 13 that human prevalence implies. Which 46$-$13 to remove
is not identified, so we bracket it:

\begin{center}\small
\begin{tabular}{lrrrr}
\toprule
Removal rule & $AUC(s,y)$ & $\Delta_{\mathrm{AUC}}$ & $\Delta_{|\cdot|}$ & routed \\
\midrule
random (1{,}000 draws) & 0.585 & $+$0.400 & $+$0.400 & 1.7 \\
remove the highest-scoring positives & 0.322 & $+$0.663 & $+$0.307 & 0.0 \\
remove the lowest-scoring positives & 0.793 & $+$0.192 & $+$0.192 & 6.0 \\
\bottomrule
\end{tabular}
\end{center}

Every arm stays past the $0.15$ cutoff on the orientation-robust gap \emph{under the superseded $\Delta_{\mathrm{AUC}}$ rule}, so the over-call does not account for the diagnosis. The same check under the deciding control is in Section~\ref{sec:label-reliability} and Appendix~\ref{app:reliability-details}: there the residual's magnitude is similar under random down-sampling, but the CAUTION verdict itself is not. The middle row is a caution about our own reporting: its raw gap is
\emph{larger} than before correction, because stripping the top-ranked positives pushes $AUC(s,y)$
below chance, and reading that as strengthened evidence would be the orientation error this paper
flags elsewhere. Reproduction: \texttt{\detokenize{scripts/judge_overcall_downsample.py}}.

\paragraph{Which control decides.} An earlier arrangement of this procedure
computed the verdict from $\Delta_{\mathrm{ext}}$. That is not defensible once Section~\ref{sec:split} shows the extended proxy leaves containment in. Verdicts are therefore computed from $\Delta_{\mathrm{dis}}$ wherever the disjoint proxy passes its preconditions, and where no off-span proxy exists the contract is UNDECIDABLE (a required field is missing, Figure~\ref{fig:verdict-tree}) with $\Delta_{\mathrm{AUC}}$ reported beside it under a \emph{containment-unverified} mark; it is likewise UNDECIDABLE where an off-span proxy exists but its complement is degenerate. Eight of the eleven external contracts take that branch, because their scores read the complete output or their pruned candidates are not retained (Table~\ref{tab:verdict-tally}); RouteLLM is the separate case, an off-span proxy whose complement is degenerate, the same exit the five long HotpotQA spans take. The verdicts this costs are tabulated in the main text (Table~\ref{tab:which-control}).

The consequence for this paper is direct: the contract we dissect at length is no longer a DIVERGENCE. Its residual after the disjoint control is
$+0.101$ $[-0.059,+0.255]$, short of the cutoff and not separable from zero, so it is CAUTION. OR-Bench, which we did not construct as a failure, carried DIVERGENCE under the two-branch rule and would carry NO DEMONSTRATED CONSTRUCT RANKING under the rule of Section~\ref{sec:verdict}: its construct AUC is shown at chance by the equivalence test ($AUC(s,y)=0.517$, 90\% interval inside the band) while the off-span proxy stays ranked. Once the resolution precondition is run on every row it carries no verdict at all, because one score value holds $64.4\%$ of its pairs. JailbreakBench carried DIVERGENCE under the same two-branch rule, but its construct AUC ($AUC(s,y)=0.508$ on 27 positives) cannot be certified at the band, and the final rule reports it undecided. They rest on 473 and 27 construct positives against the motivating contract's 46. Reproduction:
\texttt{\detokenize{scripts/verdict_rule_and_ablation.py}}.

\paragraph{Comparison with a cheaper baseline.} A cheaper prescription is available to any author: report $AUC(s,y)$ beside the proxy AUC, and report $\kappa(z,y)$. We implemented that rule under the same three preconditions, flagging the label when proxy and construct disagree and the score when it does not rank the construct, and applied it to all twenty contracts (\texttt{\detokenize{scripts/baseline_same_preconditions.py}}). It flags the two rows the audit flags, our motivating contract and the 50-character HotpotQA span, and in addition the five GSM8K spans from 120 to 600 characters as label failures, rows on which the off-span reading finds no residual gap because the same rule read beyond the span coincides with the construct; that is a different measurement, and it does not certify the same-span proxy the baseline flagged. What the cheap rule cannot do is split the two at-chance exits or reverse a reading.

The two analyses serve different purposes, so agreement between their flags is expected. $\kappa$ and $AUC(s,y)$
\emph{detect}: they say a proxy is a poor stand-in. They cannot \emph{attribute}, since neither sees
the span, so neither can say whether a large reported gap came from a shared span or from a validity
failure---and those call for different repairs, one to the contract and one to the label. Our primary
contract is the case in point: both flag it either way, and only the control takes its reported gap from $+0.393$ to $+0.101$, so we demote it. Run without the preconditions, the cheap rule would also have kept the rows our orientation and minimum-count checks withdrew.

Four readings follow. First, four of the seven refusal settings cannot carry the test at all: they
hold three to nine construct positives, and AdvBench's headline $\Delta_{\mathrm{AUC}}{=}{+}0.467$ is
an artifact of $AUC(s,y){=}0.290$ on three positives rather than of a proxy ranking well. Second,
among the testable settings the refusal family survives the partial control unanimously; under the disjoint control two rows remain DIVERGENCE-sized ($+0.248$ and $+0.245$, on 473 and 27 positives at $n{=}1319$ and $n{=}200$) while the primary XSTest row retains a CAUTION-sized residual ($+0.101$), so what the family loses in breadth it keeps in power. Third, the
correctness contracts largely do not survive: no gap among the five testable rows with a positive reported gap (HotpotQA at 50 and 80 characters, GSM8K at 120, 160 and 200) survives the partial control and two, the HotpotQA rows, survive the disjoint one, and the five long-span rows are excluded outright because
truncation leaves their complement empty on up to 41\% of sides. Fourth, and the reason both columns
are printed, the controls do not agree about which contracts pass. They correlate strongly over the eleven rows that carry both controls after the minimum-count and complement preconditions (Spearman $\rho{=}0.94$, $p{<}10^{-4}$; \texttt{\detokenize{scripts/control_agreement_spearman.py}}), so a reader who saw only one would think the other redundant, yet the
motivating contract's gap survives the partial control and not the disjoint one while the 50- and 80-character
correctness rows do the reverse.

That inverting setting isolates the field responsible. XSTest 100 at a
512-token generation cap reports $\Delta_{\mathrm{AUC}}{=}{+}0.369$ and returns
$\Delta_{\mathrm{ext}}{=}{-}0.180$ in the raw signed convention ($-0.054$ orientation-robust, the convention Table~\ref{tab:span-independent} prints): with the cap, refusal vocabulary that the opening predicts is often the
\emph{only} vocabulary the truncated output contains, so the full-output rule and the prefix rule
cease to track each other. The primary run, same benchmark and same score family under an uncapped
protocol, returns $+0.263$. Generation protocol is a contract field, and this is what it decides.

\paragraph{The contribution of the disjoint proxy.} A control that returns nothing is uninformative if
the rule it applies cannot fire on the span it is given, so we report the disjoint proxy's own
statistics beside its gap. Across the ten rows that pass both preconditions, $z^{c}$ has prevalence
between $0.027$ and $0.422$ and is never constant, so no row's $\Delta_{\mathrm{dis}}$ is an artifact
of an inert rule. Its agreement with the construct separates the two families cleanly. The three
refusal contracts give $\kappa(z^{c},y) = -0.008$, $0.000$ and $0.013$: read off the scored span, a
keyword rule is still almost unrelated to an adjudicated refusal decision, and the score's
$AUC(s,z^{c})$ of $0.693$, $0.765$ and $0.753$ is therefore genuinely about off-span refusal
vocabulary. The five GSM8K rows give $\kappa(z^{c},y) = 0.62$--$0.67$: a gold-answer match evaluated
over the remainder of a solution is close to the correctness label itself, so
$\Delta_{\mathrm{dis}}$ is near zero there --- the proxy and the construct have nearly merged, the
legitimate configuration of Claim 4 (Appendix~\ref{app:proposition}). HotpotQA sits between them ($\kappa{=}0.146$ at 50
characters, $0.105$ at 80). Reproduction: \texttt{\detokenize{scripts/complement_span_proxy_control.py}}.

\paragraph{Counts rather than a ratio.} A natural summary of this table is the median share
of the reported gap the control removes, $1-\Delta_{\mathrm{ext}}/\Delta_{\mathrm{AUC}}$. We do not
report it. The ratio is undefined at zero, unstable near it, and hard to interpret when $\Delta_{\mathrm{AUC}}$ is negative, and once the two preconditions of Section~\ref{sec:split} are enforced its median exceeds
one --- the control does not shrink these gaps so much as reverse them, which a percentage cannot
express. The twelve span-sweep rows also come from two tasks and one model pair, so a pooled median
over nineteen rows would treat correlated rows as independent evidence. Counts of contracts surviving
each control, stratified by family, are what the table supports.

\section{Span Dose--Response on Correctness Contracts}
\label{app:span-dose}

\paragraph{The two rows our preconditions withdrew.} At a 50-character GSM8K span both models emit a
fixed opening --- five distinct prefixes across 1{,}319 Qwen traces, six across the Gemma ones --- and the
score takes twelve distinct values with all eight proxy positives sharing one, so an AUC there is
pinned by the tie convention. At 120--200 characters the raw gap is positive ($+0.195$, $+0.144$,
$+0.116$) while the construct AUC is $0.374$, $0.347$, $0.358$: the score carries more information
about the construct ($0.13$--$0.15$) than about the proxy ($0.01$--$0.07$), with the sign reversed.
Five of the seventeen rows examined by the orientation audit (the set of that audit, distinct from
Appendix~\ref{app:null}'s seventeen) change under the check --- these three and two
AdvBench rows the paper already declines to use as ranking evidence.

Section~\ref{sec:sweep} reports that truncating the cleared correctness contracts re-opens the proxy--semantic gap. This appendix gives the full sweep (Table~\ref{tab:span-dose}). Traces are generated with each model's chat template (thinking enabled for Qwen, matching the cached phases), greedy decoding, 768 new tokens, on 1{,}319 GSM8K and 2{,}000 HotpotQA prompts; the construct $y$ is judged correctness disagreement (the two models' per-trace correctness labels differ), the proxy $z$ at a span is gold-answer-string disagreement over that span, and the score is TF-IDF cosine distance between the two models' spans. Judge correctness rates are 70.4\%/90.8\% (Qwen/Gemma) on GSM8K and 74.6\%/69.1\% on HotpotQA. CIs are paired bootstraps of the AUC gap; sem@$B$ counts judged disagreements in the top-10\% queue.

\begin{table}[!htbp]
\centering
\scriptsize
\begin{tabular}{llrrrrrrl}
\toprule
Task & Span & $z{+}$ & AUC$(s,z)$ & AUC$(s,y)$ & $\Delta_{\mathrm{AUC}}$ & $\Delta_{|\cdot|}$ & cov. & precondition \\
\midrule
GSM8K & 50 & 8 & -- & -- & -- & -- & 0.003 & \emph{score degenerate} \\
GSM8K & 80 & 0 & -- & -- & -- & -- & 0.011 & \emph{no proxy variation} \\
GSM8K & 120 & 12 & 0.569 & 0.374 & $+$0.195 & $-$0.058 & 0.015 & \emph{reversed} \\
GSM8K & 160 & 49 & 0.491 & 0.347 & $+$0.144 & $-$0.144 & 0.024 & \emph{reversed} \\
GSM8K & 200 & 79 & 0.473 & 0.358 & $+$0.116 & $-$0.116 & 0.030 & \emph{reversed} \\
GSM8K & 250 & 91 & 0.576 & 0.635 & $-$0.059 & $-$0.059 & 0.044 &  \\
GSM8K & 300 & 82 & 0.594 & 0.649 & $-$0.055 & $-$0.055 & 0.061 &  \\
GSM8K & 400 & 77 & 0.575 & 0.622 & $-$0.047 & $-$0.047 & 0.089 &  \\
GSM8K & 500 & 82 & 0.585 & 0.616 & $-$0.031 & $-$0.031 & 0.109 &  \\
GSM8K & 600 & 132 & 0.513 & 0.615 & $-$0.102 & $-$0.102 & 0.138 &  \\
GSM8K & 800 & 302 & 0.439 & 0.624 & $-$0.185 & $-$0.063 & 0.227 & \emph{reversed} \\
GSM8K & 1000 & 490 & 0.438 & 0.616 & $-$0.179 & $-$0.054 & 0.348 & \emph{reversed} \\
GSM8K & 1400 & 495 & 0.516 & 0.625 & $-$0.108 & $-$0.108 & 0.613 &  \\
GSM8K & full trace & 237 & 0.591 & 0.627 & $-$0.036 & $-$0.036 & 0.816 &  \\
\midrule
HotpotQA & 50 & 317 & 0.697 & 0.513 & $+$0.184 & $+$0.184 & 0.086 &  \\
HotpotQA & 80 & 446 & 0.644 & 0.528 & $+$0.116 & $+$0.116 & 0.133 &  \\
HotpotQA & 120 & 636 & 0.553 & 0.538 & $+$0.015 & $+$0.015 & 0.214 &  \\
HotpotQA & 160 & 761 & 0.541 & 0.528 & $+$0.013 & $+$0.013 & 0.266 &  \\
HotpotQA & 200 & 840 & 0.551 & 0.528 & $+$0.023 & $+$0.023 & 0.296 &  \\
HotpotQA & 250 & 892 & 0.555 & 0.534 & $+$0.021 & $+$0.021 & 0.320 &  \\
HotpotQA & 300 & 937 & 0.554 & 0.537 & $+$0.017 & $+$0.017 & 0.346 &  \\
HotpotQA & 400 & 957 & 0.563 & 0.533 & $+$0.031 & $+$0.031 & 0.396 &  \\
HotpotQA & 500 & 931 & 0.569 & 0.534 & $+$0.036 & $+$0.036 & 0.440 &  \\
HotpotQA & 600 & 868 & 0.566 & 0.536 & $+$0.030 & $+$0.030 & 0.479 &  \\
HotpotQA & 800 & 708 & 0.594 & 0.549 & $+$0.045 & $+$0.045 & 0.569 &  \\
HotpotQA & 1000 & 627 & 0.599 & 0.555 & $+$0.044 & $+$0.044 & 0.629 &  \\
HotpotQA & 1400 & 489 & 0.591 & 0.565 & $+$0.026 & $+$0.026 & 0.712 &  \\
HotpotQA & full trace & 436 & 0.627 & 0.603 & $+$0.024 & $+$0.024 & 0.780 &  \\
\bottomrule
\end{tabular}
\caption{Span sweep on the phase-14 correctness contracts (Qwen3.5-2B vs Gemma-4-E2B, judged
correctness-disagreement construct). Only the span read by score and proxy varies between rows.
$\Delta_{|\cdot|}=|AUC(s,z)-0.5|-|AUC(s,y)-0.5|$ is the orientation-robust gap, which differs from
$\Delta_{\mathrm{AUC}}$ exactly on the rows marked \emph{reversed}, where the score is inverted with
respect to one of its targets rather than uninformative about it. ``cov.'' is the fraction of traces
in which the gold answer has appeared by that span. The two rows the audit's preconditions remove are
marked in the last column; \emph{score degenerate} means one score value carries over half the items (at 50 characters the modal value holds $63.4\%$ of rows).}
\label{tab:span-dose}
\end{table}

Four observations follow. First, HotpotQA behaves as the account predicts and is
orientation-clean at every span: a significant positive gap at 50 and 80 characters, nothing
distinguishable from zero from 120 characters on, and $\kappa(z,y)$ moving from $-0.101$ to $+0.264$
by full trace. Under the reported gap alone the same score family, task, and construct would read as failing or clearing depending on the span; under the final rule the 50-character span is NO DEMONSTRATED CONSTRUCT RANKING and the long spans are undecidable.

Second, GSM8K does not, and the table shows why. Its 50-character row fails the resolution
precondition (twelve distinct score values on 1{,}319 items, all eight proxy positives sharing one),
its 80-character row has no proxy variation at all, and its 120--200 rows are reversed: the construct
AUC sits below chance, so $\Delta_{|\cdot|}$ is negative where $\Delta_{\mathrm{AUC}}$ is positive.
The main text withdraws their raw reading and reports them under the orientation-robust gap, which is the reading the tally carries.

Third, the negative-gap band deserves its own note. From 250 characters on, GSM8K's gap is negative,
meaning the score ranks the construct better than the proxy; two of those rows (800, 1000) are also
reversed, this time on the proxy side, so their raw magnitude overstates the effect and
$\Delta_{|\cdot|}$ is the figure to read. Proxy positives are also non-monotone in span on both tasks
(GSM8K peaks at 495 near 1{,}400 characters and falls to 237 over the full trace; HotpotQA peaks at
957 and falls to 436). This is expected rather than an error: $z$ is an XOR, so once \emph{both} models' spans contain the gold answer the disagreement disappears, and the count must fall as coverage approaches one. The construct AUC's orientation also changes along the grid: GSM8K's score ranks correctness backwards at 120--200 characters ($AUC(s,y)$ of $0.35$--$0.37$, intervals below chance) and forwards from 300 characters ($0.65$ and $0.62$), the transition falling where the traces have left their opening templates; we report the flip and do not attribute it.

Fourth, the labels are judge-robust. Re-adjudicating all 6{,}638 traces with a second judge from a
different vendor (gpt-4.1-mini) agrees with the Haiku judge at $\kappa$ 0.895/0.904 per model on
GSM8K and 0.769/0.875 on HotpotQA, reproduces the disagreement construct at $\kappa$ 0.836 and 0.614,
and changes no row's reading, including the sign of every reversed row. Reproduction:
\texttt{\detokenize{scripts/span_closure_fine_sweep.py}} for the sweep and arrival offsets,
\texttt{\detokenize{scripts/audit_score_resolution.py}} for the resolution precondition, and
\texttt{\detokenize{scripts/prefix_span_correctness_probe.py}} (\texttt{--judge} selects the label
set) for the cross-judge check; results in \texttt{\detokenize{analysis_results/}}.

\section{Budget Sensitivity and Statistical Uncertainty}
\label{app:budget-ci}

The main text reports routed composition at a single budget $B{=}55$ (the top 10\% of the $n{=}548$ clean run; on the $n{=}450$ primary run the same $B$ is the top 12\%). Table~\ref{tab:budget-sweep} sweeps $B\in\{25,55,90,135,225\}$ for the primary prefix contract (Phase-3 XSTest run) and for the revision scores on the clean run. The pattern is budget-stable in both directions. On the motivating contract the raw-prefix queue never exceeds semantic precision $0.13$ at any audited budget, and its agreed-refusal load grows with the budget (19 of 25 routed at $B{=}25$; 101 of 225 at $B{=}225$). On the clean run the revision ordering (composite $>$ final-span TF-IDF $>$ raw prefix) holds at every budget. Neither the mismatch nor the repair is an artifact of the reported operating point.

\begin{table}[!htbp]
\centering
\scriptsize
\begin{tabular}{llccccc}
\toprule
Run & Score & $B{=}25$ & $B{=}55$ & $B{=}90$ & $B{=}135$ & $B{=}225$ \\
\midrule
Primary run (XSTest 450) & raw prefix-50 TF-IDF & 3 & 6 & 10 & 15 & 30 \\
Clean run (548) & raw prefix-50 TF-IDF & 6 & 11 & 22 & 30 & 48 \\
Clean run (548) & final-span TF-IDF & 11 & 20 & 28 & 35 & 52 \\
Clean run (548) & audited composite score ($\lambda{=}0.7$) & 17 & 28 & 33 & 41 & 53 \\
\bottomrule
\end{tabular}
\caption{Semantic disagreements routed at each budget $B$ (out of $B$ escalation slots). Primary run: 46 semantic disagreements among 450 examples, Haiku labels. Clean run: 79 among 548, Haiku final-channel labels.}
\label{tab:budget-sweep}
\end{table}

Uncertainty is accounted for at three levels. First, the primary proxy--semantic AUC gap $\Delta_{\mathrm{AUC}}{=}{+}0.393$ has bootstrap 95\% CI $[0.324, 0.464]$ and permutation $p<10^{-4}$ (10{,}000 resamples each; the other six audited refusal settings give gap CIs bounded away from zero with $p<10^{-3}$, packaged in \texttt{\detokenize{analysis_results/gap_significance.json}} and \texttt{\detokenize{analysis_results/jailbreakbench_gap_significance.json}}). Second, Table~\ref{tab:pair-panel-full} reports bootstrap 95\% CIs (2{,}000 resamples) for the ten final-span panel gaps of Appendix Table~\ref{tab:pair-panel-full}: seven of ten CIs include zero, three show small positive gaps (at most $+0.128$), and every panel upper bound (at most $+0.232$) lies below the primary gap's lower bound $0.324$. The primary/panel contrast is therefore not attributable to sampling noise. Third, on the clean run the semantic AUC CIs are $0.622$ $[0.556, 0.687]$ (raw prefix), $0.675$ $[0.605, 0.742]$ (final TF-IDF), and $0.715$ $[0.642, 0.783]$ (composite); the paired AUC deltas over the raw prefix \citep[cf.\ the paired-AUC comparison problem of][]{delong1988comparing} cross zero at the 95\% level ($+0.053$ $[-0.053, +0.155]$ and $+0.093$ $[-0.011, +0.194]$). This is why the main text claims that the audited revision improves routed composition (11/55 to 20/55 and 28/55 on the clean run; the original raw-generation contract routes 6/55), not that it yields a statistically separated semantic AUC; the contract diagnosis, not router superiority, is the claim.

The paper applies threshold verdicts to tens of contracts and reports every
interval on its own, so we also correct for that. Pooling all 43 gap tests into a single
Benjamini--Hochberg family \citep{benjamini1995controlling} at $\alpha{=}0.05$---the seven audited refusal settings, the ten-pair
final-span panel, and the correctness span sweep---leaves the reading intact: every refusal setting
survives, both panel rows that cross the CAUTION cutoff survive, and HotpotQA's 50- and 80-character
rows survive; among rows carrying no claim, several small or negative long-span gaps also survive, including HotpotQA at 800 characters ($+0.045$). Two rows are significant before the correction but not after, and neither carries a claim: GSM8K's full-trace residual ($-0.036$) and
HotpotQA at 1{,}000 characters ($+0.044$). Permutation $p$-values are floored at $1/(B{+}1)$ rather
than reported as zero, since a permutation test cannot resolve below its resampling budget and using
zero would flatter the correction. Panel and sweep rows enter through a normal approximation to their
bootstrap intervals, which is an approximation and is the weakest link in this check. Reproduction:
\texttt{\detokenize{scripts/multiplicity_correction.py}}, with the JailbreakBench member computed by \texttt{\detokenize{scripts/jailbreakbench_gap_significance.py}}.

\paragraph{Bootstrap procedure, and what it does when positives are scarce.}
All intervals resample items i.i.d.\ with replacement at the original $n$, recomputing both AUCs and their difference on each resample; no stratification on the label is applied. The item is a prompt on the controlled contracts. On PRM800K the disjoint-control intervals resample problems, the unit of the mean-of-within-problem statistic (Appendix~\ref{app:prmcontainment}); on MixInstruct the recorded intervals resample comparisons, and Appendix~\ref{app:hybridllm} reports instruction-level cluster intervals beside them. A resample in which
either target is single-class leaves AUC undefined, and such resamples are discarded rather than
imputed. This matters only where positives are scarce, so we report the retention directly: of
$10{,}000$ draws, the primary XSTest contract keeps all $10{,}000$, and the worst case in the audited
set is AdvBench 520 at $9{,}491$ ($5.1\%$ discarded, driven by its three semantic positives), with
OR-Bench hard 1k at $9{,}500$ and AdvBench100 512 at $9{,}536$. Because discarding is conditional on
the resample containing at least one positive, the surviving intervals for those rows are conditional
intervals and are mildly optimistic. We therefore do not use the sparse-positive rows as standalone
semantic-ranking evidence anywhere in the paper; they appear only in operating-point composition,
where the quantity counted is a queue composition rather than a ranking statistic. The retained counts
are recorded per setting in \texttt{\detokenize{analysis_results/gap_significance.json}} under
\texttt{\detokenize{boot_n}}.

\section{Non-Refusal Correctness Audits}
\label{app:noncorrectness}
Table~\ref{tab:cross-domain-evidence} places the primary refusal mismatch alongside the two correctness-routing contracts audited below.

\begin{table}[!htbp]
\centering
\scriptsize
\resizebox{\linewidth}{!}{%
\begin{tabular}{lllrrrrrrl}
\toprule
Domain / score & Cheap proxy & Semantic construct & surf $n^{+}$ & sem $n^{+}$ & AUC proxy & AUC sem & AP sem & $\kappa(z,y)$ & sem@$B$ \\
\midrule
Refusal, prefix TF-IDF & refusal keyword disagreement & refusal-decision disagreement & 13/450 & 46/450 & 0.985 & 0.592 & 0.127 & 0.024 & 6/55 \\
Retrieved-context QA, full-trace TF-IDF & answer occurrence disagreement & correctness disagreement & 51/250 & 60/250 & 0.674 & 0.670 & 0.347 & 0.387 & 20/50 \\
Retrieved-context QA, full-trace Sentence-L12 & answer occurrence disagreement & correctness disagreement & 51/250 & 60/250 & 0.558 & 0.503 & 0.245 & 0.387 & 12/50 \\
Reasoning, full-trace TF-IDF & numeric disagreement & correctness disagreement & 76/250 & 92/250 & 0.570 & 0.595 & 0.478 & 0.304 & 24/50 \\
\bottomrule
\end{tabular}%
}
\caption{The same audit across three domains. Row 1 is the motivating refusal contract (DIVERGENCE under $\Delta_{\mathrm{AUC}}$, CAUTION under the deciding control); the correctness rows are full-trace scores, so they carry a same-span reading only and show no flag under it. $n^{+}$ = positives, sem@$B$ = construct disagreements in the top-$B$ queue.}
\label{tab:cross-domain-evidence}
\end{table}

To check that RLC-Audit is not tied to refusal or safety, we apply the same score--proxy--construct accounting to HotpotQA multi-hop correctness and GSM8K numeric correctness \citep{yang2018hotpotqa,cobbe2021training}. The score span is the full generated trace for the Qwen3.5-2B/Gemma-4-E2B-it pair; surface proxies are gold-answer or gold-number occurrence disagreements; the semantic construct is LLM-judge correctness disagreement. Because the score reads the whole trace, the complement is empty and the off-span control cannot run: under Algorithm~\ref{alg:rlcaudit} these contracts are UNDECIDABLE with a required field missing, and what follows is the same-span reading. Under that reading, and unlike the raw-prefix refusal contract, the TF-IDF correctness contracts show no flag and route 20/50 and 24/50 semantic correctness disagreements. Sentence-L12 is included for HotpotQA as a weak-utility boundary; no flag here means no residual above the thresholds rather than routing utility. Table~\ref{tab:cross-domain-evidence} places these rows next to the refusal contract under one accounting. They do not show that RLC is common outside refusal; they show that the protocol distinguishes aligned contracts from mismatched ones.

\section{Reliability and Robustness Details}

\subsection{Why the Control Moves the Proxy and Not the Score}
\label{app:symmetric}

\begin{figure}[t]
\centering
\includegraphics[width=0.72\linewidth]{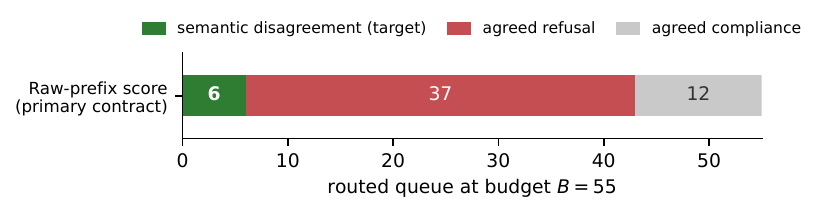}
\caption{What the budget buys on the controlled contract at $B{=}55$ (XSTest 450,
Qwen3.5-2B/Gemma-4-E2B-it, primary run); the matched clean-run baseline routes 11/55. The score's two
AUCs are in the text rather than in a second panel.}
\label{fig:contract-overview}
\end{figure}

A symmetric control suggests itself: hold the proxy fixed and re-score the complement instead. We do not use it. The score is the deployed object under audit; re-scoring a
different span does not re-examine the same contract, it constructs a new one and reports on a system
nobody runs. Whether a score would have been good had it read different text is a question about a
counterfactual system, and no release lets us answer it. The proxy is the field we are free to move:
it is a rule, evaluable over any span, and moving it keeps the audited score fixed while varying the measurement it is compared with. Section~\ref{sec:apply}'s span grid is not the symmetric control declined here: each span there is a separate contract audited on its own terms, not a re-scoring of one contract. A control must leave the audited object fixed; a contract family varies it and says so.
This asymmetry is also why NO FLAG, CAUTION, DIVERGENCE and CONTAINMENT are statements about a contract's proxy, not about the score's competence. NO DEMONSTRATED CONSTRUCT RANKING is the one exception by design: it is precisely the statement that the score is not evidence for this construct.

\subsection{Single-Draw Noise in Our Own Construct}
\label{app:singledraw}

Our construct is read off one greedy generation per model per prompt: each side is adjudicated by the
judge and the pair-level label is the XOR of the two side labels. \citet{chen2026routinggap} shows
that a construct of this shape carries noise a second draw would expose, and attributes 12--36\% of a
routing benchmark's oracle gap to it. The criticism transfers to this paper intact, and none of our
reliability work answers it: human side labels, the alternative judge and the label-noise injection
all vary the \emph{labeling} of a fixed set of generations, while single-draw noise varies the
generations themselves. Settling it requires $k$ samples per side, re-adjudicated, with the contract
re-audited over the resulting distribution of labels. We have not run that.

Two structural facts bound how far it could move a verdict, and we state them as bounds rather than
as a defence. First, the pair-level event is an XOR, so any resampling that shifts both sides in the
same direction leaves the label unchanged; only draws that flip one side matter. Second, the construct AUC enters $\Delta_{|\cdot|}$ with a negative sign, so added label-independent construct noise depresses $AUC(s,y)$ and therefore \emph{raises} $\Delta_{|\cdot|}$. That is the direction that hurts us: it
would inflate the motivating contract's gap toward the mismatch band, and inflate the two
non-clear verdicts we report. Under label-independent noise a construct measured with more draws would make our verdicts more conservative, not less; under score- or class-dependent noise the direction is not fixed, and it is unmeasured here.
\label{app:reliability-details}

\paragraph{Down-sampling protocol and extremes.} The main-text down-sampling (Section~\ref{sec:label-reliability}) removes judge positives from 46 to the 13 the human rate implies; the 2.8\% is an extrapolation, since the resolved sample is enriched toward high-scoring prompts and the rate is applied to all 450. Under removal independent of the score, $\Delta_{\mathrm{dis}}$ averages $+0.106$ over 1{,}000 draws (2.5--97.5\% band $[+0.008,+0.187]$) beside the recorded $+0.101$, and exceeds the CAUTION constant on 56.6\% of draws --- a threshold-crossing rate, not a verdict. Re-running the whole procedure on each of 200 draws, with the equivalence and orientation tests and the permutation null recomputed on the down-sampled labels, returns UNDECIDABLE on 75.0\%, NO FLAG on 24.5\% and CAUTION on 0.5\%. The down-sampling is a sensitivity analysis under an assumed error rate, not a correction of identified errors, and the verdict survives perturbations that preserve the recorded disagreement rate while mostly withdrawing under the human-implied rate. The directional extremes move the construct AUC itself, to $0.32$ when the score's highest-ranked positives are removed and $0.79$ when its lowest are, so they change which orientation branch the contract takes rather than shrinking one gap (Appendix~\ref{app:label-noise}).

Section~\ref{sec:label-reliability} states the reliability result and its consequence; Table~\ref{tab:human-reliability-main} gives the per-check accounting behind it. One check has no main-text counterpart and belongs here: the prefix-only human pass found the routed prefixes themselves insufficient for semantic adjudication, direct evidence that prefix evidence cannot license a semantic routing claim --- the conclusion the score-side analysis reaches independently.

\begin{table}[!htbp]
\centering
\small
\resizebox{0.88\linewidth}{!}{%
\begin{tabular}{lrrrr}
\toprule
Resolved human--judge check & $n$ & exact agreement & $\kappa$ & Role \\
\midrule
Qwen side refusal & 109 & 0.972 & 0.930 & side-label validation \\
Gemma side refusal & 109 & 0.899 & 0.715 & side-label validation \\
Pair refusal disagreement & 109 & 0.872 & 0.266 & harder pair construct \\
Prefix-only sufficiency pass & 258 & -- & -- & prefixes insufficient for semantic adjudication \\
\bottomrule
\end{tabular}%
}
\caption{Human reliability checks for the semantic adjudication layer. The side-label agreement supports the refusal labels used to derive pair labels; the weaker pair row is why robustness checks are reported.}
\label{tab:human-reliability-main}
\end{table}

Table~\ref{tab:label-noise} summarizes the main robustness checks. The XSTest diagnosis is not removed by small prevalence-preserving label changes: erasing the default AUC-gap diagnosis would require 36 pair-label flips, while random 20\% semantic-label flips, independent of score and class, drive semantic AUC toward chance. Sparse AdvBench rows are not used as standalone semantic-ranking evidence, but they are useful operating-point evidence because the routed queue is dominated by agreed refusals. These checks support the measurement conclusion without changing the scope of the claim.

The alternative-explanation controls clarify what RLC is not, and Table~\ref{tab:boundary-controls} collects them with their outcomes: the split is not only a lexical TF-IDF shortcut, not only a batch-fitting artifact, not only an orientation artifact, and not erased by moving to matched larger-pair scopes. These are not separate contributions; they protect the central measurement claim.

\section{Prevalence Mini-Survey: Proxy-Only Evidence in Routing and Safety Evaluation}
\label{app:prevalence}

This appendix is the survey behind the release measurement of Section~\ref{sec:release}, and it supports no prevalence claim: it records what routing and safety papers report, not how often coupling occurs. The audited failure mode is not specific to our setup: the validation configuration that permits it appears in standard practice. For each surveyed paper we extracted the same contract fields the audit uses: the routing or detection score $s$, the label $z$ used to train or evaluate it, the semantic construct $y$ the paper claims to target, and whether the paper validates $z$ against $y$ in its own setup (as opposed to citing correlations measured elsewhere).

\paragraph{Sample, coding and results.} The tables below cover systems we
had already cited, which supports a statement about those systems and no more. To put a number on the
practice we drew a sample. Six fixed arXiv queries on LLM routing, cascades, deferral and selective
prediction returned 178 distinct papers from 2023 onward; abstract-level screening for ``makes a
per-query decision among models or between answering and deferring, and evaluates that decision
against a label'' left 125, with every exclusion reason recorded. From those we drew 40 at a fixed
seed, retrieved all 40 in full text, and read them. Two were out of scope once read (one routes retrieval granularity inside a single agent, one is a speech-retrieval benchmark), leaving 38. Table~\ref{tab:survey-flow} in the main text gives the flow; the query strings, the screening sheet with every exclusion reason, and the per-paper codings are in the artifact (\texttt{\detokenize{analysis_results/prevalence_pool.json}}, \texttt{\detokenize{prevalence_codings.json}}, \texttt{\detokenize{prevalence_evidence.json}}). Coding was done by one reader; the codings are released so that a second reader can recode them, and the conclusions below are about this sample only.

Each was coded into one of three categories, defined before coding. \textbf{A}: reports an in-setup
measurement relating its operative label to a construct-level label---a human study on its own data,
an agreement statistic against expert judgment, or explicit modeling of the gap between a proxy and
a gold label. \textbf{B}: names the gap but does not measure it. \textbf{C}: neither. Validity by
citation---``LLM judges have been shown to agree with humans [ref]''---is category C, because a
citation to someone else's correlation on someone else's data is the substitution this paper is about.

\paragraph{The inference, in the authors' own words.} The step from operative label to construct is made explicitly in the sample. DiSRouter reads a classification metric against its own label as ``its ability to accurately discern solvable queries''; a cross-attention router on RouterBench states that it ``predicts both response quality and generation cost'' where the fitted quantity is a score column; Hybrid LLM routes ``based on the predicted query difficulty'' while training against a BARTScore gap. In each case the proxy-to-construct step is assumed rather than evaluated in the reported setup. The contract makes that step explicit.

\paragraph{A second coded question: can the control be run at all?} The survey above codes whether a
paper checks its label. Elsewhere we claim something different --- that nobody releases enough for a
reader to re-evaluate a proxy off the scored span --- and an earlier draft rested that claim on this
survey without having coded it. We code it here. Two conditions are separable and we record them
separately: \textbf{R1}, the paper releases per-example model generations, without which there is no
span to re-read; and \textbf{R2}, the proxy rule is reproducible from the paper, whether as a keyword
list, a string-match recipe, an exact metric definition, or a released scorer.

R2 holds for 17 of 38 ($44.7\%$). R1 holds for 0 of 38 (95\% CI $[0.0,9.2]$), and so does the
conjunction. One keyword hit was a false positive on reading --- ``we utilize third-party API providers
to obtain model outputs'' describes how the authors obtained outputs, not what they publish --- and a
second sweep for dataset-release phrasing surfaced four further papers, all read individually: two
link input benchmarks, one cites a software dependency's archive, and one releases router training
data rather than scored generations. A paper that released generations without saying so in its text
would be missed, so this is a lower bound on release. The asymmetry between R2 and R1 is the useful
part: the methods sections are adequate and the artifacts are not, which makes the recommendation
concrete. Releasing per-example generations alongside the operative label is what makes the control runnable by a third party. Reproduction:
\texttt{\detokenize{scripts/prevalence_survey_artifacts.py}}.

Three of 38 fall in A ($7.9\%$, Wilson 95\% CI $[2.7,20.8]$): one models the shift between
gold-standard and preference-based labels directly, one reports quadratic-weighted $\kappa$ against
expert graders alongside human inter-rater reliability, and one runs a 200-sample human check finding
its automatic label matched on 92\% of cases. Two more fall in B, including one that names its lexical
quality signal as ``a weak surrogate for true answer quality'' without measuring the gap. The
remaining 33 are C. Including B, $13.2\%$ $[5.8,27.3]$ acknowledge the gap at all.

Two cautions on this number. It measures reporting, not coupling: a paper that never checks its label
may still have a sound one, and nothing here says how often labels are wrong. And 38 is small---the
interval is wide, and a reader who disagrees with a coding can recompute, since every paper's
category, the passage it rests on, and the screening decisions are released in
\texttt{\detokenize{analysis_results/prevalence_codings.json}} with the pipeline in
\texttt{\detokenize{scripts/prevalence_survey_*.py}}.

\paragraph{Routing and cascade systems.}
Table~\ref{tab:prevalence-routing} summarizes eight routing and cascade systems: the claimed construct is semantic, the operative label is a text-derived proxy, and validation is delegated by citation. The closest analyses to in-setup validation treat score--label disagreement as noise to mitigate \citep{ding2024hybrid,lu2024zooter,gupta2024language}. Three of the eight also expose score and label to shared evidence --- the router of \citet{ding2024hybrid} reproduces the BARTScore that evaluates it, \citet{chen2023frugalgpt}'s scorer reads the answer its string match grades, and \citet{aggarwal2024automix} verifies on the span whose overlap defines F1 --- which is the configuration this paper audits.

\paragraph{Safety labels.}
Table~\ref{tab:prevalence-safety} summarizes nine safety-evaluation resources, which show what proxy validation finds when it is performed: the canonical keyword attack-success metric \citep{zou2023advbench} was adopted without human validation and later found unstable \citep{mazeika2024harmbench,souly2024strongreject}, and each validated resource in the table exists because an earlier proxy was found to diverge. \citet{rottger2024xstest} document the survival mechanism: string matching mislabels partial refusals instance by instance while preserving model rankings, the aggregate/instance dissociation the routed-composition check is designed to expose.

\begin{table}[!htbp]
\centering
\scriptsize
\begin{tabular}{p{2.4cm}p{2.9cm}p{3.0cm}p{1.8cm}p{1.7cm}}
\toprule
System & Score $s$ & Proxy label $z$ & Claimed $y$ & In-setup $z$-vs-$y$ validation \\
\midrule
FrugalGPT \citep{chen2023frugalgpt} & DistilBERT scorer on query+answer text & task-specific answer match & accuracy, performance & none \\
Hybrid LLM \citep{ding2024hybrid} & DeBERTa router predicting quality gap & BARTScore & response quality & judge-vs-judge only \\
MoT cascades \citep{yue2024mot} & answer consistency across sampled traces & exact match & correctness, difficulty & none \\
Token-level cascades \citep{gupta2024language} & token log-prob aggregates & string match; BLEURT & quality & none (score-side length audit) \\
Zooter \citep{lu2024zooter} & router distilled from reward model & reward-model scores; GPT-4 judge & expertise, performance & none (noise smoothing) \\
AutoMix \citep{aggarwal2024automix} & few-shot LLM self-verification & F1 / accuracy / exact match & correctness & none \\
RouterBench \citep{hu2024routerbench} & benchmark for arbitrary routers & exact match; GPT-4 judge & performance & none \\
RouteLLM \citep{ong2025routellm} & learned win-probability routers & human preferences; GPT-4 judge & response quality & none (validity by citation) \\
\bottomrule
\end{tabular}
\caption{Prevalence survey of routing and cascade systems. All eight claim a semantic construct; seven operationalize it through a text-derived proxy, and RouteLLM is the exception, training on human preference labels while augmenting with a GPT-4 judge.}
\label{tab:prevalence-routing}
\end{table}

\begin{table}[!htbp]
\centering
\scriptsize
\begin{tabular}{p{2.7cm}p{3.4cm}p{2.2cm}p{3.9cm}}
\toprule
Resource & Proxy label $z$ & Claimed $y$ & Validation / divergence evidence \\
\midrule
AdvBench keyword ASR \citep{zou2023advbench} & fixed refusal-substring list (``I'm sorry'', ``I cannot'', \ldots) & attack success & none; later found unstable \citep{mazeika2024harmbench,souly2024strongreject} \\
XSTest \citep{rottger2024xstest} & human annotation (primary); string match (secondary) & refusal decision & human $\kappa$ 0.93--0.97; string match shown to mislabel partial refusals while preserving rankings \\
HarmBench \citep{mazeika2024harmbench} & fine-tuned classifier & attack success & 93\% human agreement; documents 30\% keyword-ASR shifts from length alone \\
Llama Guard \citep{inan2023llamaguard} & classifier on single red-team labels & harmfulness & benchmark transfer only \\
WildGuard \citep{han2024wildguard} & fine-tuned classifier & harm + refusal & human-annotated test set; +26\% refusal detection over prior tools \\
SORRY-Bench \citep{xie2025sorrybench} & fine-tuned 7B judge & refusal decision & meta-evaluated on 7K+ human annotations \\
OR-Bench \citep{cui2025orbench} & keyword tier (80K); LLM judges (subsets) & over-refusal & keyword tier unvalidated; misses indirect refusals \\
JailbreakBench \citep{chao2024jailbreakbench} & LLM judge & jailbreak success & judge selected by agreement with $\sim$300 human labels \\
StrongREJECT \citep{souly2024strongreject} & usefulness-scoring evaluator & delivered harm & validated vs.\ humans; shows prior proxies overstate success \\
\bottomrule
\end{tabular}
\caption{Prevalence survey of safety-evaluation labels: proxy, claimed construct and reported validation evidence for nine resources.}
\label{tab:prevalence-safety}
\end{table}

\section{What to Report}
\label{app:reporting-guidance}

The audit is stricter than reporting one score--label correlation: it asks the same ranked list to be evaluated against the proxy, against the construct, and at the stated budget. Table~\ref{tab:audit-components} lists what each component answers and the minimum a paper must report for it. Naming the score span and the proxy span is a precondition for every row.

\begin{table}[!htbp]
\centering
\small
\resizebox{\linewidth}{!}{%
\begin{tabular}{llll}
\toprule
Audit component & Question answered & Failure hidden if omitted & Minimal output \\
\midrule
Proxy metric & Does the score predict the available proxy? & proxy utility is unknown & AUC/AP against $z$ \\
Semantic metric & Does the same ranking predict the claimed construct? & proxy success is mistaken for semantic evidence & AUC/AP against $y$ \\
Agreement & Are proxy and semantic labels measuring the same event? & two constructs are silently conflated & prevalence and $\kappa(z,y)$ \\
Routed composition & What does the top-$B$ router actually escalate? & AUC hides operating-point failure & semantic and surface counts in $R_B(s)$ \\
Span intervention & Where does the apparent evidence originate? & artifact-bearing evidence is mistaken for construct evidence & re-audited score after span relocation \\
Aligned control & Can the protocol clear a contract at all? & the audit only ever flags failures & one contract expected to clear \\
\bottomrule
\end{tabular}%
}
\caption{Why each RLC-Audit component is necessary, and the minimum a paper must report for it. Naming the score span and the proxy span is a precondition for every row.}
\label{tab:audit-components}
\end{table}
\section{Reproducibility}
\label{app:reproducibility}

\begin{table}[t]
\centering
\scriptsize
\resizebox{\linewidth}{!}{%
\begin{tabular}{lllll}
\toprule
Run & $n$ & Protocol / span & Budget & Headline numbers \\
\midrule
Primary run (phase-3) & 450 & raw prefix-50 & $B{=}55$ (12.2\%) & AUC 0.985/0.592, routed 6/55 \\
Clean run (phase-13) & 548 & tagged, raw and final spans & $B{=}55$ (10\%) & baseline 11/55, final 20/55, composite 28/55 \\
Pair panel & 450 & tagged final span & $B{=}45$ (10\%) & Appendix Table~\ref{tab:pair-panel-full} \\
\bottomrule
\end{tabular}%
}
\caption{Run-to-budget mapping for the refusal contracts. Throughout the paper these three runs are
named \emph{primary run}, \emph{clean run}, and \emph{pair panel}.
$B{=}55$ is one absolute budget: 10\% of the clean run and 12.2\% of the primary run.}
\label{tab:run-map}
\end{table}

The anonymized artifact bundle contains \texttt{\detokenize{scripts/}}, \texttt{\detokenize{data/}}, \texttt{\detokenize{analysis_results/}}, \texttt{\detokenize{figures/}}, and an upload manifest. The packaged path reproduces the reported tables without rerunning model generation or paid judge calls. Main entry points include \texttt{\detokenize{rlc_audit_cli.py}}, \texttt{\detokenize{analyze_rlc_audit.py}}, \texttt{\detokenize{analyze_label_noise_robustness.py}}, the mechanism scripts, \texttt{\detokenize{analyze_regeneration_phase.py}}, and \texttt{\detokenize{analyze_token_matched_prefix_crosspairs.py}}; composite decomposition and $\lambda$ sensitivity are generated by \texttt{\detokenize{analyze_rlc_composite_router.py}}. Cached judge outputs, bootstrap/permutation reports, label-noise reports, and generated summaries are stored in \texttt{\detokenize{analysis_results/}}. The artifact card and manifest are the source of truth for provenance: they list Hugging Face model IDs, cleanup-time pinned revisions where available, cached judge aliases, dataset sources/licenses, and the packaged files used by each reproduced table.

The packaged runs, their identifiers and the generation settings behind each are listed in the artifact's README; Table~\ref{tab:run-map} maps the primary, clean and panel runs to the sections that use them.

\clearpage
\begin{landscape}
\begin{table}[p]
\centering\small
\setlength{\tabcolsep}{3pt}
\renewcommand{\arraystretch}{1.15}
\begin{tabular}{p{3.0cm}p{3.3cm}p{2.6cm}p{3.6cm}p{3.4cm}p{3.1cm}p{3.2cm}}
\toprule
Contract & score representation & score span & proxy label & construct & model pair & protocol / budget \\
\midrule
Primary (XSTest 450) & prefix TF-IDF cosine distance & first 50 chars & same-span refusal-keyword disagreement & judged refusal-decision disagreement & Qwen3.5-2B / Gemma-4-E2B-it & tagged greedy, thinking off; $B{=}55$ \\
Refusal variants (6) & same & first 50 chars & same & same & same pair & same; $B{=}55$ \\
Correctness spans (13) & same & 50--600 chars, full & gold-answer-string disagreement over the span & judged correctness disagreement & same pair & chat template, greedy, 768 tokens; no budget audited \\
Pair panel (10) & final-span TF-IDF distance & parsed final channel & same-span keyword disagreement & judged refusal-decision disagreement & five models, four vendors & tagged greedy; $B{=}45$ \\
PRM800K & product-aggregated PRM prefix score & first $k$ steps; full solution & gold answer inside the span & sympy-graded final correctness & released sampler and PRM & as released; best-of-$n$ \\
Math-Shepherd & rollout step label & steps $1..k$ & positive step label & step correctness (as consumed) & released generator & as released \\
Search-and-learn & 8B PRM step score & partial chain & step score at a span fraction & graded final correctness & Llama-3.2-1B $+$ 8B PRM & beam / DVTS / best-of-$n$, five seeds \\
RouterBench (3 slices) & response distance & stored responses & released label column & independent adjudication & claude-instant-v1 / Mixtral-8x7B & as released; $10\%$ budget \\
Hybrid LLM / MixInstruct & BARTScore quality gap & full responses & sign of the BARTScore gap & separately elicited adequacy verdict & all released pairs with $\geq 200$ rows & as released; operating point checked \\
UltraFeedback & preference selection target & full completions & released preference & truthfulness sub-rating & completions of one instruction & as released \\
RouteLLM (gpt4\_dataset) & 200-char prefix distance & first 200 chars & same-prefix surface proxy & GPT-4 judge score & the two models of the release & as released \\
FrugalGPT-style & answer-reading scorer & query $+$ answer & string match on the answer & judged correctness & Qwen3.5-2B / Gemma-4-E2B-it (our traces) & our traces \\
\bottomrule
\end{tabular}
\caption{The seven contract fields for every audited configuration, as instantiated in the paper's text and scripts; the motivating contract is expanded with its rationale in Table~\ref{tab:contract-instantiation}. ``As released'' marks fields fixed by the artifact rather than by us.}
\label{tab:all-fields}
\end{table}
\end{landscape}

\end{document}